\documentclass[10pt,twocolumn,letterpaper]{article}

\usepackage{cvpr}
\usepackage{times}
\usepackage{epsfig}
\usepackage{graphicx}
\usepackage{amsmath}
\usepackage{amssymb}

\usepackage{subfig}
\usepackage{booktabs}

\usepackage{adjustbox} %
\usepackage{enumitem} %
\usepackage{rotating}
\usepackage{makecell}

\usepackage{siunitx}
\usepackage{amsfonts} %

\global\long\def\cE{\mathcal{E}} %
\global\long\def\btheta{\boldsymbol{\theta}} %
\global\long\def\bparams{\btheta} %
\global\long\def\bzero{\mathbf{0}}
\global\long\def\mId{\mathtt{Id}} %

\global\long\def\Warp{\mathbf{W}} %
\global\long\def\tref{t_\text{ref}} %
\global\long\def\pol{p} %
\global\long\def\IWE{I} %
\global\long\def\IWEentry{h} %

\global\long\def\Lum{L}
\global\long\def\vel{\mathbf{v}}
\global\long\def\prtl#1#2{\frac{\partial#1}{\partial#2}}

\global\long\def\R{\mathbb{R}} %
\global\long\def\bu{\mathbf{u}} %
\global\long\def\bx{\mathbf{x}}
\global\long\def\bv{\mathbf{v}} %

\global\long\def\numEvents{N_e} %
\global\long\def\numPixels{N_p} %

\global\long\def\cN{\mathcal{N}} %
\global\long\def\bmu{\boldsymbol{\mu}} %

\global\long\def\supp{\operatorname{supp}}
\global\long\def\variance{\operatorname{Var}}
\global\long\def\range{\operatorname{range}}
\global\long\def\hessian{\operatorname{Hess}}
\global\long\def\moranIndex{\operatorname{Moran}}

\global\long\def\f{\mathbf{f}}

\usepackage{xcolor}
\newcommand{\revred}[1]{#1}

\usepackage{cite}

\usepackage{titling}
\date{}\predate{}\postdate{}
\pretitle{\begin{center}\Large \bf}
\posttitle{\end{center}\vskip 0.5em}

\usepackage{url}
\usepackage[pagebackref=true,breaklinks=true,colorlinks,bookmarks=false]{hyperref}
\hypersetup{
  pdftitle={Focus Is All You Need: Loss Functions For Event-based Vision},
  pdfsubject={Computer Vision, Robotics, Optimization},
  pdfauthor={Guillermo Gallego, Mathias Gehrig, Davide Scaramuzza},
  pdfkeywords={Event Cameras, Motion Compensation, Asynchronous Sensor, Motion Estimation, High Dynamic Range, High Temporal Resolution, Low Latency}
}

\cvprfinalcopy %

\newif\ifshowpagenumbers
\showpagenumberstrue

\ifcvprfinal
\ifshowpagenumbers
\else
\fi
\fi

\definecolor{somegray}{rgb}{0.5, 0.5, 0.5}
\newcommand{\darkgrayed}[1]{\textcolor{somegray}{#1}}
\makeatletter
\newcommand*\titleheader[1]{\gdef\@titleheader{#1}}
\AtBeginDocument{%
  \let\st@red@title\@title
  \def\@title{%
    \vskip-3.45em
    \bgroup\normalfont\large\centering\@titleheader\par\egroup
    \vskip1.5em\st@red@title}
}
\makeatother

\titleheader{\darkgrayed{This paper has been accepted for publication at the\\ IEEE Conference on Computer Vision and Pattern Recognition (CVPR), Long Beach, 2019.
\copyright IEEE}}

\def\MYTITLE{Focus Is All You Need: Loss Functions For Event-based Vision}

\title{\MYTITLE}

\newif\ifaffiliationfootnote
\affiliationfootnotetrue

\ifaffiliationfootnote
\author{Guillermo Gallego \textsuperscript{\dag}
\and
Mathias Gehrig \textsuperscript{\dag}
\and
Davide Scaramuzza \textsuperscript{\dag}}
\newcommand\nomarkerfootnote[1]{%
  \begingroup
  \renewcommand\thefootnote{}\footnote{#1}%
  \addtocounter{footnote}{-1}%
  \endgroup
}
\else
\author{Guillermo Gallego, Mathias Gehrig, Davide Scaramuzza\\
Dept.~Informatics Univ.~of Zurich, and Dept.~Neuroinformatics, Univ.~of Zurich and ETH Zurich
}
\fi

\begin{document}

\twocolumn[{%
\renewcommand\twocolumn[1][]{#1}%
\maketitle
\ifaffiliationfootnote
\vspace{-2.08ex}
\fi
\begin{center}
  \centering
  \includegraphics[trim={0 0 0 2ex},clip,width=0.9\textwidth]{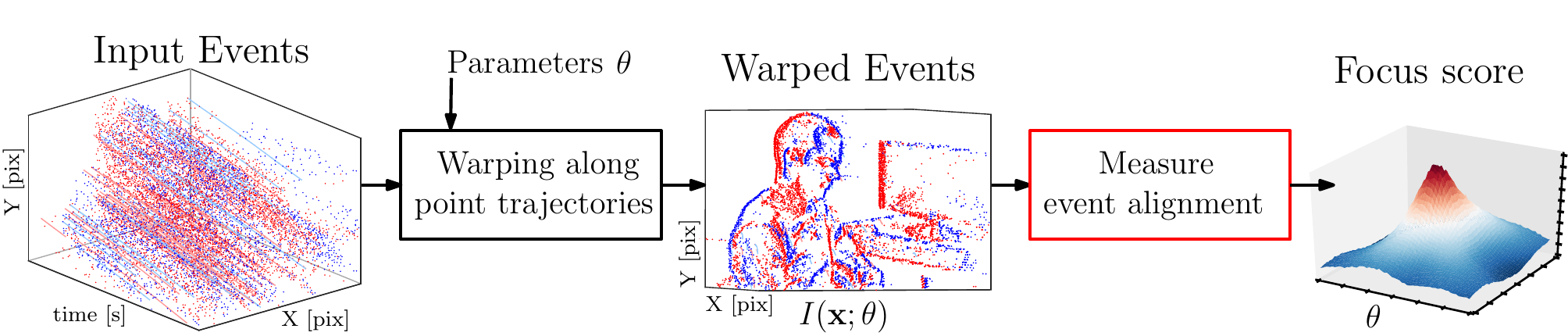}
  \captionof{figure}{\label{fig:blockDiagramSimple}
  \emph{Motion Compensation Framework}. 
  Events in a space-time window are warped according to point trajectories described by motion parameters $\btheta$,
  resulting in an image of warped events (IWE) $\IWE(\bx;\btheta)$.
  Then, a \emph{focus loss function} of $\IWE$ measures how well events are aligned along the point trajectories. 
  This work proposes multiple focus loss functions for event alignment (last block in the figure and Table~\ref{tab:focusmeasures}) for tasks such as rotational motion, depth and optical flow estimation.}
\end{center}
}]

\ifaffiliationfootnote
\nomarkerfootnote{\textsuperscript{\dag} Dept.~Informatics, Univ.~of Zurich and Dept.~Neuroinformatics, Univ.~of Zurich and ETH Zurich}
\ifshowpagenumbers
\else
\thispagestyle{empty}
\fi
\fi

\begin{abstract}
\vspace{-1ex}
Event cameras are novel vision sensors that output pixel-level brightness changes (``events'') instead of traditional video frames.
These asynchronous sensors offer several advantages over traditional cameras, such as, high temporal resolution, very high dynamic range, and no motion blur.
To unlock the potential of such sensors, motion compensation methods have been recently proposed.
We present a collection and taxonomy of twenty two objective functions to analyze event alignment in motion compensation approaches (Fig.~\ref{fig:blockDiagramSimple}).
We call them \emph{focus loss functions} since they have strong connections with functions used in traditional shape-from-focus applications.
The proposed loss functions allow bringing mature computer vision tools to the realm of event cameras.
We compare the accuracy and runtime performance of all loss functions on a publicly available dataset, and conclude that the variance, the gradient and the Laplacian magnitudes are among the best loss functions.
The applicability of the loss functions is shown on multiple tasks: 
rotational motion, depth and optical flow estimation.
The proposed focus loss functions allow to unlock the outstanding properties of event cameras.
\end{abstract}
\vspace{-1ex}

\vspace{-2ex}
\section{Introduction}
\label{sec:introduction}

Event cameras are bio-inspired sensors that work radically different from traditional cameras. Instead of capturing brightness images at a fixed rate, they measure brightness \emph{changes} asynchronously.
This results in a stream of \emph{events}, which encode the time, location and polarity (sign) of the brightness changes.
Event cameras, such as the Dynamic Vision Sensor (DVS)~\cite{Lichtsteiner08ssc} posses outstanding properties compared to traditional cameras: 
very high dynamic range (\SI{140}{\decibel} vs. \SI{60}{\decibel}), high temporal resolution (in the order of \si{\micro\second}), and do not suffer from motion blur.
Hence, event cameras have a large potential to tackle challenging scenarios for standard cameras (such as high speed and high dynamic range) 
in tracking~\cite{Ni15neco,Gehrig18eccv,Mueggler14iros,Censi14icra,Gallego17pami,Tedaldi16ebccsp,Glover17iros,Alzugaray18ral},
depth estimation~\cite{Rogister12tnnls,Schraml15cvpr,Matsuda15iccp,Zou16icip,Osswald17srep,Rebecq17ijcv,Xie17fns,Andreopoulos18cvpr,Zhou18eccv,Zhu18eccv},
Simultaneous Localization and Mapping~\cite{Weikersdorfer13icvs,Kim14bmvc,Kim16eccv,Rebecq17ral,Reinbacher17iccp,Zhu17cvpr,Rosinol18ral,Mueggler18tro},
and recognition~\cite{Lee14tnnls,Orchard15pami,Lagorce16pami,Amir17cvpr,Sironi18cvpr}, among other applications.
However, novel methods are required to process the unconventional output of these sensors in order to unlock their potential.

Motion compensation approaches \cite{Gallego17ral,Rebecq17ijcv,Zhu17icra,Gallego18cvpr,Stoffregen17acra,Mitrokhin18iros,Zhu18eccv,Ye18arxiv,Zhu18arxiv,Stoffregen19arxiv,Mitrokhin19arxiv,Stoffregen19cvpr}
have been recently introduced for processing the visual information acquired by event cameras.
They have proven successful for the estimation of motion (optical flow)~\cite{Zhu17icra,Gallego18cvpr,Stoffregen17acra,Ye18arxiv,Zhu18arxiv}, camera motion~\cite{Gallego18cvpr,Gallego17ral}, depth (3D reconstruction)~\cite{Rebecq17ijcv,Zhu18eccv,Ye18arxiv,Zhu18arxiv} as well as segmentation~\cite{Stoffregen17acra,Stoffregen19arxiv,Mitrokhin19arxiv}.
The main idea of such methods consists of searching for point trajectories on the image plane that maximize \emph{event alignment}~\cite{Gallego17ral,Gallego18cvpr} (Fig.~\ref{fig:blockDiagramSimple}, right),
which is measured using some %
loss function of the events warped according to such trajectories.
The best trajectories produce sharp, motion compensated images that reveal the brightness patterns causing the events (Fig.~\ref{fig:blockDiagramSimple}, middle).

\begin{table}
\centering
\begin{adjustbox}{max width=\columnwidth}
\begin{tabular}{l|c|c|c}
\toprule
\textbf{Focus Loss Function} & \textbf{Type} & \textbf{Spatial?} & \textbf{Goal}\\
\midrule
Variance \eqref{eq:IWEVariance} \cite{Gallego18cvpr,Gallego17ral} & Statistical & No & max\\
Mean Square \eqref{eq:MeanSquareIWE} \cite{Gallego17ral,Stoffregen17acra} & Statistical & No & max\\
Mean Absolute Deviation \eqref{eq:MeanAbsoluteDevIWE} & Statistical & No & max\\
Mean Absolute Value \eqref{eq:MeanAbsoluteValIWE} & Statistical & No & max\\
Entropy \eqref{eq:maxEntropy} & Statistical & No & max\\
Image Area \eqref{eq:minSupp} %
& Statistical & No & min\\
Image Range \eqref{eq:SupportPDFSimplified} & Statistical & No & max\\
Local Variance~\eqref{eq:local_variance_aggr} & Statistical & Yes & max\\
Local Mean Square & Statistical & Yes & max\\
Local Mean Absolute Dev. & Statistical & Yes & max\\
Local Mean Absolute Value & Statistical & Yes & max\\
Moran's Index~\eqref{eq:DefMoranIndex} & Statistical & Yes & min\\
Geary's Contiguity Ratio~\eqref{eq:DefGearyC} & Statistical & Yes & max\\
Gradient Magnitude \eqref{eq:maxGradient} & Derivative & Yes & max\\
Laplacian Magnitude \eqref{eq:maxLaplacian} & Derivative & Yes & max\\
Hessian Magnitude \eqref{eq:maxHessian} & Derivative & Yes & max\\
Difference of Gaussians & Derivative & Yes & max\\
Laplacian of the Gaussian & Derivative & Yes & max\\
Variance of Laplacian & Stat. \& Deriv. & Yes & max\\
Variance of Gradient & Stat. \& Deriv. & Yes & max\\
Variance of Squared Gradient & Stat. \& Deriv. & Yes & max\\
Mean Timestamp on Pixel~\cite{Mitrokhin18iros} & Statistical & No & min\\
\bottomrule
\end{tabular}
\end{adjustbox}
\vspace{-1ex}
\caption{\label{tab:focusmeasures}List of objective functions considered.}
\vspace{-2ex}
\end{table}

In this work, we build upon the motion compensation framework~\cite{Gallego18cvpr} and extend it to include twenty more loss functions %
for applications such as ego-motion, depth and optical flow estimation.
We ask the question: \emph{What are good metrics of event alignment?} 
In answering, we noticed strong connections between the proposed metrics (Table~\ref{tab:focusmeasures}) and those used for shape-from-focus and autofocus in conventional, frame-based cameras~\cite{Sakurikar17iccv,Pertuz13jpr}, and so, 
we called the event alignment metrics ``\emph{focus loss functions}''. 
The extended framework allows mature computer vision tools to be used on event data while taking into account all the information of the events (asynchronous timestamps and polarity).
Additionally, it sheds light on the event-alignment goal of functions used in existing motion-compensation works and provides a taxonomy of loss functions for event data.

\vspace{-2.4ex}
\paragraph*{Contributions.}
In summary, our contributions are:
\begin{enumerate}[noitemsep,nolistsep]
\item The introduction and comparison of twenty two focus loss functions for event-based processing, many of which are developed from basic principles, 
such as the ``area'' of the image of warped events.
\item Connecting the topics of shape-from-focus, autofocus and event-processing by the similar set of functions used, thus allowing to bring mature analysis tools from the former topics into the realm of event cameras.
\item A thorough evaluation on a recent dataset~\cite{Mueggler17ijrr}, comparing the accuracy and computational effort of the proposed focus loss functions,
and showing how they can be used for depth and optical flow estimation.
\end{enumerate}

The rest of the paper is organized as follows.
Section~\ref{sec:DVS_operation_application} reviews the working principle of event cameras.
Section~\ref{sec:method} summarizes the motion compensation method %
and extends it with the proposed focus loss functions. 
Experiments are carried out in Section~\ref{sec:experiments} comparing the loss functions, and conclusions are drawn in Section~\ref{sec:conclusion}.

\section{Event-based Camera Working Principle}
\label{sec:DVS_operation_application}

Event-based cameras, such as the DVS~\cite{Lichtsteiner08ssc}, have independent pixels that output ``events'' in response to \emph{brightness changes}. 
Specifically, if $\Lum(\bx,t)\doteq \log I(\bx,t)$ is the logarithmic brightness at pixel $\bx \doteq (x,y)^\top$ on the image plane, the DVS generates an event $e_k \doteq (\bx_k,t_k,\pol_k)$ 
if the change in logarithmic brightness at pixel $\bx_k$ reaches a threshold $C$ (e.g., 10-15\% relative change):
\begin{equation}
\label{eq:generativeEventCondition}
\Delta \Lum \doteq \Lum(\bx_k,t_k) - \Lum(\bx_k,t_k-\Delta t_k) = \pol_{k}\, C,
\end{equation}
where $t_k$ is the timestamp of the event, 
$\Delta t_k$ is the time since the previous event at the same pixel $\bx_k$ 
and $\pol_k \in \{+1,-1\}$ is the event polarity (i.e., sign of the brightness change).

Therefore, each pixel has its own sampling rate (which depends on the visual input) 
and outputs data proportionally to the amount of motion in the scene. 
An event camera does not produce images at a constant rate, but rather a stream of \emph{asynchronous}, sparse events in space-time (Fig.~\ref{fig:blockDiagramSimple}, left).

\section{Methodology}
\label{sec:method}
\subsection{Motion Compensation Framework}
\label{sec:contrastmax:review}
In short, the method in~\cite{Gallego18cvpr} seeks to find the point-trajectories on the image plane that maximize the alignment of corresponding events (i.e., those triggered by the same scene edge).
Event alignment is measured by the \emph{strength of the edges} of an image of warped events (IWE), which is obtained by aggregating events along candidate point trajectories (Fig.~\ref{fig:blockDiagramSimple}).
In particular, \cite{Gallego18cvpr} proposes to measure edge strength (which is directly related to \emph{image contrast}~\cite{Gonzalez09book}) using the variance of the IWE.

More specifically, the events in a set $\cE = \{e_k\}_{k=1}^{\numEvents}$ are geometrically transformed 
\vspace{-1ex}
\begin{equation}
e_k \doteq (\bx_k,t_k,\pol_k) \;\,\mapsto\;\, 
e'_k \doteq (\bx'_k,\tref,\pol_k)
\vspace{-0.8ex}
\end{equation}
according to a point-trajectory model~$\Warp$, resulting in a set of warped events $\cE' = \{e'_k\}_{k=1}^{\numEvents}$ ``flattened'' at a reference time $\tref$.
The warp $\bx'_k = \Warp(\bx_k,t_k; \bparams)$ transports each event along the point trajectory that passes through it (Fig.~\ref{fig:blockDiagramSimple}, left), until $\tref$ is reached, thus, taking into account the space-time coordinates of the event.
The vector $\bparams$ parametrizes the point trajectories, and hence contains the motion or scene parameters.

\begin{figure*}
\centering
\global\long\def\heightevol{2.45cm}
\subfloat[\label{fig:evol:dyn:wo:blur1}Iteration $it= 1$]{\frame{\includegraphics[trim={60bp 70bp 40bp 10bp},clip,height=\heightevol]{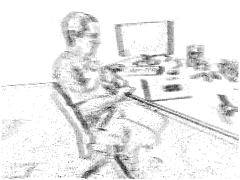}}}~
\subfloat[\label{fig:evol:dyn:wo:blur2}$it= 2$]{\frame{\includegraphics[trim={60bp 70bp 40bp 10bp},clip,height=\heightevol]{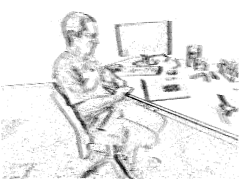}}}~
\subfloat[\label{fig:evol:dyn:wo:blur3}$it= 3$]{\frame{\includegraphics[trim={60bp 70bp 40bp 10bp},clip,height=\heightevol]{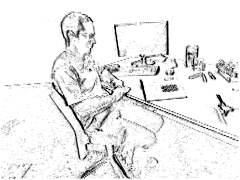}}}~
\subfloat[\label{fig:evol:dyn:wo:blur4}$it= 5$]{\frame{\includegraphics[trim={60bp 70bp 40bp 10bp},clip,height=\heightevol]{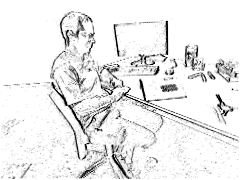}}}~
\subfloat[\label{fig:evol:dyn:wo:hist}Histograms of (a)-(d).]{\includegraphics[height=\heightevol]{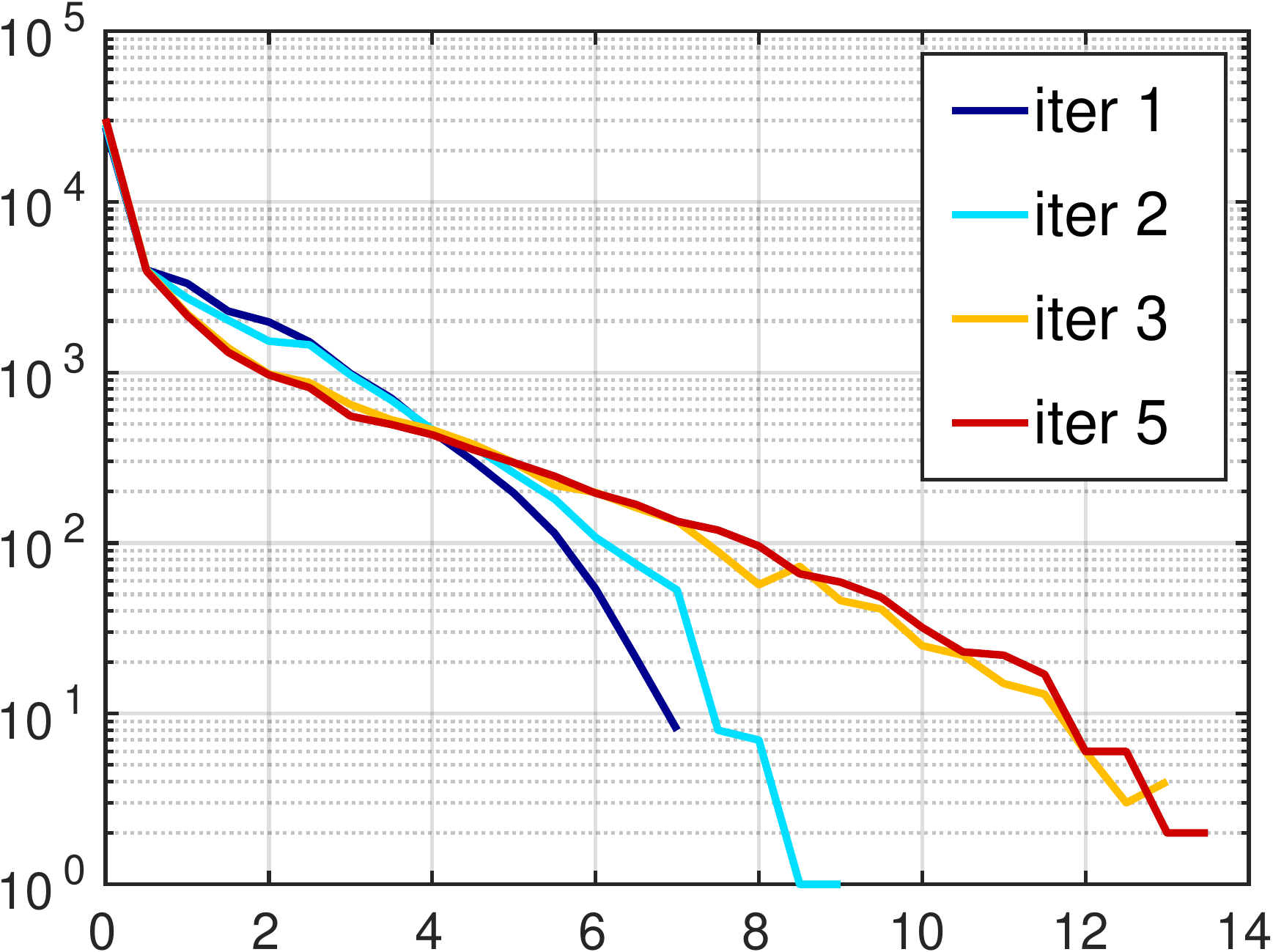}}\\[-1.5ex]
\subfloat[\label{fig:evol:dyn:w:blur1}Iteration $it= 1$]{\frame{\includegraphics[trim={60bp 70bp 40bp 10bp},clip,height=\heightevol]{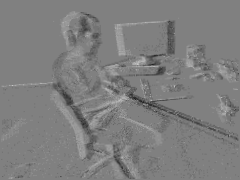}}}~
\subfloat[\label{fig:evol:dyn:w:blur2}$it= 2$]{\frame{\includegraphics[trim={60bp 70bp 40bp 10bp},clip,height=\heightevol]{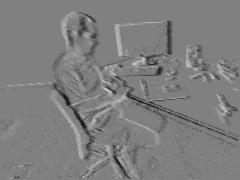}}}~
\subfloat[\label{fig:evol:dyn:w:blur3}$it= 3$]{\frame{\includegraphics[trim={60bp 70bp 40bp 10bp},clip,height=\heightevol]{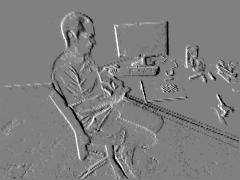}}}~
\subfloat[\label{fig:evol:dyn:w:blur4}$it= 5$]{\frame{\includegraphics[trim={60bp 70bp 40bp 10bp},clip,height=\heightevol]{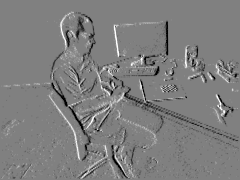}}}~
\subfloat[\label{fig:evol:dyn:w:hist}Histograms of (f)-(i).]{\includegraphics[height=\heightevol]{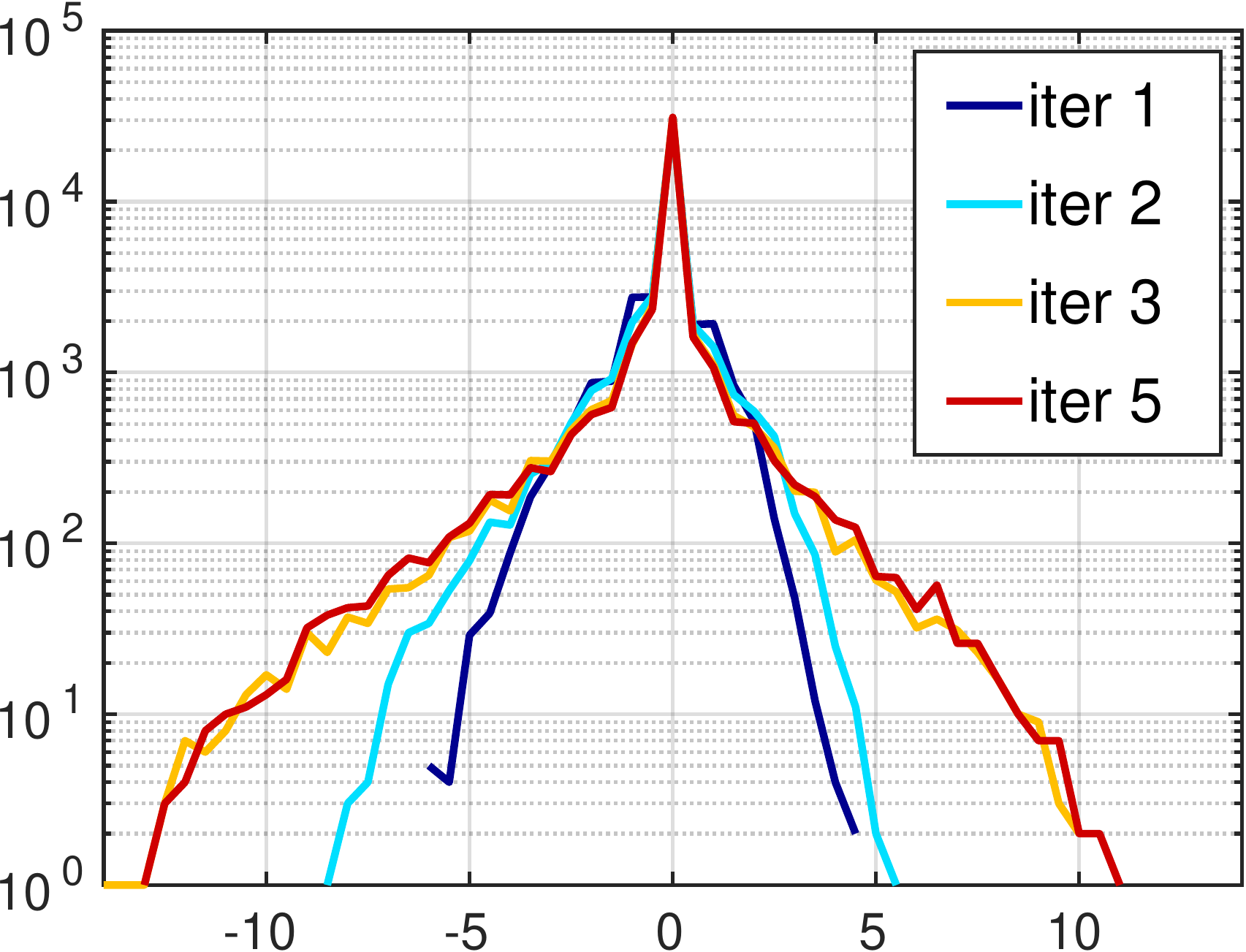}}
\vspace{-1ex}
\caption{\label{fig:evol:dyn}
Evolution of the image of warped events (IWE, \eqref{eq:IWE}) as the focus loss is optimized, showing how the IWE sharpens as the motion parameters $\bparams$ are estimated.
Motion blur (left) due to event misalignment (in the example, dominantly in horizontal direction) decreases as warped events become better aligned (right).
Top row: without polarity ($b_k=1$ in~\eqref{eq:IWE}); 
Bottom row: with polarity ($b_k=\pol_k$ in~\eqref{eq:IWE}). 
The last column shows the histograms of the images;
the peak at zero corresponds to the pixels with no events (white in the top row, gray in the bottom row).
}
\vspace{-1ex}\hrulefill\vspace{-1ex}
\end{figure*}
The image (or histogram) of warped events (IWE) is given by accumulating events along the point trajectories:
\vspace{-1ex}
\begin{equation}
\label{eq:IWE}
\IWE(\bx;\bparams) \doteq \sum_{k=1}^{\numEvents} b_k \,\delta (\bx - \bx'_k(\bparams)),
\vspace{-0.5ex}
\end{equation}
where each pixel $\bx$ sums the values $b_k$ of the warped events $\bx'_k$ that fall within it ($b_k=\pol_k$ if polarity is used or $b_k=1$ if polarity is not used; see Fig.~\ref{fig:evol:dyn}).
In practice, the Dirac delta $\delta$ is replaced by a smooth approximation, such as a Gaussian, $\delta(\bx-\bmu)\approx\cN(\bx;\bmu,\epsilon^2\mId)$, with typically $\epsilon=1$~pixel.

The contrast~\cite{Gallego18cvpr} of the IWE~\eqref{eq:IWE} is given by its variance:
\vspace{-0.5ex}
\begin{equation}
\label{eq:IWEVariance}
\variance\bigl(\IWE(\bx;\bparams)\bigr) 
\doteq \frac{1}{|\Omega|} \int_{\Omega} (\IWE(\bx;\bparams)-\mu_{\IWE})^2 d\bx
\vspace{-0.5ex}
\end{equation}
with mean $\mu_{\IWE} \doteq \frac{1}{|\Omega|} \int_{\Omega} \IWE(\bx;\bparams) d\bx$.
Discretizing into pixels, it becomes $\label{eq:IWEVarianceDiscrete}
\variance(\IWE) = \frac1{\numPixels} \sum_{i,j} (\IWEentry_{ij}-\mu_{\IWE})^2,
$
where $\numPixels$ is the number of pixels of $\IWE=(\IWEentry_{ij})$ and $\mu_{\IWE} = \frac1{\numPixels}\sum_{i,j}\IWEentry_{i,j}$.

\subsection{Our Proposal: Focus Loss Functions}
\label{sec:Focus}

In~\eqref{eq:IWE} a group of events $\cE$ has been effectively converted, using the spatio-temporal transformation, into an image representation (see Fig.~\ref{fig:evol:dyn}),
aggregating the little information from individual events into a larger, more descriptive piece of information.
Here, we propose to exploit the advantages that such an image representation offers, namely bringing in tools from image processing (histograms, convolutions, Fourier transforms, etc.) to analyze event alignment.

We study different image-based event alignment metrics (i.e., loss functions), such as~\eqref{eq:IWEVariance}; 
that is, we study different objective functions that can be used in the second block of the diagram of Fig.~\ref{fig:blockDiagramSimple} (highlighted in red).
To the best of our knowledge we are the first to address the following three related topics:
(\emph{i}) establishing connections between event alignment metrics and so-called ``focus measures'' in shape-from-focus (SFF) and autofocus (AF) in conventional frame-based imaging~\cite{Sakurikar17iccv,Pertuz13jpr},
(\emph{ii}) comparing the performance of the different metrics 
on event-based vision problems, 
and (\emph{iii}) showing that it is possible to design new focus metrics tailored to edge-like images like the IWE. %

Table~\ref{tab:focusmeasures} lists the multiple focus loss functions studied, the majority of which are newly proposed, 
and categorizes them according to their nature.
The functions are classified according to whether they are based on statistical or derivative operators (or their combination), 
according to whether they depend on the spatial arrangement of the pixel values or not,
and according to whether they are maximized or minimized.

The next sections present the focus loss functions using two image characteristics related to edge strength: \emph{sharpness} (Section~\ref{sec:Focus:Sharpness}) and \emph{dispersion} (Section~\ref{sec:Focus:Dispersion}).
But first, let us discuss the variance loss, since it motivates several focus losses.

\vspace{-2ex}
\paragraph{Loss Function: Image Variance (Contrast).}
\label{sec:Focus:Variance}

An event alignment metric used in~\cite{Gallego17ral,Gallego18cvpr} is the variance of the IWE~\eqref{eq:IWEVariance}, known as the RMS \emph{contrast} in image processing~\cite{Gonzalez09book}.
The variance is a statistical measure of \emph{dispersion}, in this case, of the pixel values of the IWE, regardless of their spatial arrangement.
Thus, event alignment (i.e., edge strength of the IWE) is here assessed using statistical principles (Table~\ref{tab:focusmeasures}).
The motion parameters $\bparams$ that best fit the events are obtained by maximizing the variance~\eqref{eq:IWEVariance}.
In the example of Fig.~\ref{fig:evol:dyn}, it is clear that as event alignment increases, so does the visual contrast of the IWE.

\emph{Fourier Interpretation:}
Using (\emph{i}) the formula relating the variance of a signal to its mean square (MS) and mean, $\variance(\IWE) = \text{MS} - \mu_{\IWE}^2$, 
and (\emph{ii}) the interpretations of MS and squared mean as the total ``energy'' and DC component of the signal, respectively,
yields that the variance represents the AC component of the signal, i.e., the energy of the oscillating, high frequency, content of the IWE.
Hence, the IWE comes into focus (Fig.~\ref{fig:evol:dyn}) by increasing its AC component; the DC component does not change significantly during the process.
This interpretation is further developed next.

\subsection{Image Sharpness}
\label{sec:Focus:Sharpness}

In the Fourier domain, sharp images (i.e., those with strong edges, e.g., Fig.~\ref{fig:evol:dyn}, right) have a significant amount of energy concentrated at high frequencies. 
The high-frequency content of an image can be assessed by measuring the magnitude of its derivative 
since derivative operators act as band-pass or high-pass filters. 
The magnitude is given by any norm; however, the $L^2$ norm is preferred since it admits an inner product interpretation and simple derivatives.
The following losses are built upon this idea, using first and second derivatives, respectively.
Similar losses, based on the DCT or wavelet transforms~\cite{Pertuz13jpr} instead of the Fourier transform are also possible.
The main idea remains the same: measure the high frequency content of the IWE (i.e., edge strength) and find the motion parameters that maximize it, therefore maximizing event alignment.

\vspace{-1ex}
\paragraph{Loss Function: Magnitude of Image Gradient.}
\label{sec:Focus:Gradient}
Event alignment is achieved by seeking the parameters $\bparams$ of the point-trajectories that maximize
\vspace{-0.5ex}
\begin{equation}
\label{eq:maxGradient}
\| \nabla I \|_{L^{2}(\Omega)}^{2}\doteq\int_{\Omega}\| \nabla I(\bx) \|^{2}d\bx,
\vspace{-0.5ex}
\end{equation}
where $\nabla I=(I_{x},I_{y})^{\top}$ is the gradient of the IWE $I$ (subscripts indicate derivative: $I_{x} \equiv \partial I/\partial x$),
and its magnitude is measured by an $L^{p}$ norm (\revred{suppl. material}), e.g., the (squared) $L^2$ norm: $\|\nabla I\|^2_{L^2(\Omega)} = \int_\Omega (I^2_x(\bx) + I^2_y(\bx)) d\bx$.

\vspace{-1ex}
\paragraph{Loss Function: Magnitude of Image Hessian.}
\label{sec:Focus:Hessian}
For these loss functions, event alignment is attained by maximizing the magnitude of the second derivatives (i.e., Hessian) of the IWE, $\hessian(I)$.
We use the (squared) norm of the Laplacian, %
\vspace{-0.5ex}
\begin{equation}
\label{eq:maxLaplacian}
\| \Delta I\|_{L^{2}(\Omega)}^{2} \doteq \|I_{xx}+ I_{yy}\|_{L^{2}(\Omega)}^{2},
\vspace{-0.5ex}
\end{equation}
or the (squared) Frobenius norm of the Hessian,
\vspace{-0.5ex}
\begin{equation}
\label{eq:maxHessian}
\|\hessian(I)\|_{L^{2}(\Omega)}^{2} \doteq \|I_{xx}\|_{L^{2}(\Omega)}^{2}+\|I_{yy}\|_{L^{2}(\Omega)}^{2}+2\|I_{xy}\|_{L^{2}(\Omega)}^{2}.
\vspace{-2ex}
\end{equation}

\paragraph{Loss Functions: DoG and LoG.}
The magnitude of the output of established band-pass filters, such as the Difference of Gaussians (DoG) and the Laplacian of the Gaussian (LoG), can also be used to assess the sharpness of the IWE.

\begin{figure}
\centering
\subfloat[\label{fig:supp:illustr:IWE}IWE~\eqref{eq:IWE} $\IWE(\bx;\bparams)$ with $b_k=1$.]{\frame{\includegraphics[height=2.8cm]{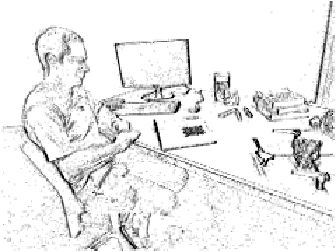}}}~~
\subfloat[\label{fig:supp:illustr:suppMap}Support map $F(\IWE(\bx;\bparams))- F(0)$.]{\includegraphics[height=2.85cm]{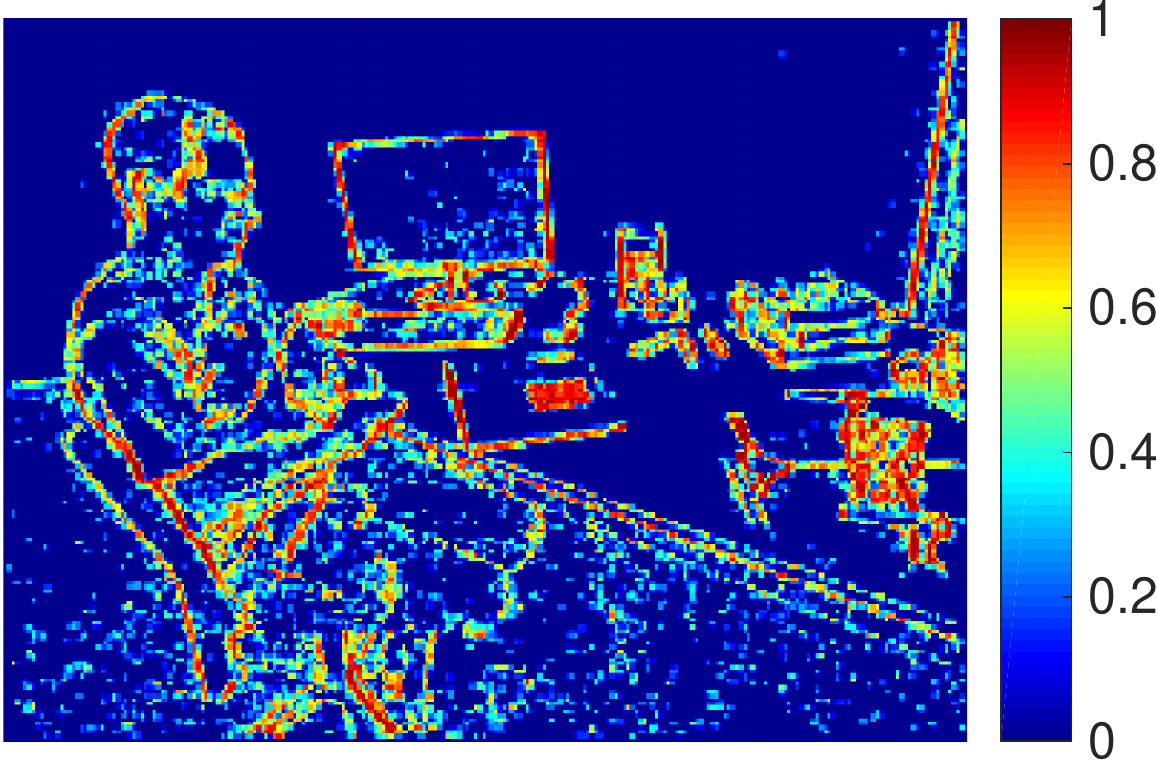}}
\vspace{-0.5ex}
\caption{\label{fig:supp:illustr}Support map of an %
IWE.
Red regions contribute more to the support than blue regions.
The integral of the support map in Fig.~\ref{fig:supp:illustr:suppMap} gives the support~\eqref{eq:minSupp} of the IWE.}
\vspace{-1ex}\hrulefill\vspace{-1ex}
\end{figure}

\vspace{-1ex}
\paragraph{Loss Function: Image Area.}
\label{sec:Focus:MinSupport}
Intuitively, sharp, motion-compensated IWEs, have \emph{thinner edges} than uncompensated ones (Fig.~\ref{fig:evol:dyn}).
We now provide a definition of the ``thickness'' or ``support'' of edge-like images (see \revred{supplementary material}) %
and use it for event alignment.
We propose to minimize the support (i.e., area) of the IWE~\eqref{eq:IWE}:
\begin{equation}
\label{eq:minSupp}
\supp(\IWE(\bx;\bparams)) = \int_{\Omega}\left( F(I(\bx;\bparams)) - F(0) \right)d\bx,
\end{equation}
where $F(\lambda)\doteq \int \rho(\lambda) d\lambda$ is the primitive of a decreasing weighting function $\rho(\lambda)\geq 0$, for $\lambda\ge 0$.
Fig.~\ref{fig:supp:illustr} shows an IWE and its corresponding support map, %
pseudo-colored as a heat map:
red regions (large IWE values) contribute more to the support~\eqref{eq:minSupp} than blue regions (small IWE values).

\begin{figure}
\centering
\subfloat[Weighting functions $\rho(\lambda)$.]{\includegraphics[width=0.48\columnwidth]{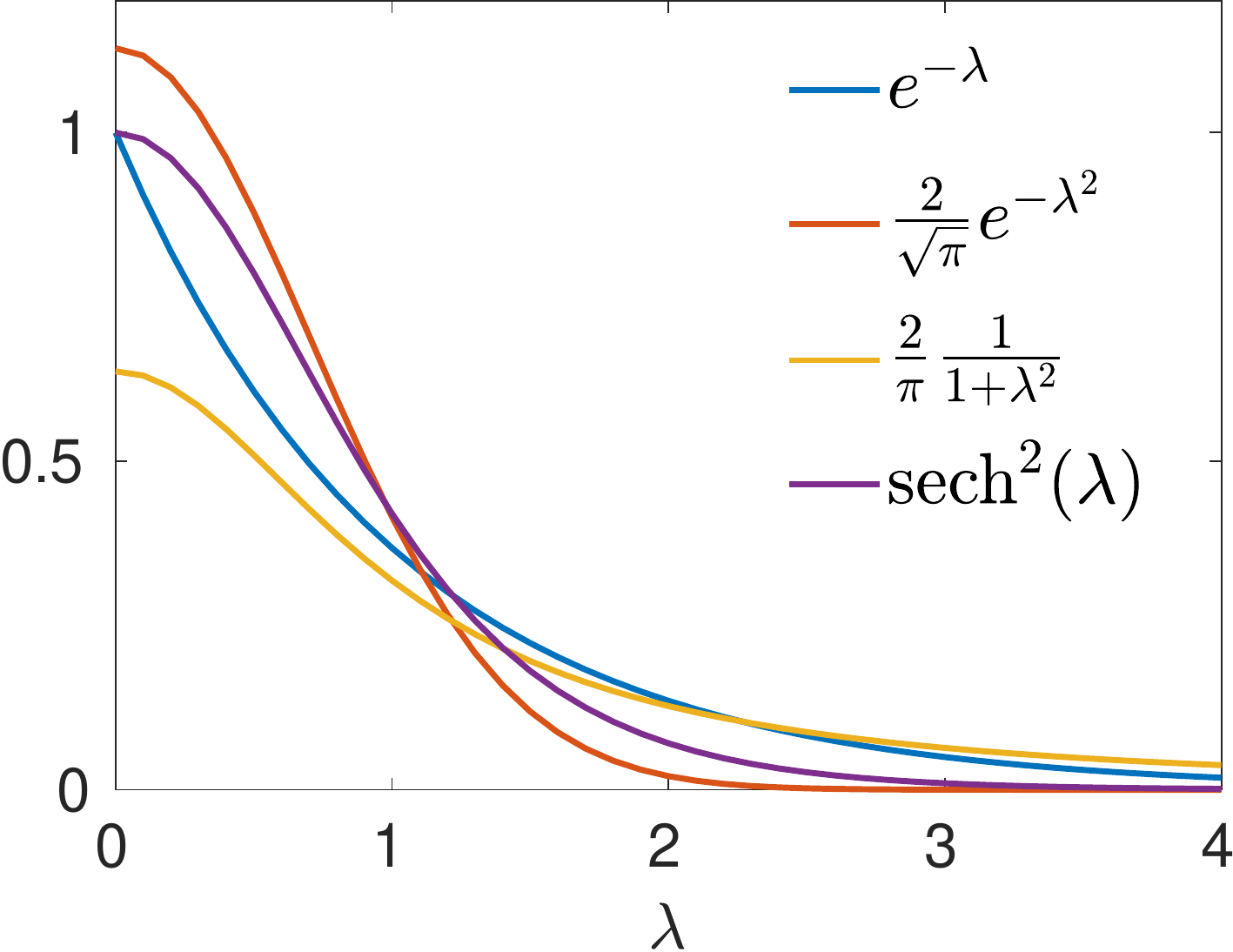}}~~
\subfloat[Primitive functions $F(\lambda)$.]{\includegraphics[width=0.48\columnwidth]{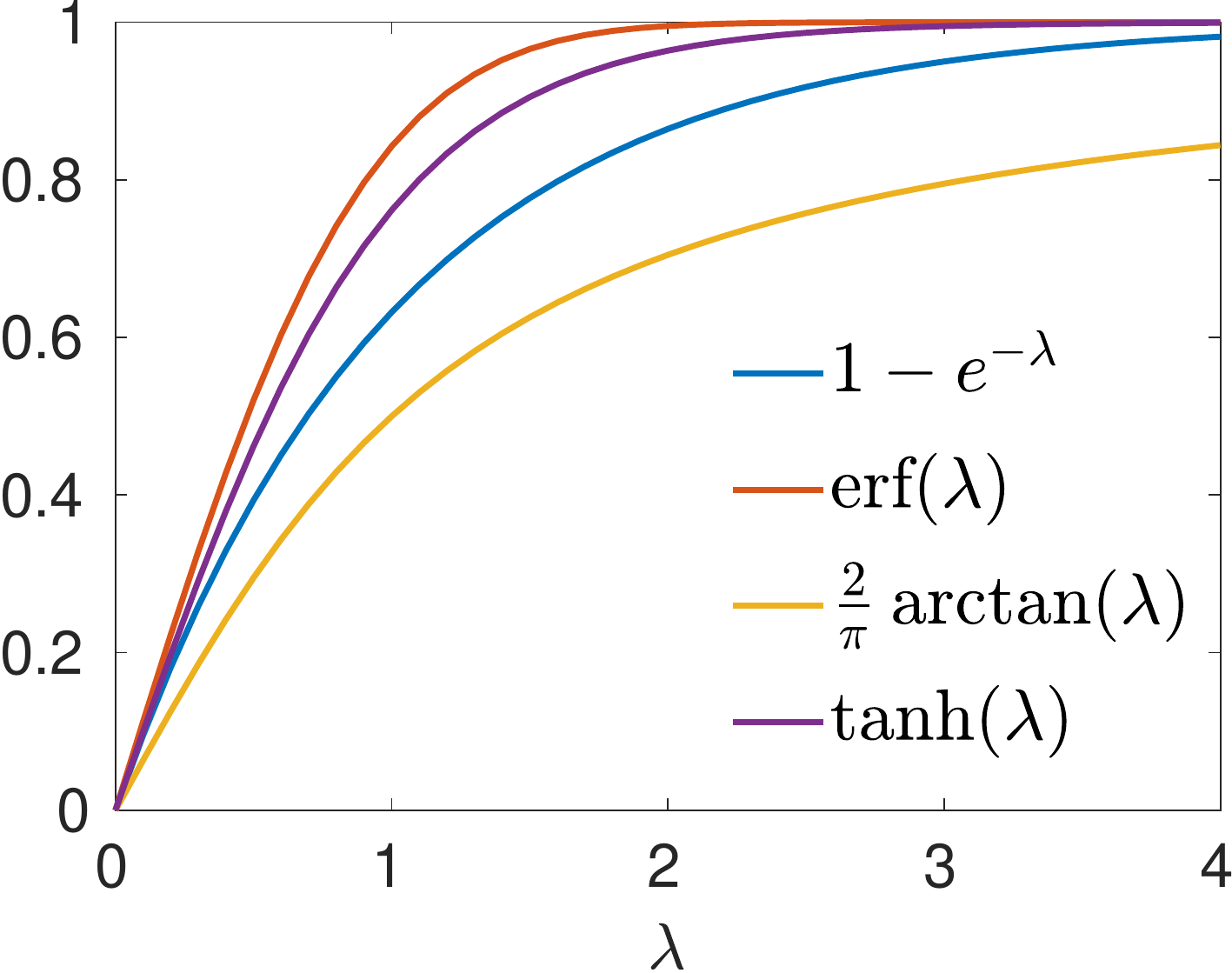}}
\vspace{-0.5ex}
\caption{\label{fig:supp-weights}Weighting functions $\rho(\lambda)$ and their primitives $F(\lambda)$, used to define the support~\eqref{eq:minSupp} of the IWE.}
\vspace{-1ex}\hrulefill\vspace{-1ex}
\end{figure}

Our definition of the support of the IWE is very flexible, since it allows for different choices of the weighting function (Fig.~\ref{fig:supp-weights}).
Specifically, we consider four choices:
\begin{enumerate}[noitemsep]
\item Exponential: $\rho(\lambda)=e^{-\lambda}$, $F(\lambda)=1 -e^{-\lambda}$.
\item Gaussian: $\rho(\lambda)=\frac{2}{\sqrt{\pi}} e^{-\lambda^{2}}$, $F(\lambda)=\text{erf}(\lambda)$.
\item Lorentzian: $\rho(\lambda)=\frac{2}{(1+\lambda^{2})\pi}$, $F(\lambda)=\frac{2}{\pi}\arctan(\lambda)$.
\item Hyperbolic: $\rho(\lambda)=\text{sech}^{2}(\lambda)$, $F(\lambda)=\tanh(\lambda)$.
\end{enumerate}

\subsection{Image Dispersion (Statistics)}
\label{sec:Focus:Dispersion}
Besides the variance~\eqref{eq:IWEVariance}, other ways to measure dispersion are possible.
These may be categorized according to their spatial character: global (operating on the IWE pixels regardless of their arrangement) or local.
Among the global metrics for dispersion we consider the mean square (MS), the mean absolute deviation (MAD), the mean absolute value (MAV), the entropy and the range.
In the group of local dispersion losses we have: 
local versions of the above-mentioned global losses (variance, MS, MAD, MAV, etc.) 
and metrics of spatial autocorrelation, such as Moran's I index \cite{Moran50biom} and Geary's Contiguity Ratio~\cite{Geary54inc}.

\vspace{-2ex}
\paragraph{Loss Function: Mean Square.}
The Mean Square (MS)
\vspace{-0.5ex}
\begin{equation}
\label{eq:MeanSquareIWE}
\text{MS}(\IWE) = \|\IWE(\bx;\bparams)\|^2_{L^2(\Omega)} \,/\, |\Omega|,
\vspace{-0.5ex}
\end{equation}
also measures the dispersion of the IWE (with respect to zero).
As anticipated in the Fourier interpretation of the variance, the MS is the total energy of the image, which comprises an oscillation (i.e., dispersion) part and a constant part (DC component or squared mean).

\vspace{-2ex}
\paragraph{Loss Function: Mean Absolute Deviation (MAD) and Mean Absolute Value (MAV).}
The analogous of the variance and the MS using the $L^1$ norm are the MAD,
\vspace{-0.5ex}
\begin{equation}
\label{eq:MeanAbsoluteDevIWE}
\text{MAD}(\IWE) = \|\IWE(\bx;\bparams) - \mu_{\IWE}\|_{L^1(\Omega)} \,/\, |\Omega|,
\vspace{-0.5ex}
\end{equation}
and the mean absolute value,
\vspace{-0.5ex}
\begin{equation}
\label{eq:MeanAbsoluteValIWE}
\text{MAV}(\IWE) = \|\IWE(\bx;\bparams)\|_{L^1(\Omega)} \,/\, |\Omega|,
\vspace{-0.5ex}
\end{equation}
respectively. 
The MAV provides a valid loss function to estimate $\bparams$ if polarity is used. 
However, if polarity is not used, the MAV coincides with the IWE mean, $\mu_I$ (which counts the warped events in $\Omega$), 
and since this value is typically constant, the MAV does not provide enough information to estimate $\bparams$ in this case (as will be noted in Table~\ref{tab:boxrot:poster:all}).

\vspace{-2ex}
\paragraph{Loss Function: Image Entropy.}
\label{sec:Focus:Entropy}

Information entropy measures the uncertainty or spread (i.e., dispersion) of a distribution, or, equivalently, the reciprocal of its concentration. %
This approach consists of maximizing Shannon's entropy of the random variable given by the IWE pixels:
\vspace{-0.5ex}
\begin{equation}
H\left(p_{\IWE}(z)\right)=-\int_{-\infty}^{\infty}p_{\IWE}(z)\log p_{\IWE}(z) dz
\label{eq:maxEntropy}
\vspace{-0.5ex}
\end{equation}

The PDF of the IWE is approximated by its histogram (normalized to unit area). 
Comparing the PDFs of the IWEs before and after motion compensation (last column of Fig.~\ref{fig:evol:dyn}), we observe two effects of event alignment:
($i$) the peak of the distribution at $\IWE=0$ increases since the image regions with almost no events grow (corresponding to homogeneous regions of the brightness signal), and 
($ii$) the distribution spreads out away from zero: 
larger values of $|\IWE(\bx)|$ are achieved.
Hence, the PDF of the motion-compensated IWE is more concentrated around zero and more spread out away from zero than the uncompensated one.
Concentrating the PDF means decreasing the entropy, whereas spreading it out means increasing the entropy. 
To obtain a high contrast IWE, with sharper edges, the second approach must dominate.
Hence, parameters $\bparams$ are obtained by maximizing the entropy of the IWE~\eqref{eq:maxEntropy}.

Entropy can also be interpreted as a measure of diversity in information content, and since 
($i$) sharp images contain more information than blurred ones~\cite{Firestone91cyto}, and 
($ii$) our goal is to have sharp images for better event alignment, thus our goal is to maximize the information content of the IWE, which is done by maximizing its entropy.

\vspace{-2ex}
\paragraph{Loss Function: Image Range.}
\label{sec:Focus:Range}
As is well known, contrast is a measure of the oscillation (i.e., dispersion) of a signal with respect to its background (e.g., Michelson contrast), 
and the range of a signal, $\range(I) = I_{\max} - I_{\min}$, measures its maximum oscillation.
Hence, maximizing the range of the IWE provides an alternative way to achieve event alignment.
However, the min and max statistics of an image are brittle, since they can drastically change by modifying two pixels.
Using the tools that lead to~\eqref{eq:minSupp} (\revred{suppl. material}), 
we propose to measure image range more sensibly by means of the support of the image PDF, 
\vspace{-1ex}
\begin{equation}
\supp(p_{I}) = \int_{-\infty}^{\infty}\left(F(p_{I}(z))-F(0)\right) dz,
\label{eq:SupportPDFSimplified}
\vspace{-1ex}
\end{equation}
where $F(\lambda)$ is a primitive of the weight function $\rho(\lambda) \ge 0$ (as in Fig.~\ref{fig:supp-weights}).
Fig.~\ref{fig:evol:dyn} illustrates how the range of the IWE is related to its sharpness: as the IWE comes into focus, the image range (support of the histogram) increases.

Next, we present focus loss functions based on local versions of the above global statistics.

\vspace{-2ex}
\paragraph{Loss Function: Local Variance, MS, MAD and MAV.}
Mimicking~\eqref{eq:maxGradient}, which aggregates local measures of image sharpness to produce a global score, 
we may aggregate the local variance, MS, MAD or MAV of the IWE to produce a global dispersion score that we seek to maximize.
For example, the aggregated local variance (ALV) of the IWE is
\vspace{-1ex}
\begin{equation}
\label{eq:local_variance_aggr}
\text{ALV}(\IWE) \doteq \int_{\Omega} \variance(\bx; \IWE) d\bx,
\vspace{-1ex}
\end{equation}
where the local variance in a neighborhood $B(\bx)\subset \Omega$ centered about the point $\bx$ is given by
\vspace{-1ex}
\begin{equation}
\label{eq:local_variance}
\variance(\bx; \IWE) \doteq \frac1{|B(\bx)|}\int_{B(\bx)} \bigl( \IWE(\bu;\bparams) - \mu(\bx;\IWE) \bigr)^2 d\bu,
\vspace{-1ex}
\end{equation} 
with local mean
$\mu(\bx;\IWE) \doteq \int_{B(\bx)} \IWE(\bv;\bparams) \,d\bv / |B(\bx)|$ 
and $|B(\bx)| = \int_{B(\bx)} d\bu$.
The local variance of an image~\eqref{eq:local_variance} is an edge detector similar to the magnitude of the image gradient, $\|\nabla \IWE\|$.
It may be estimated using a weighted neighborhood $B(\bx)$ by means of convolutions with a smoothing kernel $G_\sigma(\bx)$, such as a Gaussian: %
\vspace{-1ex}
\begin{equation}
\label{eq:local_variance_conv}
\variance(\bx;\IWE) \approx (\IWE^2(\bx) * G_{\sigma}(\bx)) - \left(\IWE(\bx) * G_{\sigma}(\bx)\right)^2.
\vspace{-1ex}
\end{equation}

Based on the above example, local versions of the MS, MAD and MAV can be derived (see \revred{suppl. material}).

\vspace{-2ex}
\paragraph{Loss Function: Spatial Autocorrelation.}
\label{sec:Focus:Moran}
Spatial autocorrelation of the IWE can also be used to assess event alignment.
We present two focus loss functions based on spatial autocorrelation: Moran's \emph{I} and Geary's $C$ indices.

\emph{Moran's \emph{I} index} is a number between $-1$ and $1$ that evaluates whether the variable under study (i.e., the pixels of the IWE) is clustered, dispersed, or random.
Letting $z_i = \IWE(\bx_i)$ be the value of the $i$-th pixel of the IWE,
Moran's index is 
\vspace{-1.2ex}
\begin{equation}
\label{eq:DefMoranIndex}
\moranIndex(\IWE) \doteq
\frac{\sum_{i,j} w_{ij}(z_{i}-{\bar{z}})(z_{j}-{\bar{z}})/W}
{\sum_{i}(z_{i}-{\bar{z}})^{2}/\numPixels},
\vspace{-0.5ex}
\end{equation}
where $\bar{z}\equiv \mu_{\IWE}$ is the mean of~$\IWE$, 
$\,\numPixels$ is the number of pixels of~$\IWE$, 
$\,(w_{ij})$ is a matrix of spatial weights with zeros on the diagonal (i.e., $w_{ii}=0$) and $W=\sum_{i,j}w_{ij}$. 
The weights $w_{ij}$ encode the spatial relation between pixels $z_i$ and $z_j$:
we use weights that decrease with the distance between pixels, e.g., $w_{ij}\propto \exp(-\|\bx_i-\bx_j\|^2 / 2\sigma_M^2)$, with $\sigma_M \approx 1$ pixel.

A positive Moran's index value indicates tendency toward clustering pixel values while a negative Moran's index value indicates tendency toward dispersion (dissimilar values are next to each other).
Edges correspond to dissimilar values close to each other, and so, we seek to minimize Moran's index of the IWE.

\emph{Geary's Contiguity Ratio} is a generalization of Von Neumann's ratio~\cite{vonNeumann41ams} of the mean square successive difference to the variance: %
\vspace{-0.5ex}
\begin{equation}
\label{eq:DefGearyC}
C(\IWE)\doteq\frac{1}{2}\frac{\sum_{i,j}w_{ij}(z_{i}-z_{j})^{2}/W}{\sum_{i}(z_{i}-\bar{z})^{2}/(\numPixels-1)}.
\vspace{-0.5ex}
\end{equation}
It is non-negative. Values around 1 indicate lack of spatial autocorrelation; values near zero are positively correlated, and values larger than 1 are negatively correlated.
Moran's \emph{I} and Geary's $C$ are inversely related, thus event alignment is achieved by maximizing Geary's $C$ of the IWE.

\subsection{Dispersion of Image Sharpness Values}
\label{sec:Focus:CompositeLosses}
We have also combined statistics-based loss functions with derivative-based ones, yielding more loss functions (Table~\ref{tab:focusmeasures}), such as the variance of the Laplacian of the IWE $\variance(\Delta \IWE(\bx;\bparams))$,
the variance of the magnitude of the IWE gradient $\variance(\|\nabla \IWE(\bx;\bparams)\|)$,
and the variance of the squared magnitude of the IWE gradient $\variance(\|\nabla \IWE(\bx;\bparams)\|^2)\equiv \variance(\IWE^2_x + \IWE^2_y)$.
They apply statistical principles to local focus metrics based on neighborhood operations (convolutions).

\subsection{Discussion of the Focus Loss Functions}
\label{sec:Focus:ProsAndCons}

\emph{Connection with Shape-from-Focus and Autofocus:}
Several of the proposed loss functions have been proven successful in shape-from-focus (SFF) and autofocus (AF) with conventional cameras~\cite{Sakurikar17iccv,Pertuz13jpr,Mir14spie}, showing that there is a strong connection between these topics and event-based motion estimation.
The principle to solve the frame-based and event-based problems is the same: maximize a focus score of the considered image or histogram.
In the case of conventional cameras, SFF and AF maximize edge strength at each pixel of a focal stack of images (in order to infer depth).
In the case of event cameras, the IWE (an image representation of the events) plays the role of the images in the focal stack: 
varying the parameters $\bparams$ produces a different ``slice of the focal stack'', which may be used not only to estimate depth, but also to estimate other types of parameters $\bparams$, such as optical flow, camera velocities, etc.%

\emph{Spatial Dependency:}
Focus loss functions based on derivatives or on local statistics imply neighborhood operations (e.g., convolutions with nearby pixels of the IWE), thus they depend on the spatial arrangement of the IWE pixels (Table~\ref{tab:focusmeasures}). 
Instead, global statistics (e.g., variance, entropy, etc.) do not directly depend on such spatial arrangement\footnote{
The IWE consists of warped events, which depend on the location of the events; however, by ``directly'' we mean that the focus loss function, by itself, does not depend on the spatial arrangement of the IWE values.}.
The image area loss functions~\eqref{eq:minSupp} are integrals of point-wise functions of the IWE, and so, they do not depend on the spatial arrangement of the IWE pixels. 
The PDF of the IWE also does not have spatial dependency, nor do related losses, such as entropy~\eqref{eq:maxEntropy} or range~\eqref{eq:SupportPDFSimplified}.
Composite focus losses (Section~\ref{sec:Focus:CompositeLosses}), however, have spatial dependency since they are computed from image derivatives.

\emph{Fourier Interpretation:}
Focus losses based on derivatives admit an intuitive interpretation in the Fourier domain: they measure the energy content in the high frequencies (i.e., edges) of the IWE (Section~\ref{sec:Focus:Sharpness}). 
Some of the statistical focus losses also admit a frequency interpretation.
For example, image variance quantifies the energy of the AC portion (i.e., oscillation) of the IWE, and the MS measures the energy of both, the AC and DC components.
Other focus functions, such as entropy, do not admit such a straightforward Fourier interpretation related to edge strength.

\section{Experiments}
\label{sec:experiments}

In this section we compare the performance of the proposed focus loss functions, which depends on multiple factors, such as, the task for which it is used, the number of events processed, the data used, and, certainly, implementation.
These factors produce a test space with an intractable combinatorial size, 
and so, we choose the task for best assessing accuracy on real data and then provide qualitative results on other tasks: depth and optical flow estimation.

\subsection{Accuracy and Timing Evaluation}

Rotational camera motion estimation provides a good scenario for assessing accuracy
since camera motion can be reliably obtained with precise motion capture systems.
The acquisition of accurate per-event optical flow or depth is less reliable, since it depends on additional depth sensors, such as RGB-D or LiDAR, 
which are prone to noise.

\begin{figure}[t]
\vspace{-1.5ex}
\centering
  \subfloat[\label{fig:rot:boxplot:without}Without event polarity.]{\includegraphics[width=0.46\columnwidth]{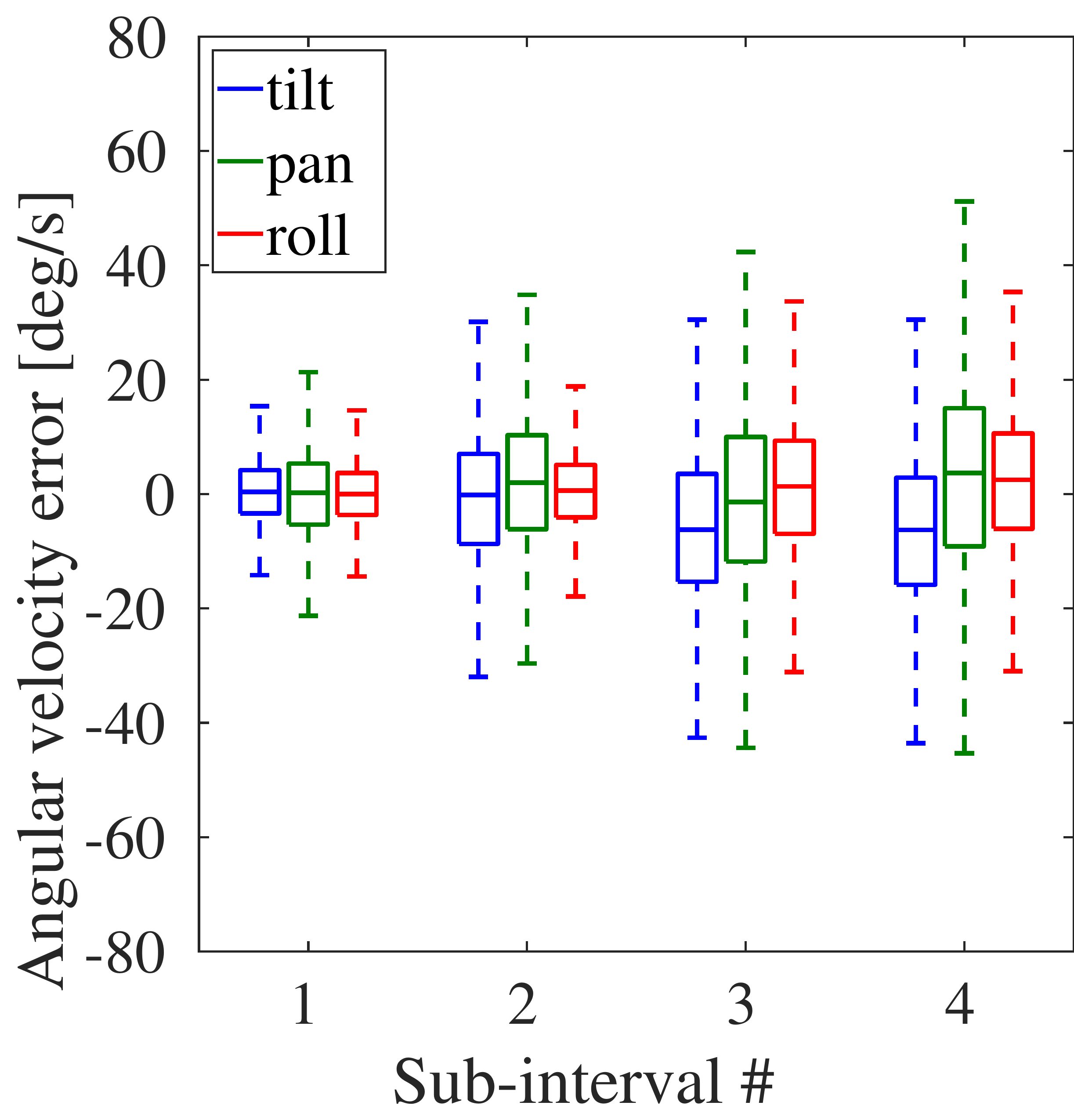}}~~
  \subfloat[\label{fig:rot:boxplot:with}With event polarity.]{\includegraphics[width=0.46\columnwidth]{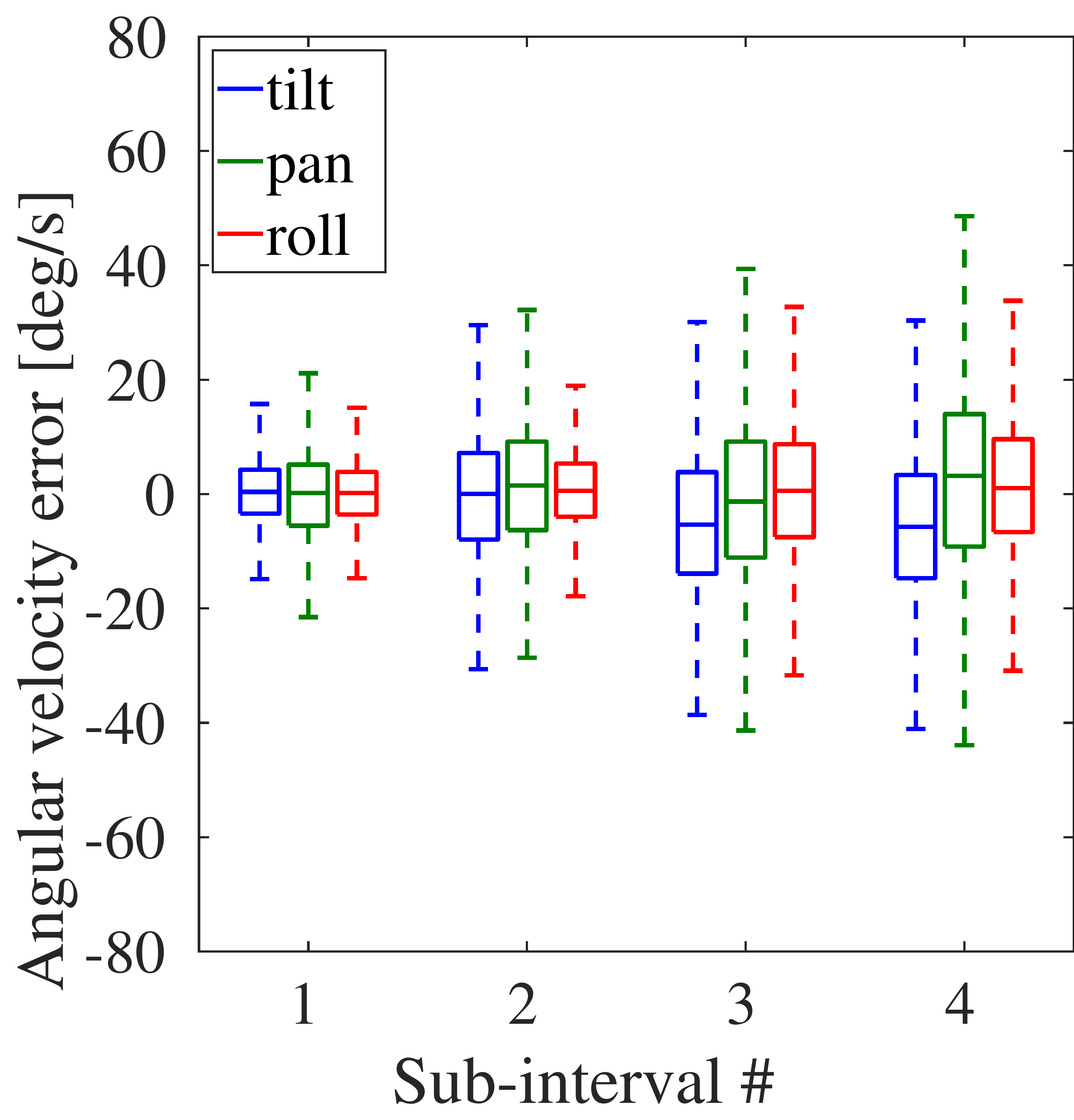}}
  \vspace{-1ex}
  \caption{\label{fig:rot:boxplot}
  Box plots of angular velocity error (estimated vs. ground truth), per axis, for four subintervals of \SI{15}{\second}, using the gradient magnitude~\eqref{eq:maxGradient} as focus loss function. 
  The method produces small errors: $\approx$\SI{18}{\degree/\second} RMS (see Table~\ref{tab:boxrot:poster:all}), i.e., \SI{2.7}{\percent} with respect to $|\omega_{\max}|=\SI{670}{\degree/\second}$.
  }
\vspace{-1ex}\hrulefill\vspace{-2ex}
\end{figure}

Table~\ref{tab:boxrot:poster:all} compares the accuracy and timing of the focus loss functions on the above scenario, using data from~\cite{Mueggler17ijrr}. %
The data consists of sequences of one minute duration, with increasing camera motion, reaching up to $\approx\pm$\SI{1000}{\degree/\second}, that is recorded using a motion-capture system with sub-millimeter precision at \SI{200}{\hertz}.
Each entry in Table~\ref{tab:boxrot:poster:all} consists of an experiment with more than 160 million events, processed in groups of $\numEvents = \si{30000}$ events.
The focus loss functions are optimized using non-linear conjugate gradient, 
initialized by the estimated angular velocity $\bparams$ for the previous group of events.
As the table reports, the angular velocity errors (difference between the estimated angular velocity $\bparams$ that optimizes the focus loss function and the velocity provided by the motion-capture system) vary across loss functions.
These columns summarize the RMS errors (with and without using polarity), which are reported in more detail in the supplementary material.
Among the most accurate focus loss functions are the ones based on the derivatives of the IWE, high-pass filters and the ones based on the variance (global or local), with less than \SI{2.6}{\percent} error.
The entropy and support-based losses (area and range) are not as accurate, yet the errors are small compared to the angular velocity excursions ($\omega_{\max}$) in the sequences (less than \SI{7}{\percent} error).
In general, Table~\ref{tab:boxrot:poster:all} shows that using polarity is beneficial; 
the results are slightly more accurate using polarity than not using it.
The error boxplots of the gradient magnitude~\eqref{eq:maxGradient}, one of the most accurate focus loss functions on Table~\ref{tab:boxrot:poster:all}, are given in Fig.~\ref{fig:rot:boxplot}.

\vspace{-2ex}
\paragraph{Computational Cost.}

\begin{table}
\centering
\begin{adjustbox}{max width=\columnwidth} %
\begin{tabular}{l|cc|cc|r}
\toprule 
\textbf{Focus Loss Function}  & \multicolumn{2}{c|}{\textbf{Boxes (RMS)}} & \multicolumn{2}{c|}{\textbf{Poster (RMS)}} & \textbf{Time}\\ 
 & \multicolumn{2}{c|}{$\omega_{\max}\approx\pm \SI{670}{\degree/\second}$} & \multicolumn{2}{c|}{$\omega_{\max}\approx\pm\SI{1000}{\degree/\second}$} & \\
 & w/o & w/ & w/o & w/ & {[}\si{\micro\second}{]} \\
\midrule 
Variance \eqref{eq:IWEVariance} \cite{Gallego18cvpr,Gallego17ral}  & 18.52  & 18.94  & 25.96  & 24.39  & \textbf{16.90} \\
Mean Square \eqref{eq:MeanSquareIWE} \cite{Gallego17ral,Stoffregen17acra}  & 19.93  & 19.02  & 34.10  & 26.31  & 25.11 \\
Mean Absolute Deviation \eqref{eq:MeanAbsoluteDevIWE}  & 19.46  & 19.58  & 30.70  & 29.62 & 78.11 \\
Mean Absolute Value  & -  & 19.77  & -  & 29.90  & 23.89 \\
Entropy \eqref{eq:maxEntropy}  & 28.50  & 26.54  & 47.54  & 33.21  & 271.85 \\
Area \eqref{eq:minSupp} (Exp)  & 31.50  & 19.54  & 43.12  & 26.40  & 160.56 \\
Area \eqref{eq:minSupp} (Gaussian)  & 25.85  & 18.85  & 34.50  & 25.35  & 1098.64\\
Area \eqref{eq:minSupp} (Lorentzian)  & 32.43  & 20.98  & 35.86  & 26.57  & 777.98\\
Area \eqref{eq:minSupp} (Hyperbolic)  & 29.13  & 19.15  & 32.94  & 25.88  & 1438.15\\
Range \eqref{eq:SupportPDFSimplified} (Exp)  & 28.66  & 28.72  & 65.33  & 32.23 & 263.11\\
Local Variance~\eqref{eq:local_variance_aggr}  & 18.21  & 18.40  & 25.44  & 24.15  & 78.48\\
Local Mean Square  & 24.81  & 19.86  & 33.95  & 26.47  & 137.20\\
Local Mean Absolute Dev.  & 21.37  & 18.74  & 61.89  & 25.29 & 177.15\\
Local Mean Absolute Value  & -  & 24.10  & -  & 30.37  & 243.58 \\
Moran's Index~\eqref{eq:DefMoranIndex}  & 24.28  & 23.43  & 32.40  & 30.96 & 116.39 \\
Geary's Contiguity Ratio~\eqref{eq:DefGearyC}  & 23.87  & 19.50  & 26.61  & 25.23  & 181.73\\
Gradient Magnitude \eqref{eq:maxGradient}  & \textbf{17.83}  & 18.10  & \textbf{23.93}  & 23.58 & 128.46\\
Laplacian Magnitude \eqref{eq:maxLaplacian}  & 18.32  & \textbf{17.58}  & 24.91  & 23.67 & 293.80 \\
Hessian Magnitude \eqref{eq:maxHessian}  & 18.41  & 17.93  & 25.47  & 23.74 & 569.55\\
Difference of Gauss. (DoG)  & 20.85  & 19.25  & 24.50  & \textbf{22.15} & 189.90 \\
Laplacian of the Gauss. (LoG)  & 20.36  & 17.77  & 25.15  & 24.01  & 127.65 \\
Variance of Laplacian  & 18.26  & 18.01  & 26.59  & 23.62 & 327.60 \\
Variance of Gradient  & 18.69  & 19.08  & 26.60  & 24.22 & 872.03\\
Variance of Squared Gradient  & 18.72  & 18.95  & 26.10  & 24.43 & 653.62\\
Mean Time on Pixel~\cite{Mitrokhin18iros}  & 82.89  & -  & 121.20  & - & 24.43\\
\bottomrule
\end{tabular}
\end{adjustbox}
\vspace{-0.5ex}
\caption{\label{tab:boxrot:poster:all}\emph{Accuracy and Timing Comparison of Focus Loss Functions}.
RMS angular velocity errors (in \si{\degree/\second}) of the motion compensation method~\cite{Gallego18cvpr} (with (w/) or without (w/o) polarity) with respect to motion-capture system. 
Processing $\numEvents = \SI{30000}{events}$, warped onto an image of $240\times 180$ pixels (DAVIS camera~\cite{Brandli14ssc}), takes \SI{2040.24}{\micro\second},
the runtime of the focus loss functions is given in the last column.
Sequences \texttt{\small{}boxes} and \texttt{\small{}poster}
from dataset~\cite{Mueggler17ijrr}.
The best value per column is highlighted in bold.}
\vspace{-1ex}
\hrulefill\vspace{-1ex}
\end{table}

The last column of Table~\ref{tab:boxrot:poster:all} compares the losses in terms of computational effort, on a single-core \SI{2.7}{\giga\Hz} CPU.
The variance is the fastest, with \SI{17}{\micro\second} (possibly due to efficient code in OpenCV), and its runtime is negligible compared to the time required to warp the events ($\approx$\SI{2}{\milli\second}). 
The MS and the MAV are also fast.
Other focus functions take longer, due to effort spent in spatial operations (convolutions), and/or transcendental functions (e.g., $\exp$).
The variance is a notable trade-off between accuracy and speed\footnote{Implementation plays a major role in runtime; thus the above figures are illustrative. 
We built on top of the C++ Standard Library and OpenCV, but did not optimize the code or used function approximation~\cite{Gallego14tcsi,Berjon16tcyb}.
\cite{Mitrokhin18iros,Stoffregen19arxiv} suggest considerable speed-up factors if warping is done on a GPU.
}.
Computing the IWE has $O(\numEvents)$ complexity, but it is, to some extent, parallelizable.
Loss functions act on the IWE (with $\numPixels$ pixels), thus they have $O(\numPixels)$ complexity.

\subsection{Depth Estimation}
\begin{figure}[t]
\centering
\includegraphics[width=\columnwidth]{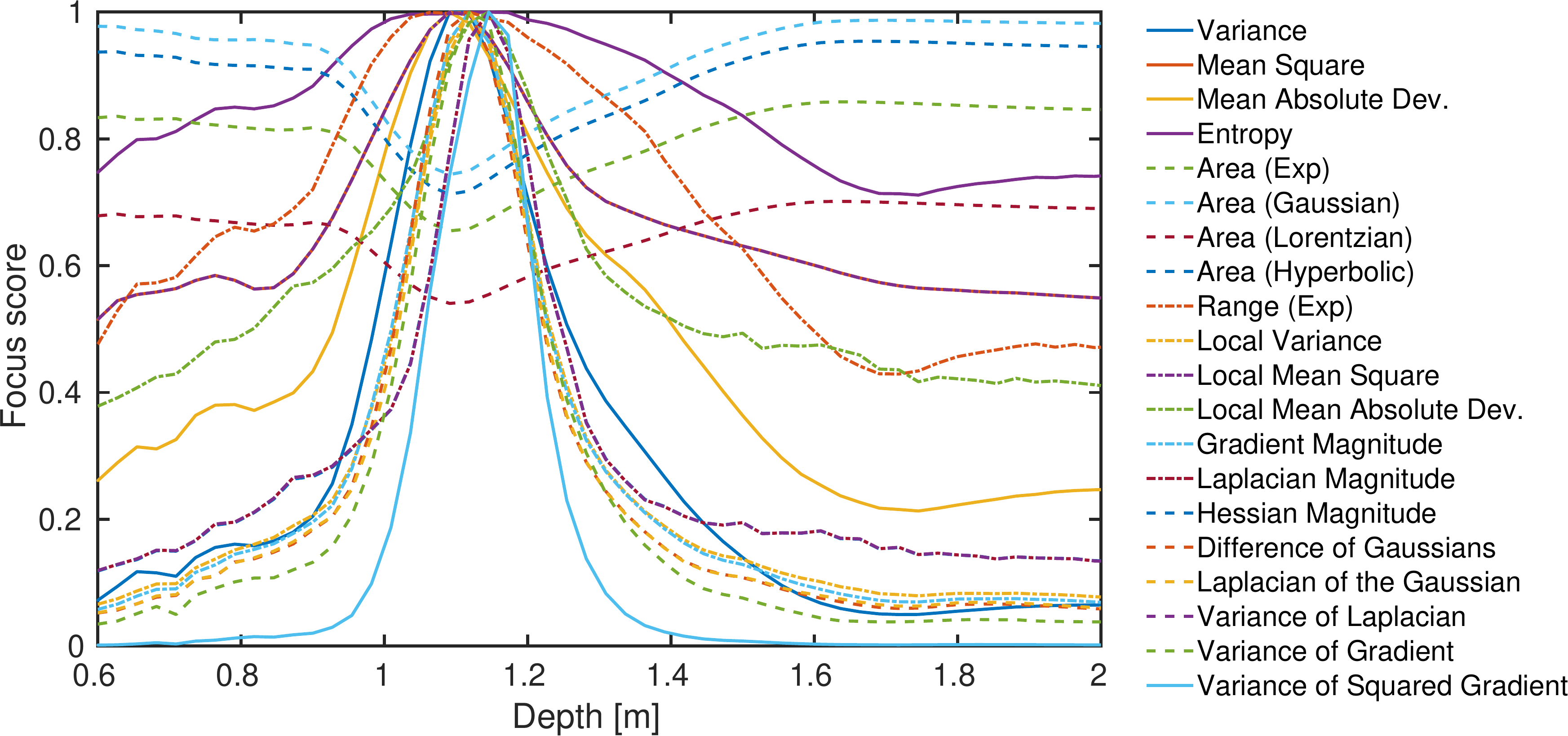}
\vspace{-4ex}
\caption{\label{fig:depthVsDepth}Depth estimation for a patch.
Plots of several focus losses as a function of depth.
Each focus loss curve is normalized (e.g., by its maximum), to stay in $[0,1]$.
}
\vspace{-1ex}\hrulefill\vspace{-2ex}
\end{figure}

We also qualitatively compared the loss functions in the context of depth estimation.
Here, the IWE is a slice of the disparity space image (DSI) computed by back-projecting events into a volume~\cite{Rebecq17ijcv}.
The DSI plays the role of the focal stack in SFF and AF (Section~\ref{sec:Focus:ProsAndCons}).
Depth at a pixel of the reference view ($\bparams\equiv Z$) is estimated by selecting the DSI slice with extremal focus along the optical ray of the pixel.
Fig.~\ref{fig:depthVsDepth} shows the focus loss functions along an optical ray for the events in a small space-time window around the pixel.
The values of the focus losses come to an extremal at a common focal point (at depth$\,\approx$\SI{1.1}{\meter}).
The plots show that, in general, the focus functions monotonically decrease/increase as the distance from the focal plane increases, forming the basin of attraction of the local extrema. 
Composite losses (Section~\ref{sec:Focus:CompositeLosses}) have a narrow peak, whereas others, such as the area losses have a wider peak.

\subsection{Optical Flow Estimation}
\begin{figure}[t]
\subfloat[Patch and three candidate flow vectors]{\includegraphics[width=0.32\columnwidth]{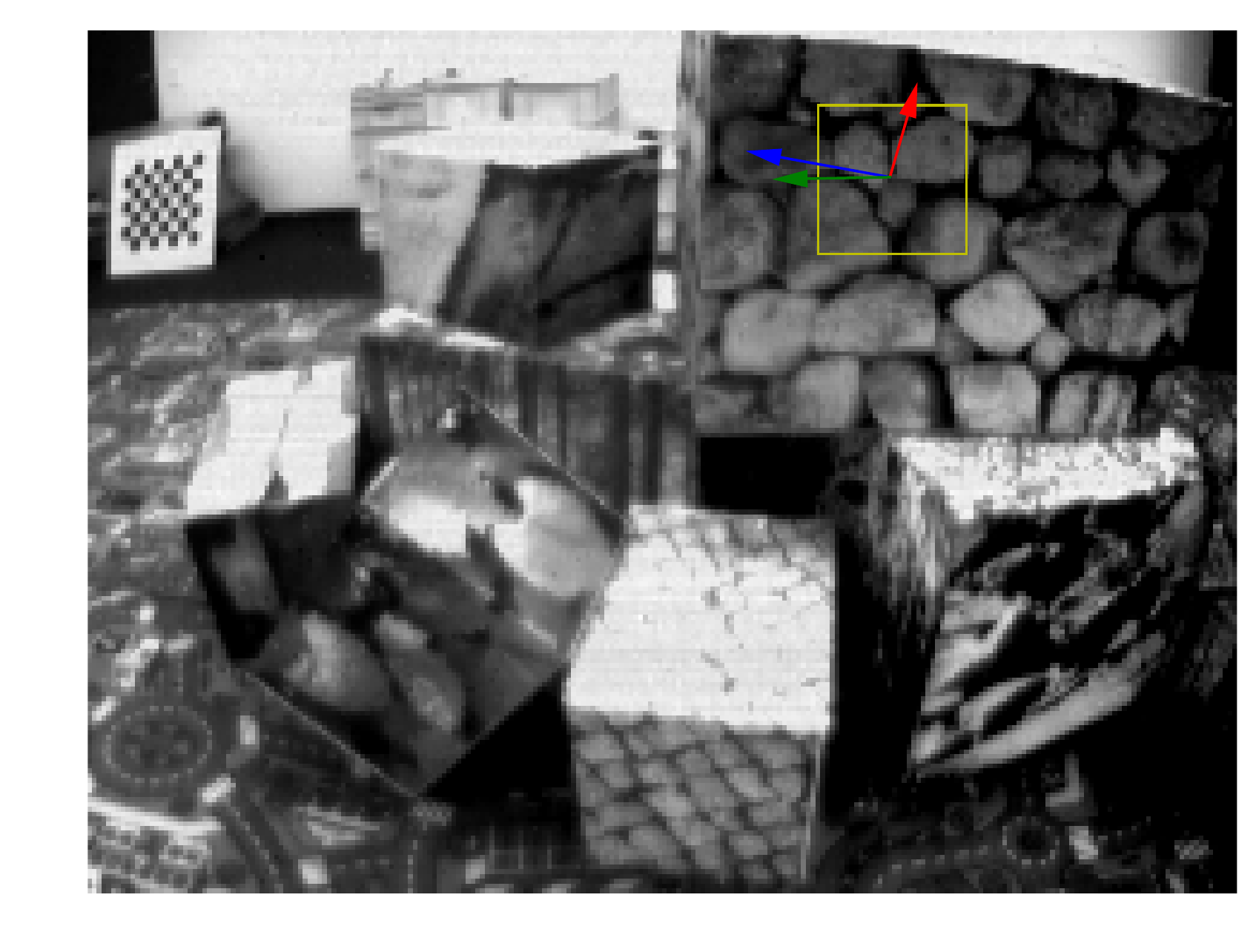}}~
\subfloat[Variance~\eqref{eq:IWEVariance}.]{\includegraphics[width=0.32\columnwidth]{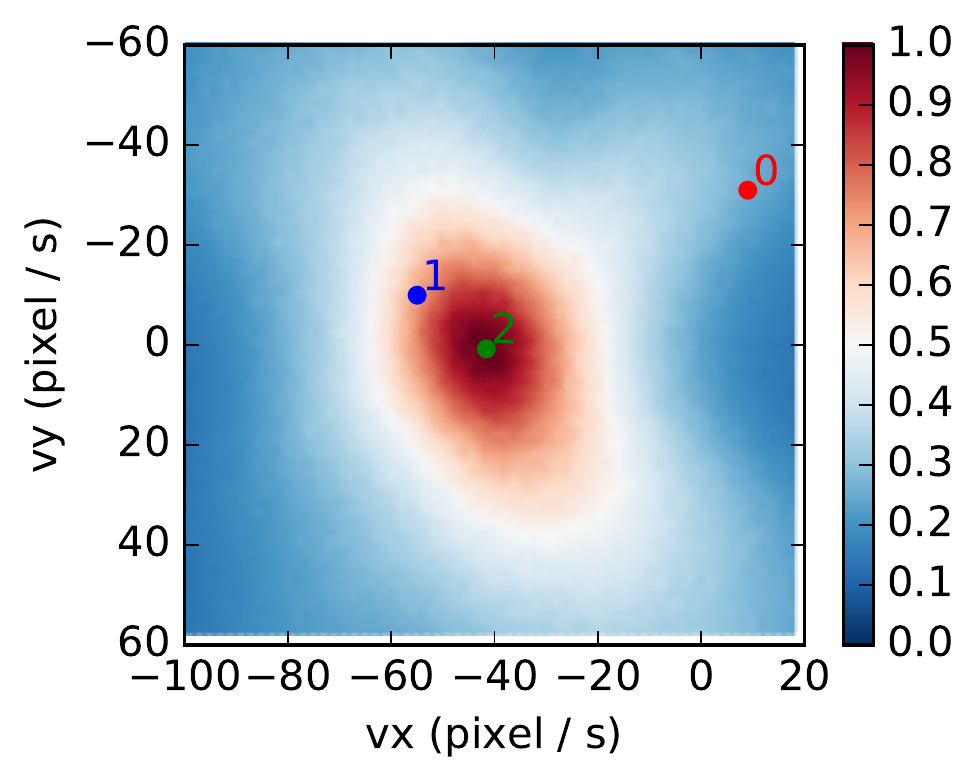}}~
\subfloat[Gradient Magnitude~\eqref{eq:maxGradient}.]{\includegraphics[width=0.32\columnwidth]{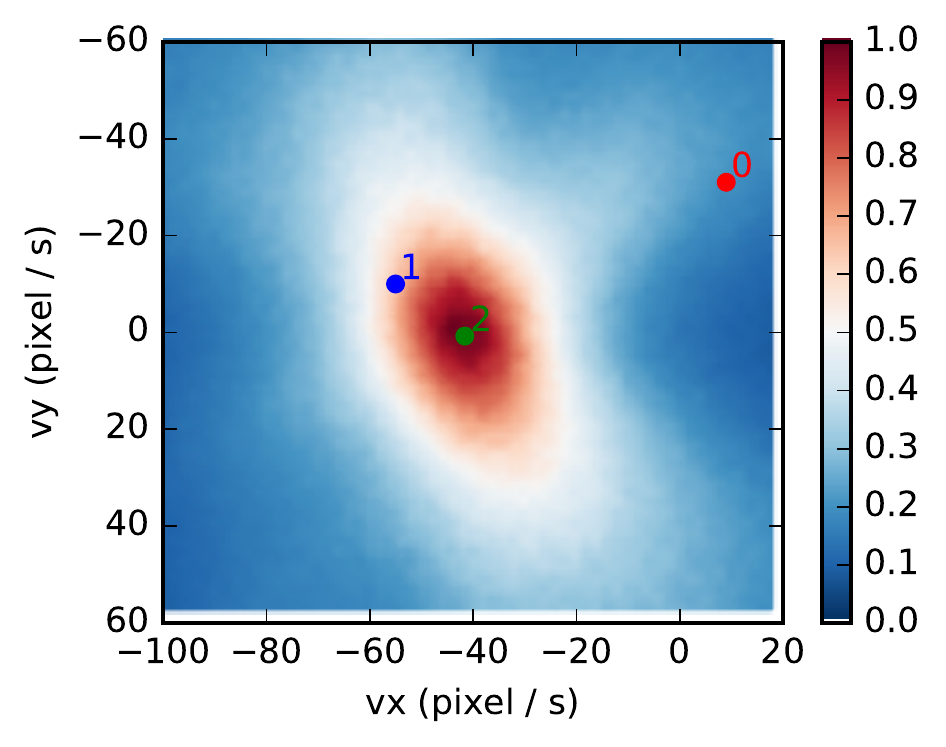}}\\[-1ex]
\subfloat[Laplacian Magnitude~\eqref{eq:maxLaplacian}.]{\includegraphics[width=0.32\columnwidth]{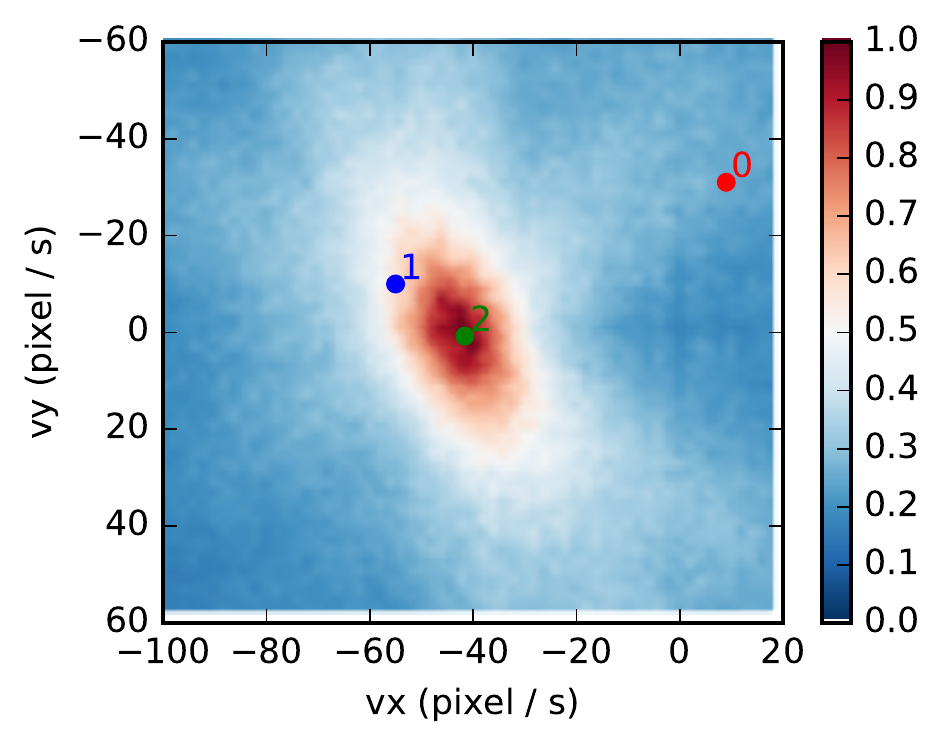}}~
\subfloat[Variance of Gradient Magnitude.]{\includegraphics[width=0.32\columnwidth]{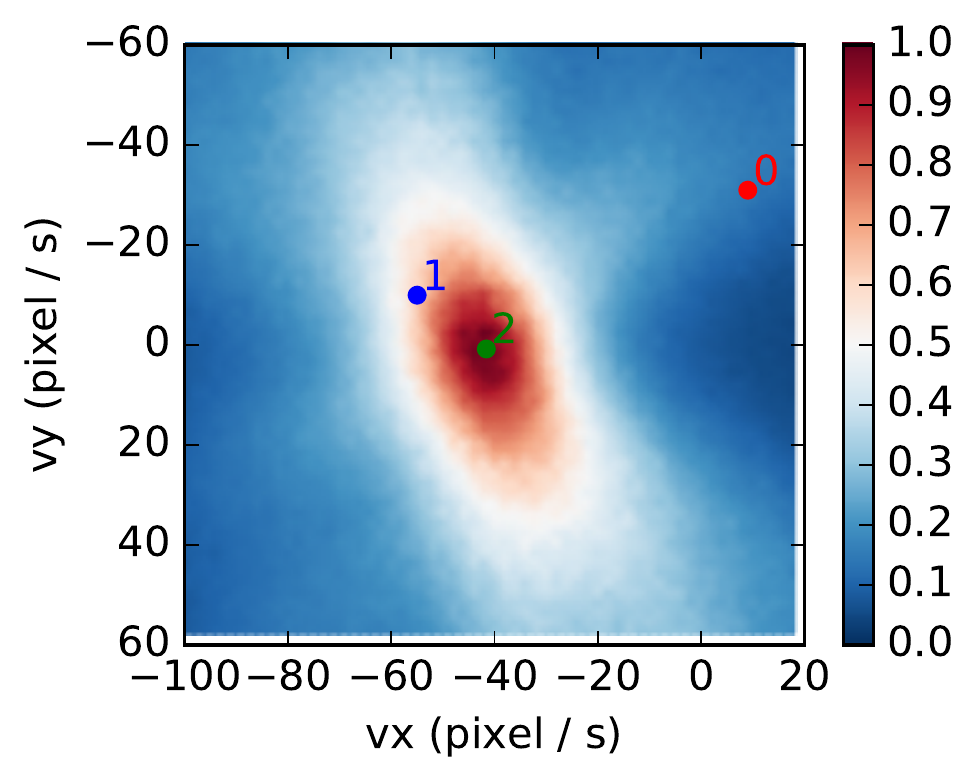}}~
\subfloat[Variance of Laplacian.]{\includegraphics[width=0.32\columnwidth]{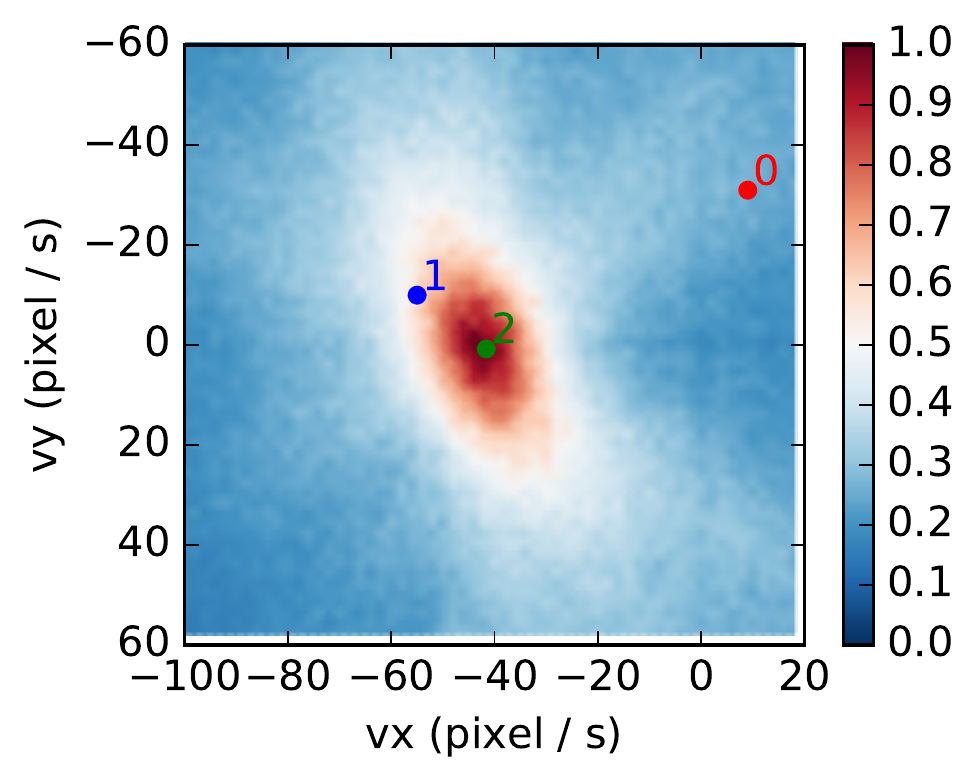}}
\vspace{-1ex}
\caption{\label{fig:oflow}
Optical flow of a patch (feature). %
Plots of several focus losses as a function of the two optical flow parameters of the feature.
Each focus loss surface is normalized by its maximum.
At the correct optical flow vector, some surfaces show a narrow peak, whereas others present a broader peak.
}
\vspace{-1ex}\hrulefill\vspace{-2ex}
\end{figure}

The profile of the loss function can also be visualized for 2D problems such as patch-based optical flow~\cite{Gallego18cvpr}. 
Events in a small space-time window (e.g., $15\times 15$ pixels) are warped according to a feature flow vector $\btheta\equiv\vel$: 
$\bx'_k= \bx_k - (t_k-t_\text{ref}) \vel$.
Fig.~\ref{fig:oflow} shows the profiles of several competitive focus losses. 
They all have a clear extremal at the location of the visually correct ground truth flow (point 0, green arrow in Fig.~\ref{fig:oflow}a).
The Laplacian magnitude and its variance show the narrowest peaks.
Plots of more loss functions are provided in the \revred{supplementary material}.

\subsection{Unsupervised Learning of Optical Flow}
The last row of Table~\ref{tab:boxrot:poster:all} evaluates the accuracy and timing of a loss function inspired in the time-based IWE of~\cite{Mitrokhin18iros}:
the variance of the per-pixel average timestamp of warped events.
This loss function has been used in~\cite{Zhu18arxiv} for unsupervised training of a neural network (NN) that predicts dense optical flow from events~\cite{Zhu18rss}.
However, this loss function is considerably less accurate than all other functions (Table~\ref{tab:boxrot:poster:all}). 
This suggests that
(\emph{i}) loss functions based on timestamps of warped events are not as accurate as those based on event count (the event timestamp is already taken into account during warping),
(\emph{ii}) there is considerable room for improvement in unsupervised learning of optical flow if better loss function is used.

\begin{figure}[t]
\centering
	\frame{\includegraphics[width=0.445\linewidth]{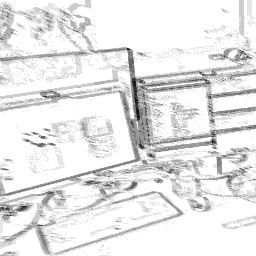}}~~~
    \frame{\includegraphics[width=0.445\linewidth]{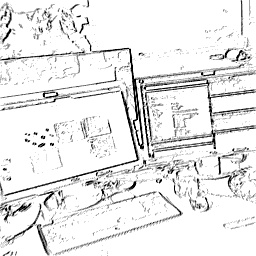}}
    \caption{
    \label{fig:networkflow}
    IWE before (Left) and after (Right) motion compensation by a neural network similar to~\cite{Zhu18arxiv}, trained using the gradient magnitude loss~\eqref{eq:maxGradient}.}
    \vspace{-1ex}\hrulefill\vspace{-2ex}
\end{figure}

To probe the applicability of the loss functions to estimate dense optical flow, we trained a network inspired in~\cite{Zhu18rss} using~\eqref{eq:maxGradient} and a Charbonier prior on the flow derivative.
The flow produced by the NN increases event alignment (Fig.~\ref{fig:networkflow}).
A deeper evaluation is left for future work.

\section{Conclusion}
\label{sec:conclusion}
\vspace{-1ex}
We have extended motion compensation methods for event cameras with a library of twenty more loss functions that measure event alignment.
Moreover, we have established a fundamental connection between the proposed loss functions and metrics commonly used in shape from focus with conventional cameras. 
This connection allows us to bring well-established analysis tools and concepts from image processing into the realm of event-based vision.
The proposed functions act as focusing operators on the events, enabling us to estimate the point trajectories on the image plane followed by the objects causing the events.
We have categorized the loss functions according to their capability to measure edge strength and dispersion, metrics of information content.
Additionally, we have shown how to design new focus metrics tailored to edge-like images like the image of warped events.
We have compared the performance of all focus metrics in terms of accuracy and time.
Similarly to comparative studies in autofocus for digital photography applications~\cite{Mir14spie}, 
we conclude that the variance, the gradient magnitude and the Laplacian are among the best functions.
Finally, we have shown the broad applicability of the functions to tackle essential problems in computer vision: ego-motion, depth, and optical flow estimation.
We believe the proposed focus loss functions are key to taking advantage of the outstanding properties of event cameras, specially in unsupervised learning of structure and motion.

\subsection*{Acknowledgment}
\vspace{-1ex}
This work was supported by the Swiss National Center of Competence Research Robotics, the Swiss National Science Foundation and the SNSF-ERC Starting Grant.

\cleardoublepage
\appendix
\title{\MYTITLE\\---Supplementary Material---}
\maketitle
\ifshowpagenumbers
\else
\thispagestyle{empty}
\fi

\section{Notation}

\subsection{$L^{p}$ norm of a vector-valued function}
\label{sec:app:LpNormDef}

We define the $L^{p}$ norm of a vector-valued function $\f:\Omega\subset\R^{d}\to\R^{n}$ with components $\{f_i\}_{i=1}^{n}$ (i.e., $\f=(f_{1},\ldots,f_{n})^{\top}$) by
\begin{equation}
\label{eq:LpNormFunction}
\|\f\|_{L^{p}(\Omega)}\doteq\left(\int_{\Omega}\|\f(\bx)\|^{p}d\bx\right)^{1/p},
\end{equation}
where $\|\f(\bx)\| = (\sum_{i=1}^{n}|f_{i}(\bx)|^{p})^{1/p}$ is the $p$-norm in $\R^{n}$. 
With this convention,
\begin{equation}
\|\f\|_{L^{p}(\Omega)}^{p}=\sum_{i=1}^{n}\int_{\Omega}|f_{i}(\bx)|^{p}d\bx =\sum_{i=1}^{n}\|f_{i}\|_{L^{p}(\Omega)}^{p},
\end{equation}
that is, the $p$-th power of the norm of $\f$ is the sum of the $p$-th power of the norms of its components.

The two most common cases are $p=1$ and $p=2$, which, for the gradient of an image, $\nabla\IWE = (I_x,I_y)^\top$, yield simple expressions:
\begin{equation}
\label{eq:LOneNormVectorValued}
\|\nabla I\|_{L^1(\Omega)} = \int_\Omega \bigl(|I_x(\bx)| + |I_y(\bx)| \bigr) \, d\bx\end{equation}
and 
\begin{equation}
\label{eq:LTwoNormVectorValued}
\|\nabla I\|^2_{L^2(\Omega)} = \int_\Omega \bigl(I^2_x(\bx) + I^2_y(\bx)\bigr)\, d\bx.
\end{equation}

\subsection{Hessian Matrix}

The Hessian matrix of a function, such as the IWE (used in~\eqref{eq:maxHessian}), is denoted by
\begin{equation}
\label{eq:HessianMatrix}
\hessian(I) = \left(\begin{array}{cc}
I_{xx} & I_{xy}\\
I_{xy} & I_{yy}
\end{array}\right),
\end{equation}
where the subscripts indicate derivatives.

The trace of the Hessian matrix is the Laplacian, which is used to define loss function~\eqref{eq:maxLaplacian}.

\section{Area of the Image of Warped Events}
\label{sec:app:MinSupport}

To measure the ``thickness'' of the edges of the IWE (e.g., Fig.~\ref{fig:supp:illustr:IWE}), one could count the number of pixels with count of warped events above a threshold, e.g., one event. 
However, this is brittle since it depends on this arbitrary threshold.
We propose to define the above-mentioned edge thickness or ``area'' of an edge-like image like the IWE~\eqref{eq:IWE} in a more sensible way as a weighted sum of the interior of the level sets of the image, as we show next.

\subsection{Definition of the Area of an Image}

\global\long\def\widthAreaOneDim{0.9\columnwidth}
\begin{figure}
\centering
\subfloat[$\Delta x=1.4$ pixel]{
\includegraphics[width=\widthAreaOneDim]{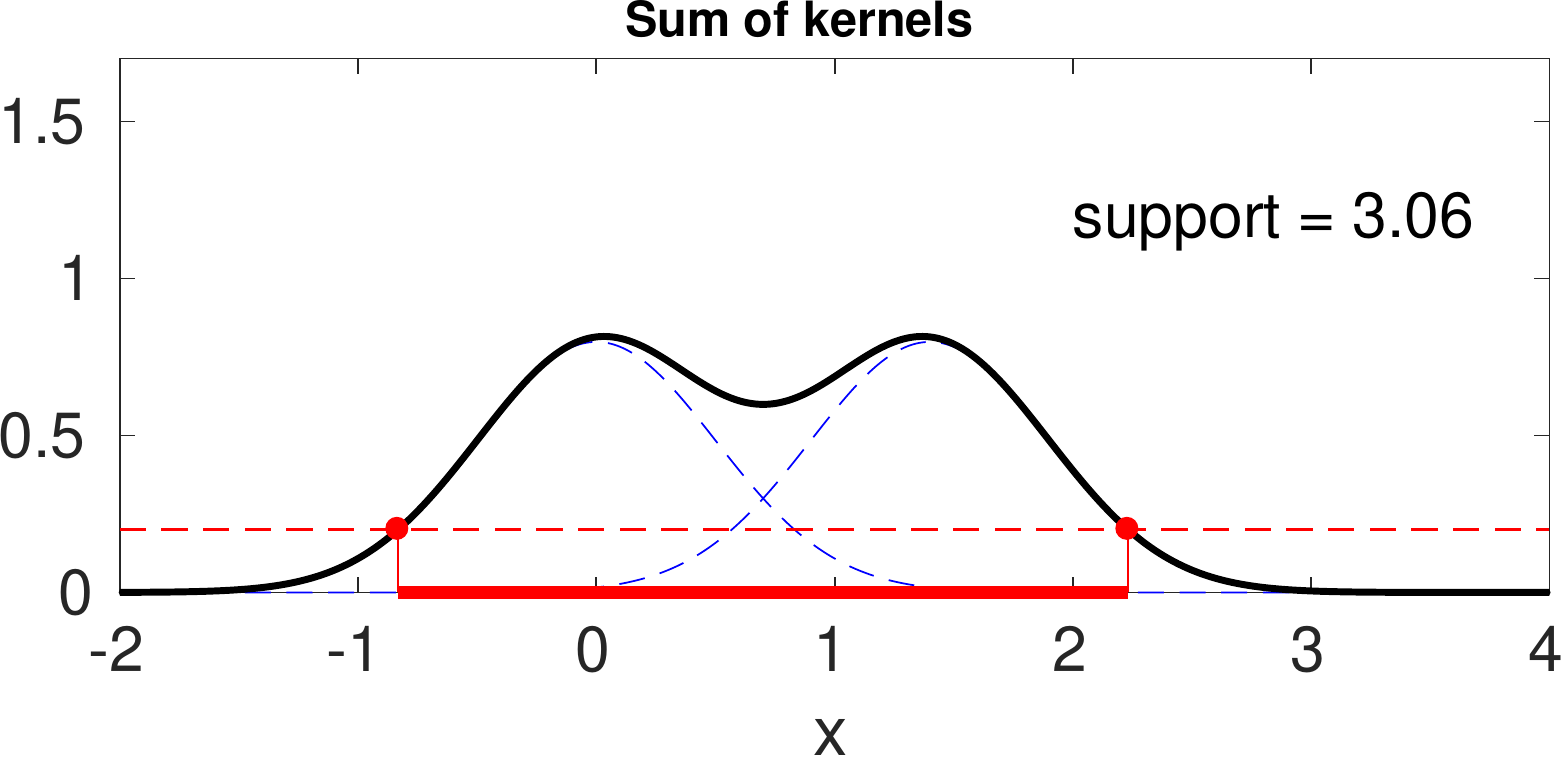}}\\
\subfloat[$\Delta x=0.7$ pixel]{
\includegraphics[width=\widthAreaOneDim]{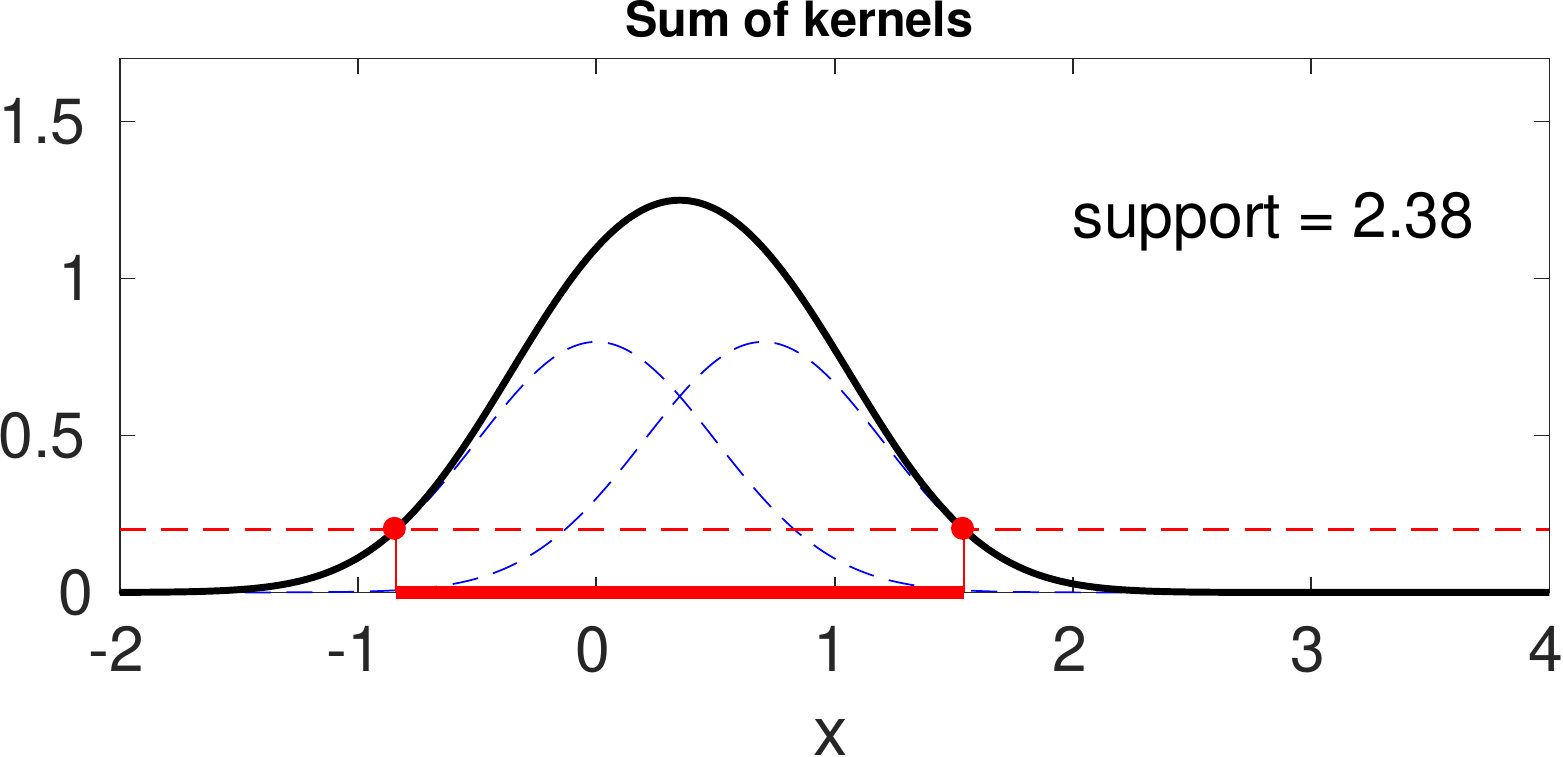}}\\
\subfloat[$\Delta x=0$ pixel]{
\includegraphics[width=\widthAreaOneDim]{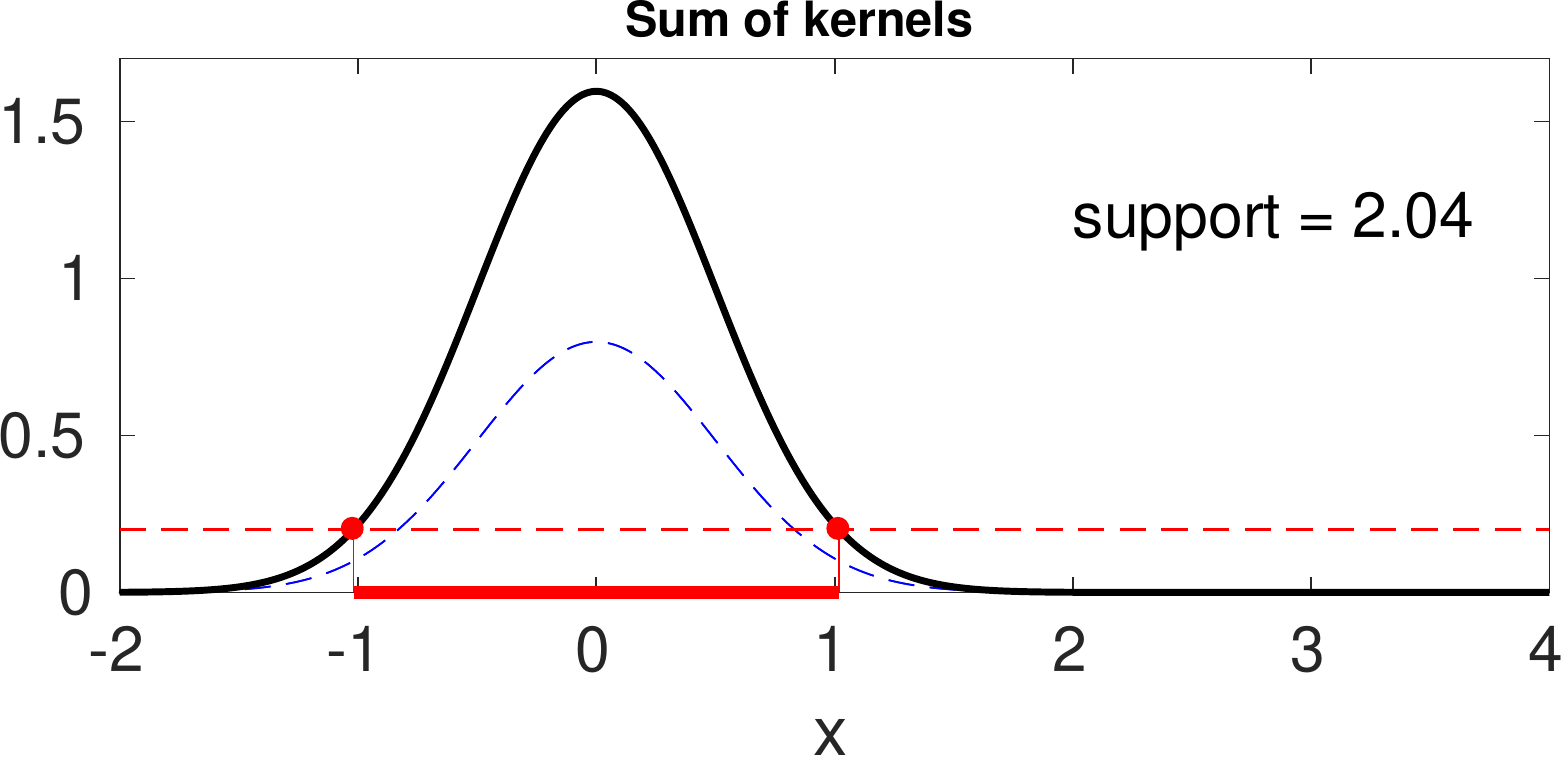}}
\caption{\label{fig:supp-example}
Illustration of the ``area'' or support of a 1-D image~\eqref{eq:IWE} with two events.
The sum of the Gaussian kernels centered on each warped event (in blue) produces $\IWE(x)$ (in black), whose support (i.e., the set $\{x\in\R\,|\,I(x)>\lambda = 0.2\}$) is displayed in solid red. 
The more aligned the events (smaller $\Delta x = x'_{2}-x'_{1}$), the smaller the support of $I(x)$.}
\end{figure}

Using a Gaussian function (kernel) as a smooth approximation to the Dirac delta, $\delta(\bx-\bmu)\approx\cN(\bx;\bmu,\sigma^2\mId)$, the image of warped events~\eqref{eq:IWE} has, strictly speaking, an unbounded support (area of pixels with non-zero value).
To have a meaningful support measure, 
we instead count the number of pixels with value greater than\footnote{We assume that $I(\bx)\geq 0$ either because event polarity is not used ($b_k=1$ in~\eqref{eq:IWE}) or because the support here defined is applied to images of positive and negative events separately, and the results are added.}
$\lambda$, 
\begin{equation}
\supp(I;\lambda) \doteq \int_{\Omega}H(I(\bx)>\lambda)\,d\bx,
\label{eq:app:SuppDefHeaviside}
\end{equation}
where $H(\cdot)$ is the Heaviside function.
Fig.~\ref{fig:supp-example} shows several examples of it.
This figure also illustrates the principle of area minimization, for a 1-D signal~\eqref{eq:IWE} with just two warped events. 
As observed, the area or thickness of $I$ is minimized if the events are warped to the same location ($\Delta\bx'=\bzero\iff\bx'_{i}=\bx'_{j}$), which is the desired event alignment condition of corresponding events.

To have a support metric that does not depend on the particular value of the threshold $\lambda$ used (for fixed kernel width $\sigma$), 
we sum~\eqref{eq:app:SuppDefHeaviside} over all threshold values,
\begin{equation}
\supp(\IWE)\doteq\int_{0}^{\infty}\rho(\lambda)\supp(\IWE;\lambda)\,d\lambda,\label{eq:app:AggrSuppWeighted}
\end{equation}
using a decreasing weighting function $\rho$, such as $e^{-\lambda}$, thus emphasizing the areas corresponding to $\lambda\approx0$ over those associated to $\lambda\gg 0$.
In this way, an algorithm minimizing~\eqref{eq:app:AggrSuppWeighted} will focus its attention on decreasing the area contribution of small thresholds, which are more important since the areas of larger thresholds are smaller due to the $\lambda$-support sets $\{\bx\in\R^{2}\,|\, I(\bx)>\lambda\}$ forming a family of nested subsets.

Notice that it is not possible to use $\rho = \text{const}$ since this leads to $\supp(\IWE)=\numEvents$, which does not depend on the motion parameters $\bparams$ we wish to optimize for.
Using weighting functions with unit area (i.e., $\int_0^\infty \rho(\lambda)d\lambda = 1$) allows us to interpret~\eqref{eq:app:AggrSuppWeighted} as a convex combination of supports~\eqref{eq:app:SuppDefHeaviside}, thus setting the correct scale so that~\eqref{eq:app:AggrSuppWeighted} has the same units as~\eqref{eq:app:SuppDefHeaviside}.

\subsection{Simplification of the Area of an Image}

Substituting~(\ref{eq:app:SuppDefHeaviside}) in~(\ref{eq:app:AggrSuppWeighted})
and swapping the order of integration gives
\begin{align}
\supp(\IWE) 
&=\int_{\Omega}\int_{0}^{\infty}\rho(\lambda) \, H(I(\bx)>\lambda)\,d\lambda \,d\bx \nonumber\\
&=\int_{\Omega}\int_{0}^{\IWE(\bx)}\rho(\lambda) \,d\lambda \,d\bx \nonumber\\
&=\int_{\Omega}\left[F(\lambda)\right]_{0}^{\IWE(\bx)}\,d\bx \nonumber\\
&=\int_{\Omega}\bigl(F(\IWE(\bx))-F(0)\bigr)\,d\bx,
\label{eq:app:WeightedSupp}
\end{align}
where $F(\lambda) \doteq \int\rho(\lambda)d\lambda$ is a primitive of $\rho$, and $F(0)$ is constant. 
This is an advantageous expression compared to~\eqref{eq:app:AggrSuppWeighted}, since it states that $\supp(\IWE)$ can be computed using the values of $\IWE(\bx)$ directly,
without having to compute~\eqref{eq:app:SuppDefHeaviside} for every threshold $\lambda$ and then sum up the results.
By using a continuous image formulation, we have analytically integrated the partial sums~\eqref{eq:app:SuppDefHeaviside}.

\begin{figure}[t]
\centering
\global\long\def\widthpatchsupp{0.47\columnwidth}
\subfloat[Warped events for $\bparams_1$\label{fig:suppPatch:NotBestPatch}]{\includegraphics[width=\widthpatchsupp]{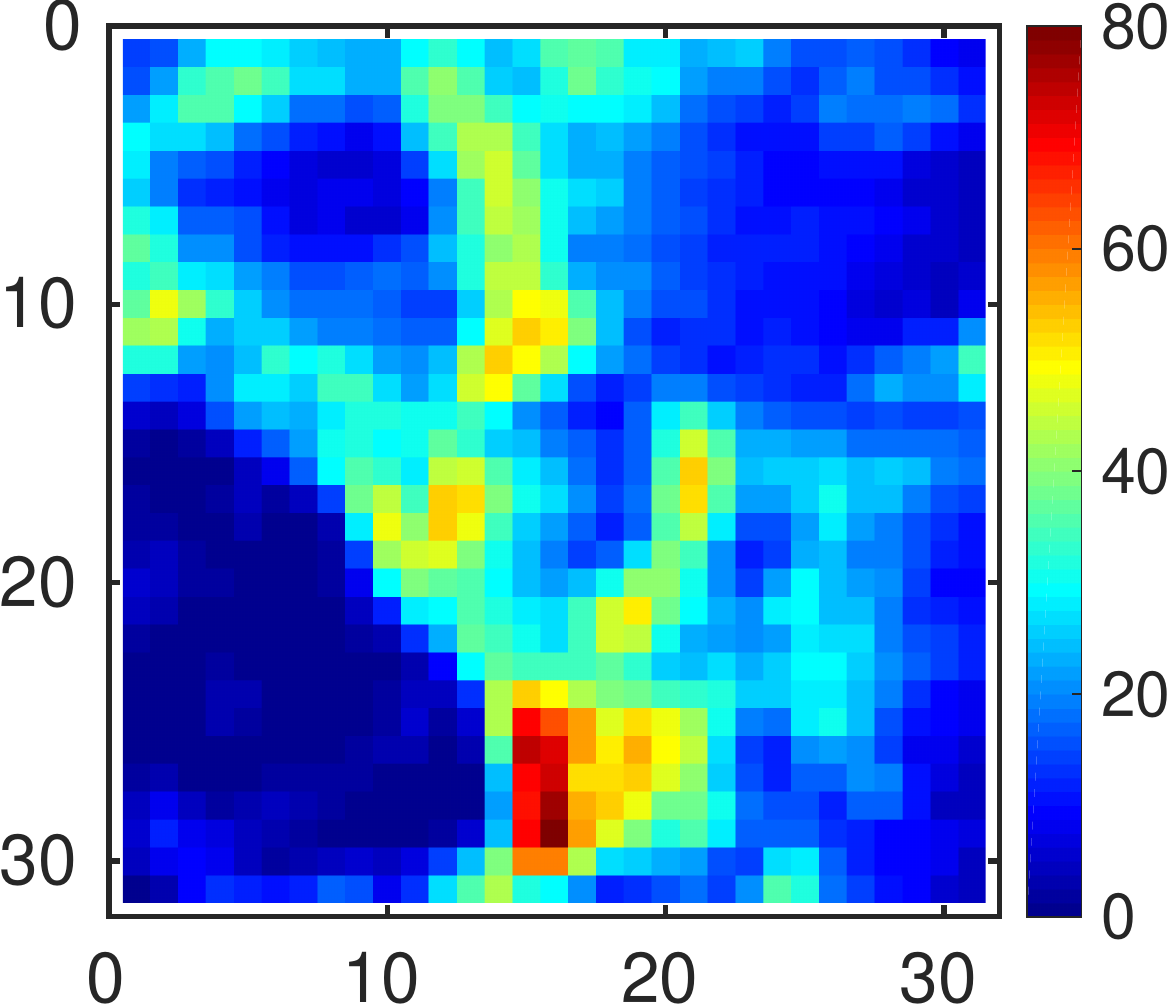}}~~
\subfloat[Support map for $\bparams_1$\label{fig:suppPatch:NotBestSupp}]{\includegraphics[width=\widthpatchsupp]{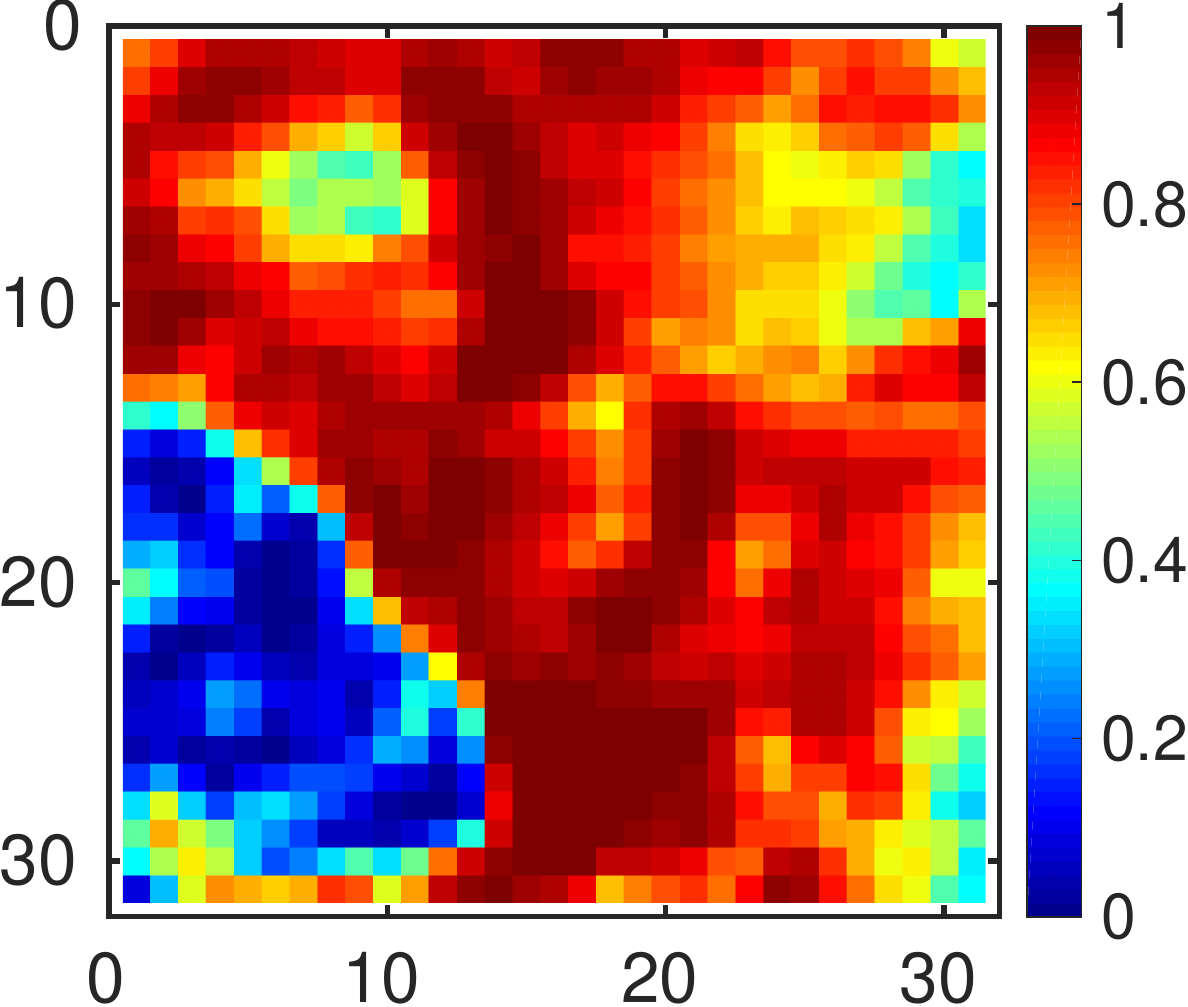}}\\
\subfloat[Warped events for $\bparams^\ast$\label{fig:suppPatch:BestPatch}]{\includegraphics[width=\widthpatchsupp]{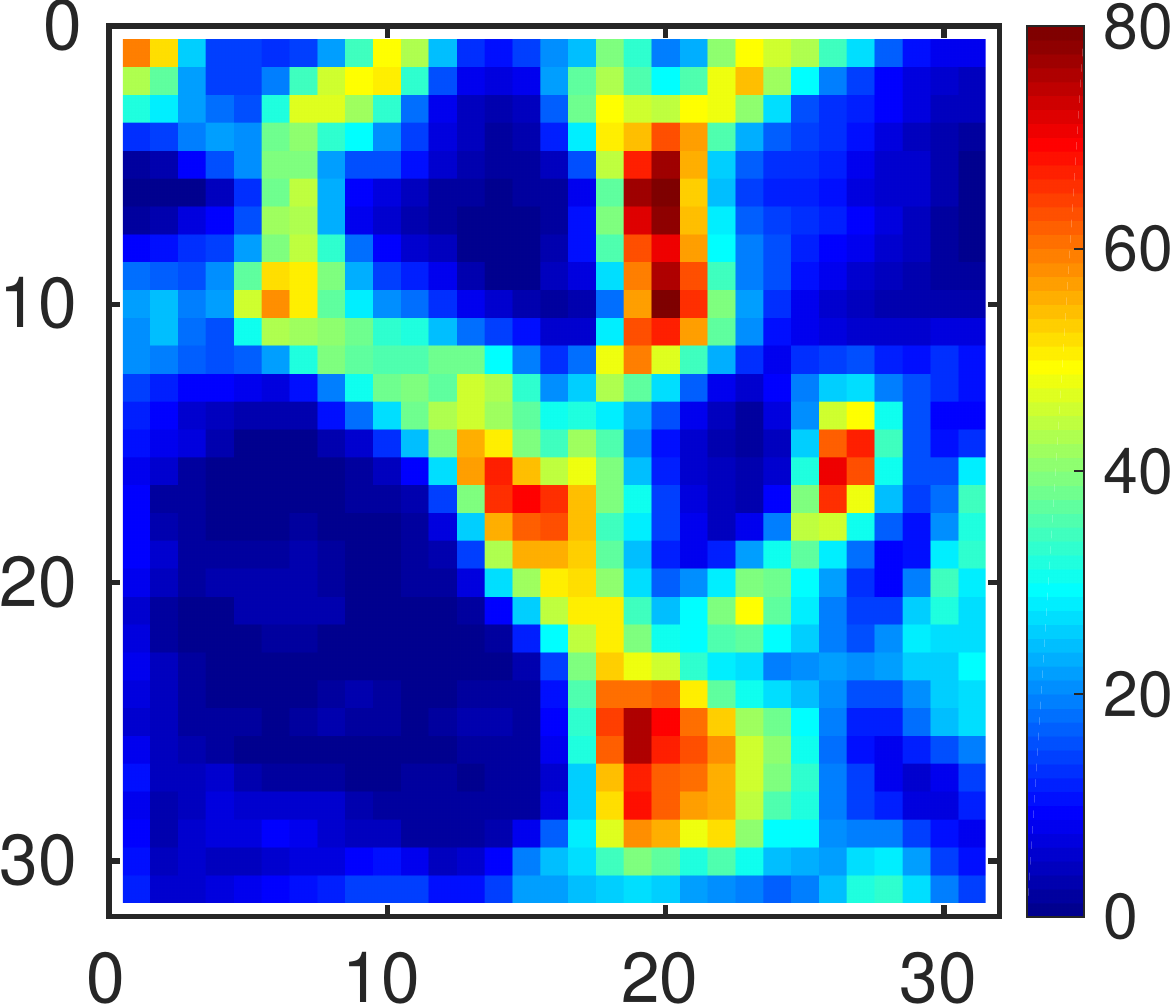}}~~
\subfloat[Support map for $\bparams^\ast$\label{fig:suppPatch:BestSupp}]{\includegraphics[width=\widthpatchsupp]{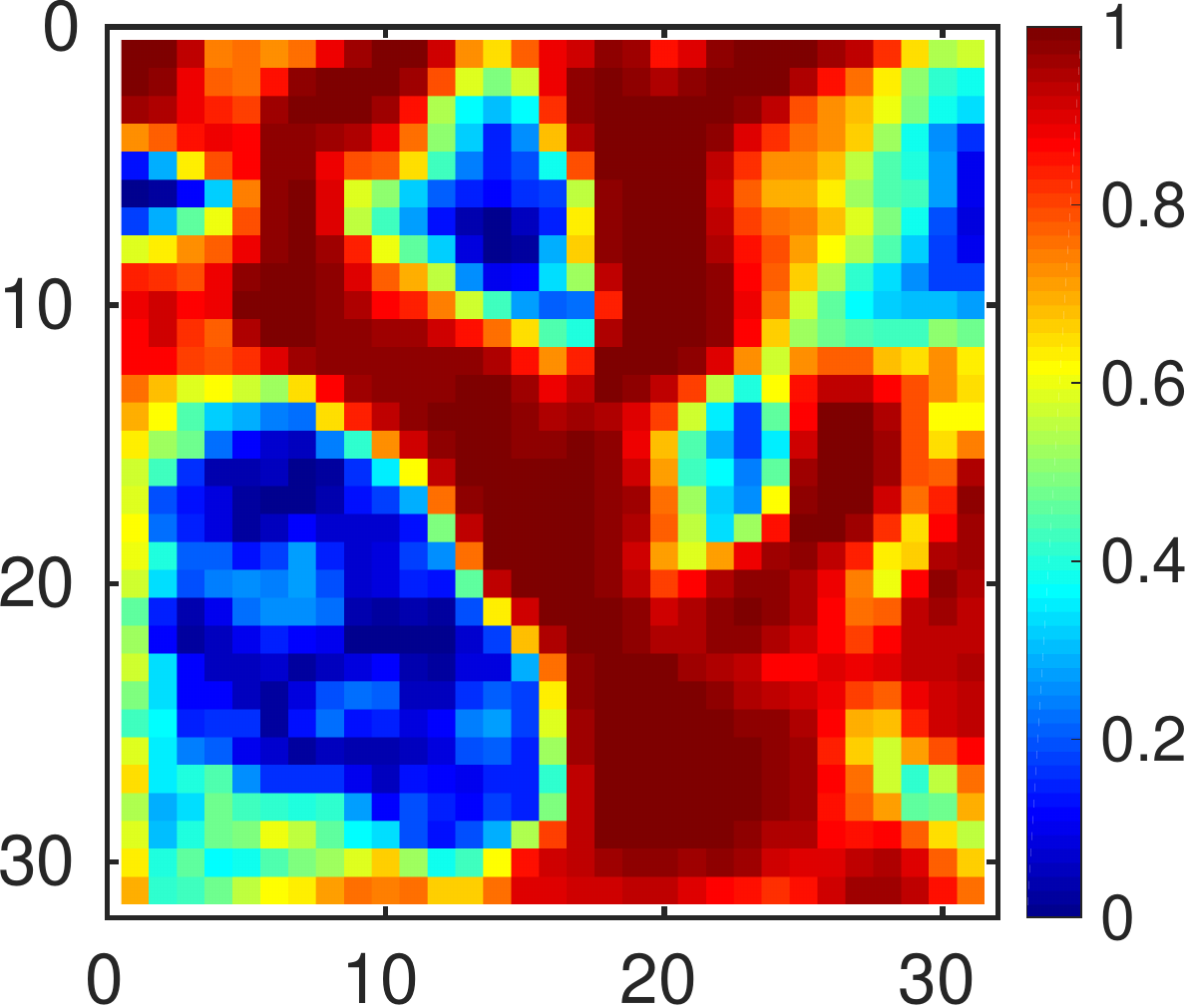}}
\caption{\label{fig:suppPatch}Illustration of the area~\eqref{eq:app:WeightedSupp} of a patch/image of warped events.
Warped events $\IWE(\bx;\bparams)$ (left column) and corresponding per-pixel support maps $F(\IWE(\bx;\bparams) / 10)-F(0)$ in~\eqref{eq:app:WeightedSupp} (right column), 
for two different motion parameters $\bparams$ (top: $\bparams_1\equiv$ suboptimal; bottom: $\bparams_2\equiv\bparams^\ast$ optimal).
Support scores~\eqref{eq:app:WeightedSupp}: 
$\supp(\IWE)=0.72\,|\Omega|$ (top) and $\supp(\IWE)=0.65\,|\Omega|$ (bottom), with $|\Omega|=31^2 = 961$ pixels.
The bottom patch has a smaller area (i.e., thinner edges) than the top patch,
thus showing a better event alignment.}
\end{figure}

Fig.~\ref{fig:suppPatch} illustrates~\eqref{eq:app:WeightedSupp}. 
It shows the warped events $\IWE(\bx;\bparams)$ on a $31\times 31$ image patch for two different parameters $\bparams_1, \bparams_2$ (depth values, in this example~\cite{Gallego18cvpr}). %
It also shows the corresponding integrands of~\eqref{eq:app:WeightedSupp}, or ``per-pixel support maps'' $F(\IWE(\bx;\bparams) / \lambda_0)-F(0) \equiv 1-\exp(- (\IWE(\bx;\bparams) / \lambda_0) )$, with $\lambda_0=10$ warped events. 
Pixels with $\IWE(\bx) \gtrsim \lambda_0$ events contribute more to the support~\eqref{eq:app:WeightedSupp} than pixels with $\IWE(\bx) \lesssim \lambda_0$ events, as shown in the support maps (right column of Fig.~\ref{fig:suppPatch}), which are color-coded from blue (low contribution) to red (high contribution).

Basically, the red regions of the support maps approximately indicate the area of the IWE, 
whereas the blue regions indicate the pixels where few warped events accumulate and therefore do not effectively contribute to the area of the IWE.
Clearly, the bottom patch has a smaller area (i.e., thinner edges) than the top patch, as indicated by the smaller area of the red regions.
The image area~\eqref{eq:app:WeightedSupp} is used to define the focus loss function~\eqref{eq:minSupp}.

\section{Loss Function: Image Entropy}
\label{sec:app:Focus:Entropy}

As anticipated in~\eqref{eq:maxEntropy}, event alignment may be achieved by maximizing the entropy of the IWE, where
\begin{equation}
H\left(p(z)\right) \doteq -\int_{-\infty}^{\infty}p(z)\log p(z) dz
\label{eq:DefEntropy}
\end{equation}
is Shannon's (differential) entropy for a continuous random variable whose density function (PDF) is $p(z)$.

The PDF of an image is approximated by its histogram, normalized to have unitary area. 
In a continuous formulation, this is written as (see~\cite{Gallego13tip}) 
\begin{equation}
p_{\IWE}(z)\doteq\frac{1}{|\Omega|}\int_{\Omega}\delta\left(z-\IWE(\bx;\bparams)\right) d\bx,
\label{eq:PDFImage}
\end{equation}
using the Dirac delta. 
This equation intuitively says that $p_{\IWE}(z)$ is computed as a ratio of areas: 
the ``number of pixels'' of the IWE with value $z$, divided by the total ``number of pixels'', $|\Omega|=\int_\Omega d\bx = \numPixels$.

Substituting~\eqref{eq:PDFImage} into~\eqref{eq:DefEntropy}, the entropy of the IWE becomes
\begin{align}
H\left(p_{I}(z)\right)
&=-\int_{-\infty}^{\infty}p_{\IWE}(z)\log p_{\IWE}(z)dz \nonumber \\
&=-\int_{-\infty}^{\infty}\frac{1}{|\Omega|}\int_{\Omega}\delta(z-\IWE(\bx))d\bx\log p_{\IWE}(z)dz \nonumber \\
&=-\frac{1}{|\Omega|}\int_{\Omega}\left(\int_{-\infty}^{\infty}\delta(z-\IWE(\bx))\log p_{\IWE}(z)dz\right)d\bx \nonumber \\
&=-\frac{1}{|\Omega|}\int_{\Omega}\log p_{I}\left(\IWE(\bx)\right)d\bx.
\label{eq:EntropyLogPDFImageStatement}
\end{align}
Observe that the entropy is maximized by favoring large values of $\log (1/p_{\IWE}(\IWE(\bx)))$ over smaller ones. 
Since $\log$ is concave, it means that large values of $1/p_{I}(I(\bx))$ are favored, i.e., small values of $p_{\IWE}(\IWE(\bx))$ are favored.
For a PDF that is concentrated around $\IWE=0$ (large $p_{\IWE}(0)$, as shown on the last column of Fig.~\ref{fig:evol:dyn}), favoring small density values implies that they must be achieved away from $\IWE=0$,
i.e., for large $|\IWE|$ values (which are caused by the aggregation of aligned events). 
Thus, maximizing the entropy increases the range of $\IWE(\bx)$, producing a higher contrast image.

\section{Loss Function: Image Range}
\label{sec:app:Focus:Range}

We measure the image range by means of the support of its PDF~\eqref{eq:PDFImage}, 
\begin{equation}
\supp(p_{\IWE})\doteq\int_{0}^{\infty}\rho(\lambda)\supp(p_{\IWE}(z);\lambda)\,d\lambda,
\label{eq:app:DefSupportPDF}
\end{equation}
where the weight function $\rho(\lambda)\ge 0$ 
emphasizes the contributions of small $|\lambda|$ over those of large $|\lambda|$, according to the typical shape of the PDF of the event image (concentrated around $\lambda=0$).

Mimicking the steps in Section~\ref{sec:app:MinSupport}, Eq.~\eqref{eq:app:DefSupportPDF} can be rewritten as
\begin{align}
\supp(p_{\IWE}) 
&=\int_{-\infty}^{\infty}\int_{0}^{\infty}\rho(\lambda) \, H(p_{\IWE}(z)>\lambda)\,d\lambda \,dz \nonumber\\
&=\int_{-\infty}^{\infty}\int_{0}^{p_{\IWE}(z)}\rho(\lambda) \,d\lambda \,dz \nonumber\\
&=\int_{-\infty}^{\infty}\left[F(\lambda)\right]_{0}^{p_{\IWE}(z)}\,dz \nonumber\\
& =\int_{-\infty}^{\infty}\bigl(F(p_{\IWE}(z))-F(0)\bigr)\,dz, 
\label{eq:app:SupportPDFSimplified}
\end{align}
where $F(\lambda)$ is a primitive of $\rho(\lambda)$, and $F(0)$ is constant.

The motion parameters are found by maximizing~\eqref{eq:app:SupportPDFSimplified}, 
i.e.,~\eqref{eq:SupportPDFSimplified}. 
The same weighting functions and primitives as for the image area (Section~\ref{sec:Focus:MinSupport}) may be used 
(with even symmetry if event polarity is used in the IWE~\eqref{eq:IWE}).
This approach is inspired by the maximization of the entropy of the PDF of the image of warped events, as explained in Section~\ref{sec:app:Focus:Entropy}.

\section{Loss Function: Spatial Autocorrelation}
\label{sec:app:Focus:Moran}

\subsection{Moran's \emph{I} Index}

\begin{figure}[t]
\includegraphics[width=\columnwidth]{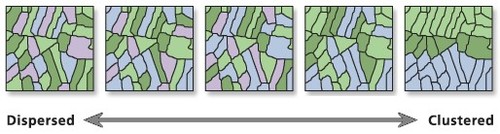}
\caption{\label{fig:MoranI}Illustration of spatial autocorrelation by Moran's index.
A negative index indicates dispersion, whereas a positive index indicates clustering.
The IWE differs from the figure above in the sense that its pixel values vary continuously with respect to the warping parameters, i.e., they are not fixed values that move around as in the figure. 
Image courtesy of ArcGIS.com \url{https://pro.arcgis.com/en/pro-app/tool-reference/spatial-statistics/spatial-autocorrelation.htm}}
\end{figure}

Moran's \emph{I} index~\eqref{eq:DefMoranIndex} (or ``serial correlation coefficient''~\cite{Moran50biom})
is a measure of spatial autocorrelation, i.e., it measures how similar is one object with respect to its neighbors. 
It is a concept that applies to variables whose values are known in unstructured grids (spatial units), in general (see Fig.~\ref{fig:MoranI}).

If the variable of interest $z$ consists of the intensity values of an image, $z_{i}=I(\bx_{i})$, which is defined on a regular (pixel) lattice $\{\bx_{i}\}$,
and the weights $w_{ij}$ are shift-invariant (they do not depend on the particular location of pixels $i$ and $j$, only on their relative spatial arrangement) and symmetric $w_{ij}=w_{ji}$, 
then it is possible to write Moran's \emph{I} index using a convolution.
In the formalism of continuous images $z(\bx)$ over a domain $\Omega$, Moran's \emph{I} index becomes
\begin{equation}
\text{Moran}(z)\doteq\frac{1}{|\Omega|}\int_{\Omega} z^{s}(\bx)\left(z^{s}(\bx)*\tilde{w}(\bx)\right)d\bx,\label{eq:MoranIcontinuousDef}
\end{equation}
where the standardized image
$z^{s}(\bx)\doteq (z(\bx)-\bar{z})/\sigma_{z}$ 
is obtained by normalizing $z$ with its mean $\bar{z}$ and variance $\sigma_{z}^{2}$ over $\Omega$.
The weights $\tilde{w}(\bx)$ should produce, in the convolution $z^{s}(\bx)*\tilde{w}(\bx)$, a sum of the neighboring values of $z^{s}(\bx)$ (excluding the central value at $\bx$). 
Thus, it is natural to consider the weights from a Gaussian kernel $G_{\sigma}(\bx)$ with a zero at the origin:
\begin{equation}
\tilde{w}(\bx)=\frac{G_{\sigma}(\bx)-G_{\sigma}(\bzero)\delta(\bx)}{1-G_{\sigma}(\bzero)}.
\label{eq:MoranWeightContNormalized}
\end{equation}

The integrand of~\eqref{eq:MoranIcontinuousDef} is the local Moran's \emph{I} index, and it is the element-wise product of the standardized variable $z^{s}$ with a low-pass filtered version of itself.
It is positive if both $z^{s}$ and neighboring values $z^{s}(\bx)*\tilde{w}(\bx)$ are higher or lower than the mean; 
and it is negative if the value and neighboring values are on opposite sides of the mean (one higher, the other lower).
Increasing event alignment corresponds to favoring negative local Moran indices (dissimilar IWE pixels next to each other), and therefore, a negative (global) Moran's \emph{I} index.

\subsection{Geary's Contiguity Ratio}

Geary's contiguity ratio is a generalization of Von Neumann's ratio~\cite{vonNeumann41ams} of the mean square successive difference (numerator of~\eqref{eq:DefGearyC}) to the variance (denominator of~\eqref{eq:DefGearyC}). 
Geary's contiguity ratio is non-negative, and its mean is 1 for random images. 
Values of $C$ significantly lower than 1 demonstrate positive spatial autocorrelation (the variable of interest is regarded as \emph{contiguous}), while values significantly higher than 1 illustrate negative autocorrelation.

In the formalism of continuous images, Geary's contiguity ratio can be written as 
\begin{equation}
\label{eq:GearyGlobalFromLocal}
C(z) = \frac{1}{2}\frac{1}{|\Omega|}\int_{\Omega}c(\bx)d\bx,
\end{equation}
with local score efficiently computed using convolutions:
\begin{equation}
\label{eq:GearyLocalConvs}
c(\bx) \doteq (z^{s}(\bx))^{2} + (z^{s}(\bx))^{2}*\tilde{w}(\bx) - 2z^{s}(\bx)\, (z^{s}*\tilde{w})(\bx).
\end{equation}
Notice that the last term in~\eqref{eq:GearyLocalConvs} also appears in Moran's \emph{I} index~\eqref{eq:MoranIcontinuousDef}. 
Thus, Geary's $C$ is inversely related to Moran's \emph{I}, but they are not identical.

Homogeneous regions of an image $z(\bx)$ have a positive spatial autocorrelation, indicated by $c(\bx) < 1$.
The regions with large values of~\eqref{eq:GearyLocalConvs}, i.e., negative spatial autocorrelation,
are those corresponding to the edges of the objects 
(dissimilar intensity values on each side of the edge).
Thus, Geary's local statistic~\eqref{eq:GearyLocalConvs} acts as an edge detector (see Figs.~\ref{fig:dynamic:localscores2} or \ref{fig:boxes:localscores2}).

\section{Loss Function: Aggregation of Local Statistics}
Similarly to~\eqref{eq:local_variance_aggr}, aggregating other local statistics of the IWE also yield focus measures.
For example, the aggregation of the local mean absolute deviation (ALMAD),
\begin{equation}
\label{eq:local_mean_abs_dev_aggr}
\text{ALMAD}(\IWE) \doteq \int_{\Omega} \text{MAD}(\bx; \IWE) d\bx,
\end{equation}
with
\begin{equation}
\label{eq:local_mean_abs_dev}
\text{MAD}(\bx; \IWE) \doteq \frac1{|B(\bx)|}\int_{B(\bx)} | \IWE(\bu;\bparams) - \mu(\bx;\IWE) | d\bu,
\end{equation}
is closely related to~\eqref{eq:local_variance_aggr} since both aggregate local measures that are edge-detectors of $\IWE$ (\eqref{eq:local_variance} uses the the $L^2$ norm, whereas \eqref{eq:local_mean_abs_dev} uses the $L^1$ norm).
Using a weighted neighborhood (e.g., Gaussian kernel $G_\sigma$), 
\eqref{eq:local_mean_abs_dev} can be efficiently approximated by the formula with two convolutions:
\begin{equation}
\label{eq:local_mean_abs_dev_conv}
\text{MAD}(\bx; \IWE) \approx \bigl| \IWE(\bx) - \bigl(\IWE(\bx) * G_\sigma (\bx)\bigr) \bigr| * G_\sigma (\bx),
\end{equation}
where the inner convolution approximates the local mean, 
$\mu(\bx;\IWE) \approx \IWE(\bx) * G_\sigma (\bx)$,
and the outer convolution averages the magnitude of the local, centered IWE (integrand of~\eqref{eq:local_mean_abs_dev}) over the neighborhood around~$\bx$.

Omitting the local mean in~\eqref{eq:local_variance} and \eqref{eq:local_mean_abs_dev} leads to local versions of the MS and the MAV, respectively.
These operators, however, are not edge detectors; nevertheless, they also work as focus loss functions since the images on which they are applied, the IWEs, are edge-like images (the events are brightness changes, i.e., they are related to the temporal derivative of the brightness signal).
Similarly to the (global) MAV, the local MAV does not provide enough information to estimate the warp parameters $\bparams$ if polarity is not used (as indicated in Table~\ref{tab:boxrot:poster:all}).

\section{Plots of the Local Loss Maps}
\global\long\def\layerwidth{0.225\linewidth}
\begin{figure*}[t!]
	\centering
    {\small
    \setlength{\tabcolsep}{2pt}
	\begin{tabular}{
	>{\centering\arraybackslash}m{\layerwidth} 
	>{\centering\arraybackslash}m{\layerwidth}
	>{\centering\arraybackslash}m{3pt} 
	>{\centering\arraybackslash}m{\layerwidth} 
	>{\centering\arraybackslash}m{\layerwidth}}
	 Scene & & & IWE before optimization & IWE after optimizing~\eqref{eq:IWEVariance} \\ [1ex]
		\frame{\includegraphics[width=\linewidth]{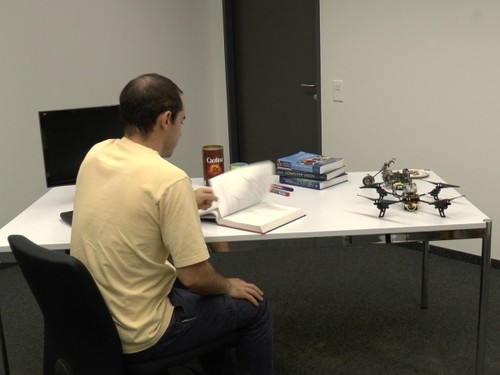}}
		&
		&&
		\frame{\includegraphics[trim={60bp 70bp 40bp 10bp},clip,width=\linewidth]{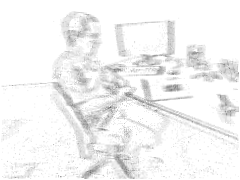}}
		&
		\frame{\includegraphics[trim={60bp 70bp 40bp 10bp},clip,width=\linewidth]{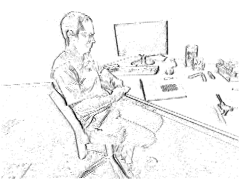}}
		\\
		\textbf{Zero parameters} & \textbf{Best parameters} $\bparams^\ast$ & & \textbf{Zero parameters} & \textbf{Best parameters} $\bparams^\ast$ \\ [1ex]
		\frame{\includegraphics[trim={60bp 70bp 40bp 10bp},clip,width=\linewidth]{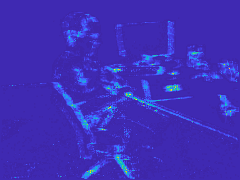}}
		&
		\frame{\includegraphics[trim={60bp 70bp 40bp 10bp},clip,width=\linewidth]{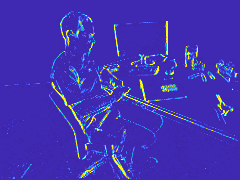}}
		&&
		\frame{\includegraphics[trim={60bp 70bp 40bp 10bp},clip,width=\linewidth]{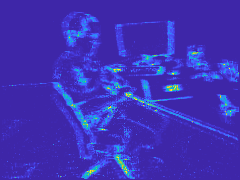}}
		&
		\frame{\includegraphics[trim={60bp 70bp 40bp 10bp},clip,width=\linewidth]{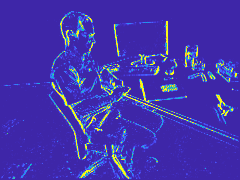}}
		\\
		\multicolumn{2}{c}{Variance~\eqref{eq:IWEVariance}: $|\IWE(\bx)-\mu_\IWE|^2$} && \multicolumn{2}{c}{MS~\eqref{eq:MeanSquareIWE}: $|\IWE(\bx)|^2$} \\[1ex]

		\frame{\includegraphics[trim={60bp 70bp 40bp 10bp},clip,width=\linewidth]{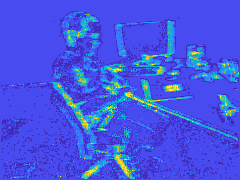}}
		&
		\frame{\includegraphics[trim={60bp 70bp 40bp 10bp},clip,width=\linewidth]{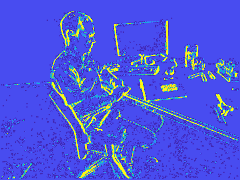}}
		&&
		\frame{\includegraphics[trim={60bp 70bp 40bp 10bp},clip,width=\linewidth]{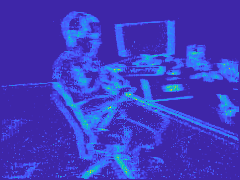}}
		&
		\frame{\includegraphics[trim={60bp 70bp 40bp 10bp},clip,width=\linewidth]{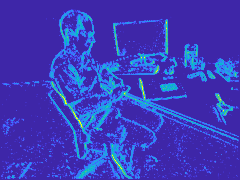}}
		\\
		\multicolumn{2}{c}{MAD~\eqref{eq:MeanAbsoluteDevIWE}: $|\IWE(\bx)-\mu_\IWE|$} && \multicolumn{2}{c}{Entropy~\eqref{eq:maxEntropy} \eqref{eq:EntropyLogPDFImageStatement}: $-\log p_{\IWE}(\IWE(\bx))$}\\[1ex]

		\frame{\includegraphics[trim={60bp 70bp 40bp 10bp},clip,width=\linewidth]{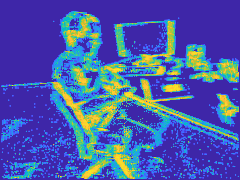}}
		&
		\frame{\includegraphics[trim={60bp 70bp 40bp 10bp},clip,width=\linewidth]{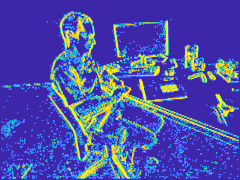}}
		&&
		\frame{\includegraphics[trim={60bp 70bp 40bp 10bp},clip,width=\linewidth]{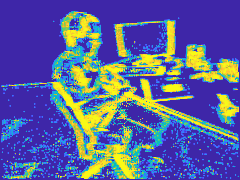}}
		&
		\frame{\includegraphics[trim={60bp 70bp 40bp 10bp},clip,width=\linewidth]{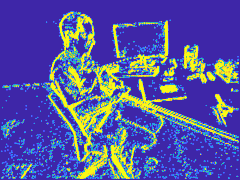}}
		\\
		\multicolumn{2}{c}{Area~\eqref{eq:minSupp} (Exponential): $F(\IWE(\bx))$} && \multicolumn{2}{c}{Area~\eqref{eq:minSupp} (Gaussian): $F(\IWE(\bx))$} \\[1ex]

		\frame{\includegraphics[trim={60bp 70bp 40bp 10bp},clip,width=\linewidth]{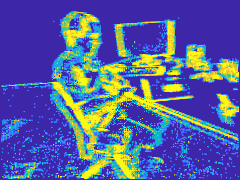}}
		&
		\frame{\includegraphics[trim={60bp 70bp 40bp 10bp},clip,width=\linewidth]{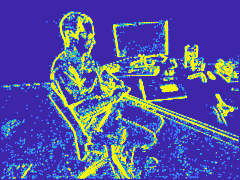}}
		&&
		\frame{\includegraphics[trim={60bp 70bp 40bp 10bp},clip,width=\linewidth]{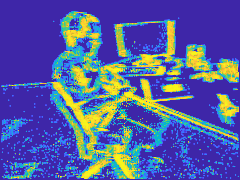}}
		&
		\frame{\includegraphics[trim={60bp 70bp 40bp 10bp},clip,width=\linewidth]{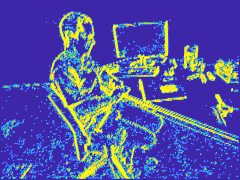}}
		\\
		\multicolumn{2}{c}{Area~\eqref{eq:minSupp} (Lorentzian): $F(\IWE(\bx))$} && \multicolumn{2}{c}{Area~\eqref{eq:minSupp} (Hyperbolic): $F(\IWE(\bx))$} \\[1ex]

		\frame{\includegraphics[trim={60bp 70bp 40bp 10bp},clip,width=\linewidth]{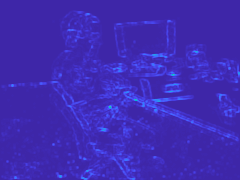}}
		&
		\frame{\includegraphics[trim={60bp 70bp 40bp 10bp},clip,width=\linewidth]{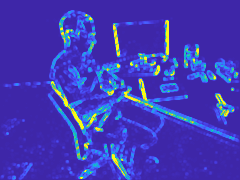}}
		&&
		\frame{\includegraphics[trim={60bp 70bp 40bp 10bp},clip,width=\linewidth]{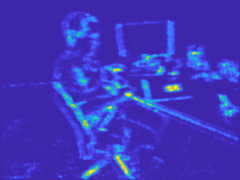}}
		&
		\frame{\includegraphics[trim={60bp 70bp 40bp 10bp},clip,width=\linewidth]{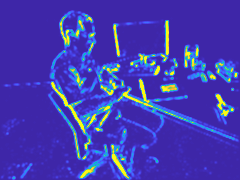}}
		\\
		\multicolumn{2}{c}{Local Variance~\eqref{eq:local_variance_conv} 
		} 
		&& \multicolumn{2}{c}{Local MS: $\IWE^2(\bx) * G_{\sigma}(\bx)$} \\[1ex]
	\end{tabular}
	}
	\caption{Visualization of the local loss (i.e., integrands of the Focus Loss Functions), pseudo-colored from blue (small) to yellow (large).
    Same scene as in the top row of Fig.~\ref{fig:evol:dyn} (i.e., without using polarity).
    Images are given in pairs: local loss before optimization (no motion compensation, Left), and after optimization of the corresponding focus loss function (motion-compensated, Right).
    The local loss of area-based loss functions is the ``support map'', as in Figs.~\ref{fig:supp:illustr:suppMap} and~\ref{fig:suppPatch}.}
	\label{fig:dynamic:localscores}
\end{figure*}

\begin{figure*}[t!]
	\centering
    {\small
    \setlength{\tabcolsep}{2pt}
	\begin{tabular}{
	>{\centering\arraybackslash}m{\layerwidth} 
	>{\centering\arraybackslash}m{\layerwidth}
	>{\centering\arraybackslash}m{3pt} 
	>{\centering\arraybackslash}m{\layerwidth} 
	>{\centering\arraybackslash}m{\layerwidth}}
	 \textbf{Zero parameters} & \textbf{Best parameters} $\bparams^\ast$ & & \textbf{Zero parameters} & \textbf{Best parameters} $\bparams^\ast$ \\ [1ex]
     
		\frame{\includegraphics[trim={60bp 70bp 40bp 10bp},clip,width=\linewidth]{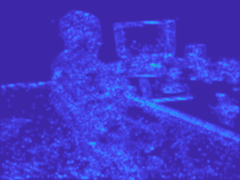}}
		&
		\frame{\includegraphics[trim={60bp 70bp 40bp 10bp},clip,width=\linewidth]{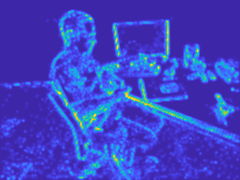}}
		&&
		\frame{\includegraphics[trim={60bp 70bp 40bp 10bp},clip,width=\linewidth]{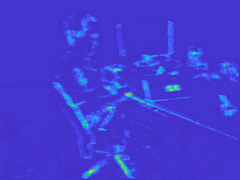}}
		&
		\frame{\includegraphics[trim={60bp 70bp 40bp 10bp},clip,width=\linewidth]{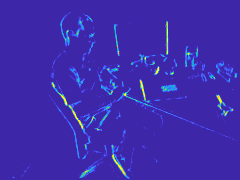}}
		\\
		\multicolumn{2}{c}{Local MAD~\eqref{eq:local_mean_abs_dev_conv}: $|\IWE(\bx) - (\IWE(\bx)*G_{\sigma}(\bx))|*G_{\sigma}(\bx)$} 
		&& \multicolumn{2}{c}{(Local) Moran's Index~\eqref{eq:DefMoranIndex},\eqref{eq:MoranIcontinuousDef}:
		$\IWE^{s}(\bx)\,(\IWE^{s}*\tilde{w})(\bx)$
		} \\[1ex]
        
		\frame{\includegraphics[trim={60bp 70bp 40bp 10bp},clip,width=\linewidth]{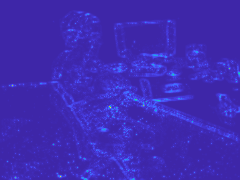}}
		&
		\frame{\includegraphics[trim={60bp 70bp 40bp 10bp},clip,width=\linewidth]{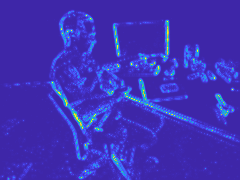}}
		&&
		\frame{\includegraphics[trim={60bp 70bp 40bp 10bp},clip,width=\linewidth]{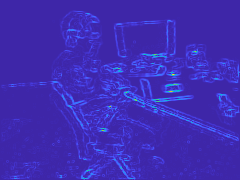}}
		&
		\frame{\includegraphics[trim={60bp 70bp 40bp 10bp},clip,width=\linewidth]{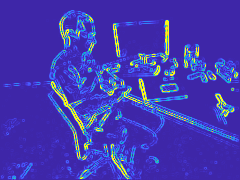}}
		\\
		\multicolumn{2}{c}{(Local) Geary's $C$ Contiguity Ratio~\eqref{eq:GearyLocalConvs}} && \multicolumn{2}{c}{Gradient Magnitude~\eqref{eq:maxGradient}: $\|\nabla \IWE(\bx)\|^2$} \\[1ex]

		\frame{\includegraphics[trim={60bp 70bp 40bp 10bp},clip,width=\linewidth]{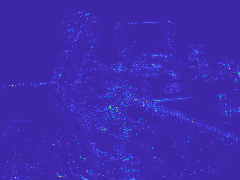}}
		&
		\frame{\includegraphics[trim={60bp 70bp 40bp 10bp},clip,width=\linewidth]{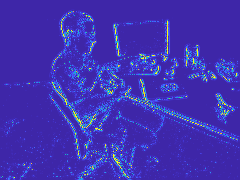}}
		&&
		\frame{\includegraphics[trim={60bp 70bp 40bp 10bp},clip,width=\linewidth]{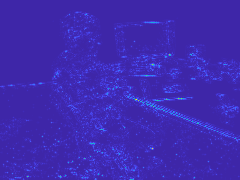}}
		&
		\frame{\includegraphics[trim={60bp 70bp 40bp 10bp},clip,width=\linewidth]{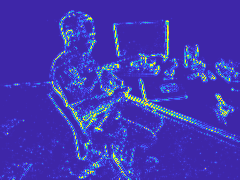}}
		\\
		\multicolumn{2}{c}{Laplacian Magnitude~\eqref{eq:maxLaplacian} $\|\Delta\IWE(\bx)\|^2$} && \multicolumn{2}{c}{Hessian Magnitude~\eqref{eq:maxHessian}: $\|\text{Hess}(\IWE(\bx))\|^2$} \\[1ex]

		\frame{\includegraphics[trim={60bp 70bp 40bp 10bp},clip,width=\linewidth]{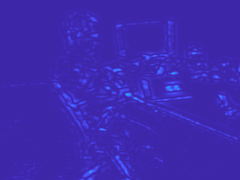}}
		&
		\frame{\includegraphics[trim={60bp 70bp 40bp 10bp},clip,width=\linewidth]{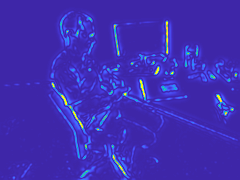}}
		&&
		\frame{\includegraphics[trim={60bp 70bp 40bp 10bp},clip,width=\linewidth]{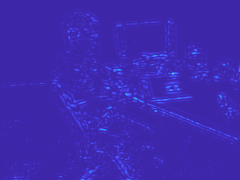}}
		&
		\frame{\includegraphics[trim={60bp 70bp 40bp 10bp},clip,width=\linewidth]{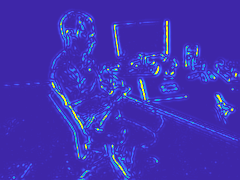}}
		\\
		\multicolumn{2}{c}{DoG: $|(\IWE * G_{\sigma_1})(\bx) - (\IWE * G_{\sigma_2})(\bx)|^2$} && 
		\multicolumn{2}{c}{LoG: $|(\IWE * G_{\sigma})(\bx) - (\IWE * G_{1.6\sigma})(\bx)|^2$} \\[1ex]

		\frame{\includegraphics[trim={60bp 70bp 40bp 10bp},clip,width=\linewidth]{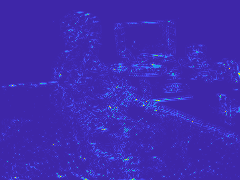}}
		&
		\frame{\includegraphics[trim={60bp 70bp 40bp 10bp},clip,width=\linewidth]{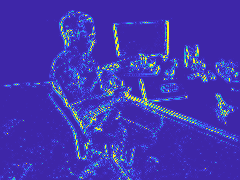}}
		&&
		\frame{\includegraphics[trim={60bp 70bp 40bp 10bp},clip,width=\linewidth]{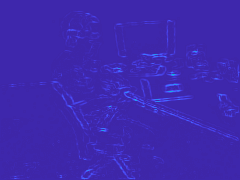}}
		&
		\frame{\includegraphics[trim={60bp 70bp 40bp 10bp},clip,width=\linewidth]{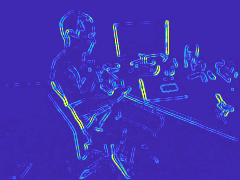}}
		\\
		\multicolumn{2}{c}{Variance of Laplacian: $|\Delta\IWE(\bx) - \mu_{\Delta\IWE} |^2$} && 
		\multicolumn{2}{c}{Variance of Gradient Magnitude: $|\|\nabla\IWE(\bx)\| - \mu_{\|\nabla\IWE\|}|^2$} \\[1ex]

		\frame{\includegraphics[trim={60bp 70bp 40bp 10bp},clip,width=\linewidth]{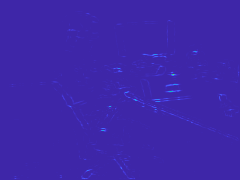}}
		&
		\frame{\includegraphics[trim={60bp 70bp 40bp 10bp},clip,width=\linewidth]{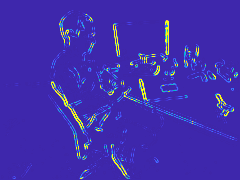}}
		&&
		\frame{\includegraphics[trim={60bp 70bp 40bp 10bp},clip,width=\linewidth]{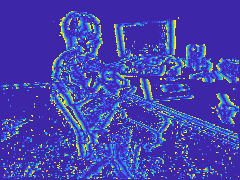}}
		&
		\frame{\includegraphics[trim={60bp 70bp 40bp 10bp},clip,width=\linewidth]{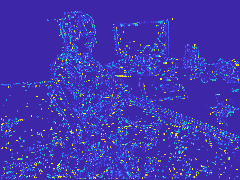}}
		\\
		\multicolumn{2}{c}{Variance of Squared Gradient Magnitude} && 
		\multicolumn{2}{c}{Variance of Mean Timestamp on Pixel~\cite{Mitrokhin18iros}} \\[1ex]
	\end{tabular}
	}
	\caption{Visualization of local scores (i.e., integrands) of the Focus Loss Functions (continuation).}
	\label{fig:dynamic:localscores2}
\end{figure*}

\global\long\def\layerwidth{0.225\linewidth}
\begin{figure*}[t!]
	\centering
    {\small
    \setlength{\tabcolsep}{2pt}
	\begin{tabular}{
	>{\centering\arraybackslash}m{\layerwidth} 
	>{\centering\arraybackslash}m{\layerwidth}
	>{\centering\arraybackslash}m{3pt} 
	>{\centering\arraybackslash}m{\layerwidth} 
	>{\centering\arraybackslash}m{\layerwidth}}
	 Scene & & & IWE before optimization & IWE after optimizing~\eqref{eq:IWEVariance} \\ [1ex]
		\frame{\includegraphics[width=\linewidth]{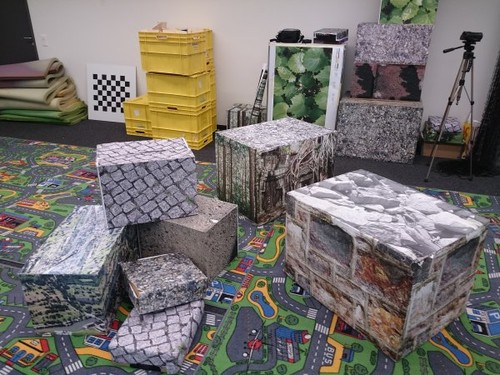}}
		&
		&&
		\frame{\includegraphics[width=\linewidth]{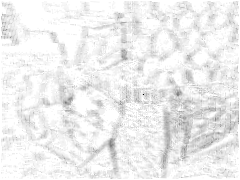}}
		&
		\frame{\includegraphics[width=\linewidth]{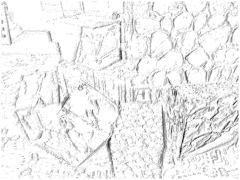}}
		\\	
	 \textbf{Zero parameters} & \textbf{Best parameters} $\bparams^\ast$ & & \textbf{Zero parameters} & \textbf{Best parameters} $\bparams^\ast$ \\ [1ex]
		\frame{\includegraphics[width=\linewidth]{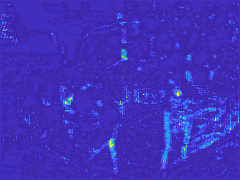}}
		&
		\frame{\includegraphics[width=\linewidth]{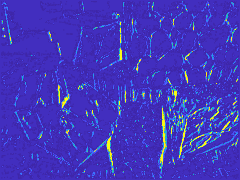}}
		&&
		\frame{\includegraphics[width=\linewidth]{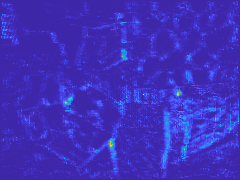}}
		&
		\frame{\includegraphics[width=\linewidth]{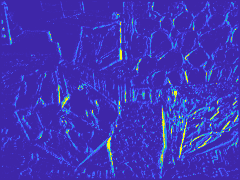}}
		\\
		\multicolumn{2}{c}{Variance~\eqref{eq:IWEVariance}: $|\IWE(\bx)-\mu_\IWE|^2$} && \multicolumn{2}{c}{MS~\eqref{eq:MeanSquareIWE}: $|\IWE(\bx)|^2$} \\[1ex]

		\frame{\includegraphics[width=\linewidth]{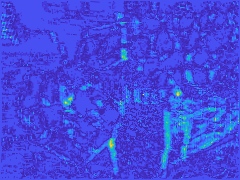}}
		&
		\frame{\includegraphics[width=\linewidth]{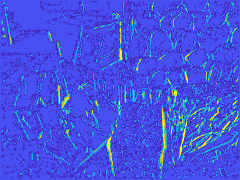}}
		&&
		\frame{\includegraphics[width=\linewidth]{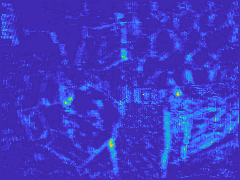}}
		&
		\frame{\includegraphics[width=\linewidth]{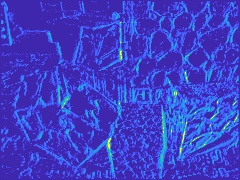}}
		\\
		\multicolumn{2}{c}{MAD~\eqref{eq:MeanAbsoluteDevIWE}: $|\IWE(\bx)-\mu_\IWE|$} && \multicolumn{2}{c}{Entropy~\eqref{eq:maxEntropy} \eqref{eq:EntropyLogPDFImageStatement}: $-\log p_{\IWE}(\IWE(\bx))$}\\[1ex]

		\frame{\includegraphics[width=\linewidth]{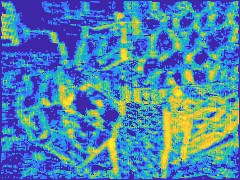}}
		&
		\frame{\includegraphics[width=\linewidth]{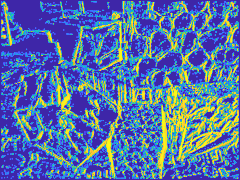}}
		&&
		\frame{\includegraphics[width=\linewidth]{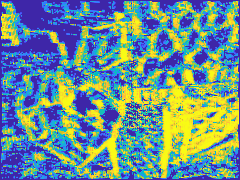}}
		&
		\frame{\includegraphics[width=\linewidth]{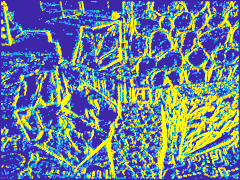}}
		\\
		\multicolumn{2}{c}{Area~\eqref{eq:minSupp} (Exponential): $F(\IWE(\bx))$} && \multicolumn{2}{c}{Area~\eqref{eq:minSupp} (Gaussian): $F(\IWE(\bx))$} \\[1ex]

		\frame{\includegraphics[width=\linewidth]{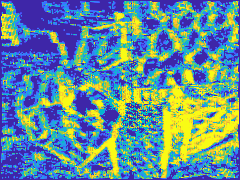}}
		&
		\frame{\includegraphics[width=\linewidth]{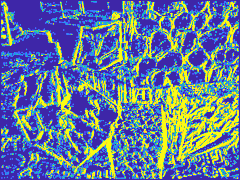}}
		&&
		\frame{\includegraphics[width=\linewidth]{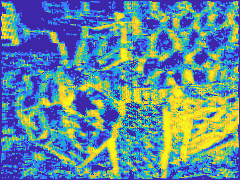}}
		&
		\frame{\includegraphics[width=\linewidth]{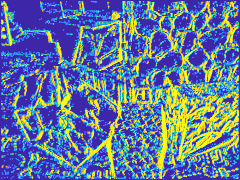}}
		\\
		\multicolumn{2}{c}{Area~\eqref{eq:minSupp} (Lorentzian): $F(\IWE(\bx))$} && \multicolumn{2}{c}{Area~\eqref{eq:minSupp} (Hyperbolic): $F(\IWE(\bx))$} \\[1ex]

		\frame{\includegraphics[width=\linewidth]{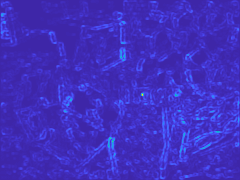}}
		&
		\frame{\includegraphics[width=\linewidth]{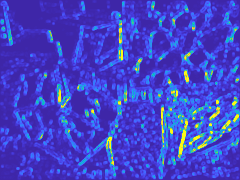}}
		&&
		\frame{\includegraphics[width=\linewidth]{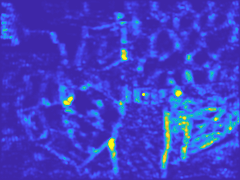}}
		&
		\frame{\includegraphics[width=\linewidth]{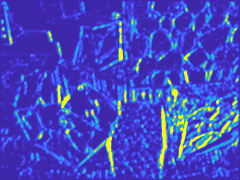}}
		\\
		\multicolumn{2}{c}{Local Variance~\eqref{eq:local_variance_conv}} && \multicolumn{2}{c}{Local MS: $\IWE^2(\bx) * G_{\sigma}(\bx)$} \\[1ex]
	\end{tabular}
	}
	\caption{Visualization of the local loss (i.e., integrands of the Focus Loss Functions).
    Scene {\small\texttt{boxes}}, IWE not without using polarity.
    Same notation as Fig.~\ref{fig:dynamic:localscores}.}
	\label{fig:boxes:localscores}
\end{figure*}

\begin{figure*}[t!]
	\centering
    {\small
    \setlength{\tabcolsep}{2pt}
	\begin{tabular}{
	>{\centering\arraybackslash}m{\layerwidth} 
	>{\centering\arraybackslash}m{\layerwidth}
	>{\centering\arraybackslash}m{3pt} 
	>{\centering\arraybackslash}m{\layerwidth} 
	>{\centering\arraybackslash}m{\layerwidth}}
	 \textbf{Zero parameters} & \textbf{Best parameters} $\bparams^\ast$ & & \textbf{Zero parameters} & \textbf{Best parameters} $\bparams^\ast$ \\ [1ex]
     
		\frame{\includegraphics[width=\linewidth]{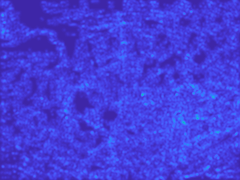}}
		&
		\frame{\includegraphics[width=\linewidth]{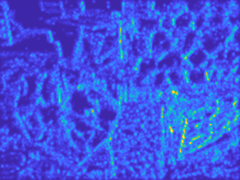}}
		&&
		\frame{\includegraphics[width=\linewidth]{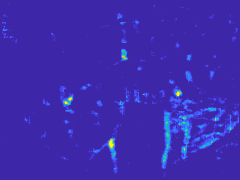}}
		&
		\frame{\includegraphics[width=\linewidth]{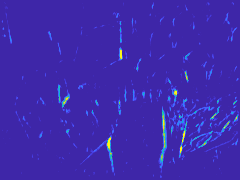}}
		\\
		\multicolumn{2}{c}{Local MAD~\eqref{eq:local_mean_abs_dev_conv}: $|\IWE(\bx) - (\IWE(\bx)*G_{\sigma}(\bx))|*G_{\sigma}(\bx)$} 
		&& \multicolumn{2}{c}{(Local) Moran's Index~\eqref{eq:DefMoranIndex}, \eqref{eq:MoranIcontinuousDef}:
		$\IWE^{s}(\bx)\,(\IWE^{s}*\tilde{w})(\bx)$} \\[1ex]
        
		\frame{\includegraphics[width=\linewidth]{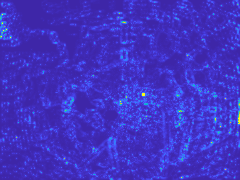}}
		&
		\frame{\includegraphics[width=\linewidth]{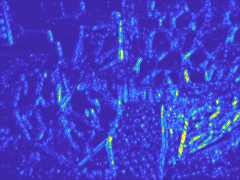}}
		&&
		\frame{\includegraphics[width=\linewidth]{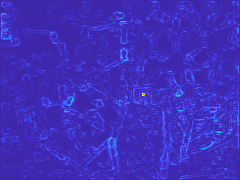}}
		&
		\frame{\includegraphics[width=\linewidth]{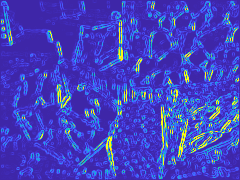}}
		\\
		\multicolumn{2}{c}{(Local) Geary's $C$ Contiguity Ratio~\eqref{eq:GearyLocalConvs}} && \multicolumn{2}{c}{Gradient Magnitude~\eqref{eq:maxGradient}: $\|\nabla \IWE(\bx)\|^2$} \\[1ex]

		\frame{\includegraphics[width=\linewidth]{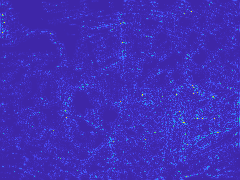}}
		&
		\frame{\includegraphics[width=\linewidth]{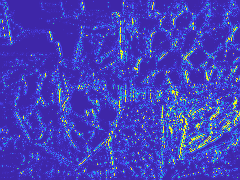}}
		&&
		\frame{\includegraphics[width=\linewidth]{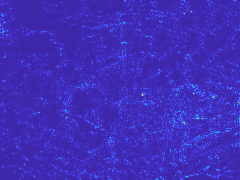}}
		&
		\frame{\includegraphics[width=\linewidth]{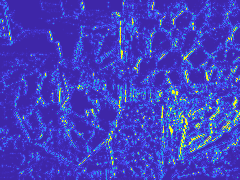}}
		\\
		\multicolumn{2}{c}{Laplacian Magnitude~\eqref{eq:maxLaplacian} $\|\Delta\IWE(\bx)\|^2$} && \multicolumn{2}{c}{Hessian Magnitude~\eqref{eq:maxHessian}: $\|\text{Hess}(\IWE(\bx))\|^2$} \\[1ex]

		\frame{\includegraphics[width=\linewidth]{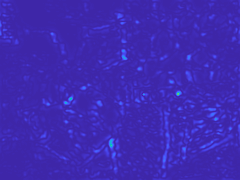}}
		&
		\frame{\includegraphics[width=\linewidth]{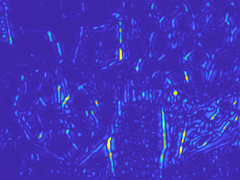}}
		&&
		\frame{\includegraphics[width=\linewidth]{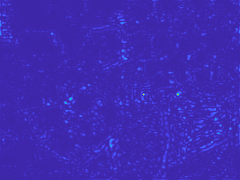}}
		&
		\frame{\includegraphics[width=\linewidth]{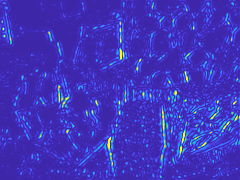}}
		\\
		\multicolumn{2}{c}{DoG: $|(\IWE * G_{\sigma_1})(\bx) - (\IWE * G_{\sigma_2})(\bx)|^2$} && 
		\multicolumn{2}{c}{LoG: $|(\IWE * G_{\sigma})(\bx) - (\IWE * G_{1.6\sigma})(\bx)|^2$} \\[1ex]

		\frame{\includegraphics[width=\linewidth]{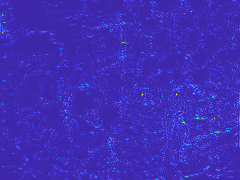}}
		&
		\frame{\includegraphics[width=\linewidth]{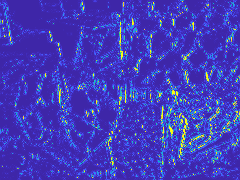}}
		&&
		\frame{\includegraphics[width=\linewidth]{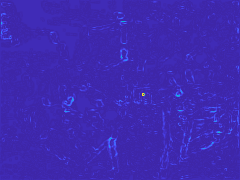}}
		&
		\frame{\includegraphics[width=\linewidth]{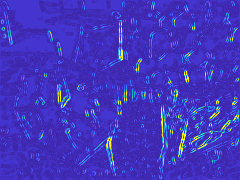}}
		\\
		\multicolumn{2}{c}{Variance of Laplacian: $|\Delta\IWE(\bx) - \mu_{\Delta\IWE} |^2$} && 
		\multicolumn{2}{c}{Variance of Gradient Magnitude: $|\|\nabla\IWE(\bx)\| - \mu_{\|\nabla\IWE\|}|^2$} \\[1ex]

		\frame{\includegraphics[width=\linewidth]{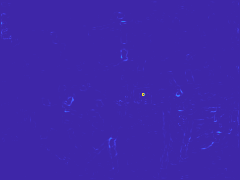}}
		&
		\frame{\includegraphics[width=\linewidth]{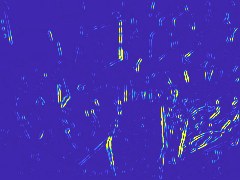}}
		&&
		\frame{\includegraphics[width=\linewidth]{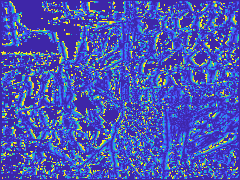}}
		&
		\frame{\includegraphics[width=\linewidth]{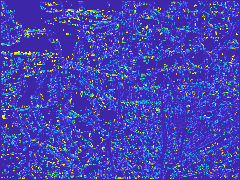}}
		\\
		\multicolumn{2}{c}{Variance of Squared Gradient Magnitude} && 
		\multicolumn{2}{c}{Variance of Mean Timestamp on Pixel~\cite{Mitrokhin18iros}} \\[1ex]
	\end{tabular}
	}
	\caption{Visualization of local scores (i.e., integrands) of the Focus Loss Functions (continuation).}
	\label{fig:boxes:localscores2}
\end{figure*}

Most of the loss functions considered can be written as integrals over the IWE domain.
Figs.~\ref{fig:dynamic:localscores},
\ref{fig:dynamic:localscores2},
\ref{fig:boxes:localscores} and 
\ref{fig:boxes:localscores2} visualize the integrands (i.e., ``local loss'') of most focus loss functions,
for two scenes: {\small\texttt{dynamic}} and {\small\texttt{boxes}} from the dataset~\cite{Mueggler17ijrr}.
Images are given in pairs (with a common caption below the images): 
local loss before optimization (without motion compensation, on the left), 
and after optimization of the corresponding focus loss function (motion-compensated, on the right).
Each image pair shares the same color scale for proper visualization of how the local loss changes before and after optimization.
For reference, since the local loss are transformations of the IWE, Figs.~\ref{fig:dynamic:localscores} and~\ref{fig:boxes:localscores} also provide, on the top right, the IWE before and after optimization with one of the loss functions (the variance).

The local loss of area-based loss functions is the support map, as in Figs.~\ref{fig:supp:illustr:suppMap} and~\ref{fig:suppPatch}. 
The focus loss given by the IWE range~\eqref{eq:SupportPDFSimplified} is not expressed as an integral over the image domain, therefore, no image integrand is visualized in the above-mentioned figures. 
MAV and local MAV are not displayed either since they cannot be optimized with respect to the parameters.

Notice that all local loss maps are represented using the same color scheme, from blue (small) to large (yellow).
For objective functions formulated as maximization problems (variance, gradient magnitude, etc.), 
visually good maps are those that are almost ``blue'' for IWEs with bad event-alignment parameters, 
and that clearly show ``yellow'' regions where events align (due to good parameters $\bparams$).
For objective functions formulated as minimization problems (e.g., area-based, loss based on the mean timestamp per pixel), the situation is the opposite: 
good local loss maps become less yellowish and more blueish as event alignment improves due to good parameters $\bparams$.

\section{Additional Experiments on Accuracy Evaluation}
\begin{table*}
\centering
\begin{adjustbox}{max width=\textwidth}
\begin{tabular}{l|cccccc|cccccc}
\toprule 
\textbf{Focus Loss Function}  & \multicolumn{6}{c|}{\textbf{Without polarity}} & \multicolumn{6}{c}{\textbf{With polarity}}\\
 & $\omega_{x}$  & $\omega_{y}$  & $\omega_{z}$  & $\mu$  & $\sigma$  & \textbf{RMS}  & $\omega_{x}$  & $\omega_{y}$  & $\omega_{z}$ & $\mu$  & $\sigma$  & \textbf{RMS} \\
\midrule 
Variance \eqref{eq:IWEVariance} \cite{Gallego18cvpr,Gallego17ral}  & 15.69  & 19.53  & 20.34  & -0.31  & 18.42  & 18.52  & 16.03  & 19.67  & 21.12 & -0.64  & 18.78  & 18.94 \\
Mean Square \eqref{eq:MeanSquareIWE} \cite{Gallego17ral,Stoffregen17acra} & 16.05  & 20.03  & 23.70  & -0.49  & 19.83  & 19.93  & 16.04  & 19.67  & 21.36 & -0.57  & 18.86  & 19.02 \\
Mean Absolute Deviation \eqref{eq:MeanAbsoluteDevIWE}  & 15.40  & \textbf{18.99}  & 23.99  & -0.49  & 19.40  & 19.46  & 15.93  & 19.15  & 23.67 & -0.57  & 19.40  & 19.58 \\
Mean Absolute Value \eqref{eq:MeanAbsoluteValIWE}  & -  & -  & -  & -  & -  & -  & 15.95  & 19.16  & 24.22 & -0.61  & 19.60  & 19.77 \\
Entropy \eqref{eq:maxEntropy}  & 19.29  & 21.73  & 44.48  & 0.03  & 28.14  & 28.50  & 18.03  & 20.50  & 41.10 & -0.20  & 26.19  & 26.54 \\
Area \eqref{eq:minSupp} (Exp)  & 19.70  & 21.57  & 53.23  & 0.10  & 31.41  & 31.50  & 15.89  & 19.12  & 23.60 & -0.27  & 19.39  & 19.54 \\
Area \eqref{eq:minSupp} (Gaussian)  & 18.12  & 20.20  & 39.22  & 0.10  & 25.78  & 25.85  & 15.67  & \textbf{18.77}  & 22.11 & -0.33  & 18.69  & 18.85 \\
Area \eqref{eq:minSupp} (Lorentzian)  & 20.56  & 21.69  & 55.04  & 0.42  & 32.30  & 32.43  & 15.48  & 19.36  & 28.11 & -0.02  & 20.85  & 20.98 \\
Area \eqref{eq:minSupp} (Hyperbolic)  & 18.71  & 20.67  & 47.99  & 0.07  & 29.05  & 29.13  & 15.69  & 18.90  & 22.86 & -0.27  & 18.99  & 19.15 \\
Range \eqref{eq:SupportPDFSimplified} (Exp)  & 18.68 & 22.72  & 44.59 & 0.39  & 28.16  & 28.66  & 17.29 & 19.38 & 49.50 & -0.48 & 28.23 & 28.72 \\ %
Local Variance~\eqref{eq:local_variance_aggr}  & 16.17  & 19.55  & 18.90  & -0.41  & 17.94  & 18.21  & 16.04  & 19.55  & 19.61 & -0.21  & 18.15  & 18.40 \\
Local Mean Square  & 18.37  & 21.42  & 34.65  & 0.59  & 24.31  & 24.81  & 16.68  & 20.21  & 22.67 & -0.05  & 19.53  & 19.86 \\
Local Mean Absolute Deviation  & 16.36  & 19.39  & 28.35  & -0.35  & 21.10  & 21.37  & 15.78  & 19.06  & 21.37  & -0.20  & 18.47  & 18.74 \\
Local Mean Absolute Value  & -  & -  & -  & -  & -  & -  & 18.13  & 20.17  & 34.00  & -0.42  & 23.61  & 24.10 \\
Moran's Index~\eqref{eq:DefMoranIndex}  & 17.87  & 20.74  & 34.24  & -0.99  & 23.87  & 24.28  & 16.97  & 20.01  & 33.31  & -0.08  & 23.18  & 23.43 \\
Geary's Contiguity Ratio~\eqref{eq:DefGearyC}  & 17.48  & 20.21  & 33.93  & -0.58  & 23.50  & 23.87  & 16.00  & 19.42  & 23.07  & 0.04  & 19.25  & 19.50 \\
Gradient Magnitude \eqref{eq:maxGradient}  & 16.12  & 19.53  & \textbf{17.84}  & -0.71  & \textbf{17.58}  & \textbf{17.83}  & 15.90  & 19.46  & 18.93 & -0.79  & 17.91  & 18.10 \\
Laplacian Magnitude \eqref{eq:maxLaplacian}  & 15.63  & 20.92  & 18.42  & -0.01  & 18.09  & 18.32  & 14.52  & 19.93  & \textbf{18.28} & \textbf{0.03}  & \textbf{17.36} & \textbf{17.58} \\
Hessian Magnitude \eqref{eq:maxHessian}  & 16.26  & 19.69  & 19.29  & -0.22  & 18.14  & 18.41  & 15.89  & 19.43  & 18.46 & -0.30  & 17.70  & 17.93 \\
Difference of Gaussians (DoG)  & 14.79  & 20.22  & 27.54  & 0.66  & 20.50  & 20.85  & 14.99  & 20.52  & 22.25 & 0.43  & 18.92  & 19.25 \\
Laplacian of the Gaussian (LoG)  & \textbf{14.64}  & 20.45  & 26.00  & 0.50  & 20.09  & 20.36  & \textbf{14.42}  & 20.09  & 18.80 & 0.47  & 17.49  & 17.77 \\
Variance of Laplacian  & 16.30  & 19.70  & 18.78  & -0.45  & 18.02  & 18.26  & 15.95  & 19.45  & 18.71 & -0.32  & 17.81  & 18.01 \\
Variance of Gradient  & 16.26  & 19.71  & 20.11  & \textbf{-0.00}  & 18.49  & 18.69  & 16.10  & 19.75  & 21.39 & 0.31  & 18.84  & 19.08 \\
Variance of Squared Gradient  & 16.50  & 20.02  & 19.64  & -0.10  & 18.46  & 18.72  & 16.22  & 20.08  & 20.56 & 0.06  & 18.72  & 18.95 \\
Mean Timestamp on Pixel~\cite{Mitrokhin18iros}  & 42.48  & 39.40  & 166.81  & 0.54  & 82.87  & 82.89  & -  & -  & -  & -  & -  & - \\
\bottomrule
\end{tabular}
\end{adjustbox}
\vspace{-0.5ex}
\caption{\label{tab:rot:boxes:all}\emph{Accuracy and Timing Comparison of Focus Loss Functions}.
Angular velocity errors (in \si{\deg/\second}) of the motion compensation method~\cite{Gallego18cvpr} (with or without polarity) with respect to motion-capture system. 
The six columns per case are the errors in each component of the angular velocity and their mean, standard deviation and RMS values.
Processing $\numEvents = \SI{30000}{events}$, warped onto an image of $240\times 180$ pixels (DAVIS camera~\cite{Brandli14ssc}).
Sequence: \texttt{\small{}boxes\_rotation} from the Event Camera Dataset~\cite{Mueggler17ijrr}.
Best value per column is in bold.}
\end{table*}

\begin{table*}
\centering
\begin{adjustbox}{max width=\textwidth}
\begin{tabular}{l|cccccc|cccccc}
\toprule 
\textbf{Focus Loss Function}  & \multicolumn{6}{c|}{\textbf{Without polarity}} & \multicolumn{6}{c}{\textbf{With polarity}}\\
 & $\omega_{x}$  & $\omega_{y}$  & $\omega_{z}$  & $\mu$  & $\sigma$  & \textbf{RMS}  & $\omega_{x}$  & $\omega_{y}$  & $\omega_{z}$ & $\mu$  & $\sigma$  & \textbf{RMS} \\
\midrule 
Variance \eqref{eq:IWEVariance} \cite{Gallego18cvpr,Gallego17ral}  & 26.67  & 20.24  & 30.95  & \textbf{0.45}  & 25.93  & 25.96  & 26.50  & 19.86  & 26.80 & \textbf{-0.20}  & 24.36  & 24.39 \\
Mean Square \eqref{eq:MeanSquareIWE} \cite{Gallego17ral,Stoffregen17acra}  & 28.44  & 25.89  & 47.97  & 3.70  & 33.43  & 34.10  & 26.90  & 21.23  & 30.79 & 1.19  & 26.14  & 26.31 \\
Mean Absolute Deviation \eqref{eq:MeanAbsoluteDevIWE}  & 27.25  & 22.16  & 42.66  & 1.25  & 30.43  & 30.70  & 27.11  & 21.73  & 40.03 & 1.71  & 29.40  & 29.62\\
Mean Absolute Value \eqref{eq:MeanAbsoluteValIWE} & -  & -  & -  & -  & -  & -  & 27.18  & 22.19  & 40.33 & 2.18  & 29.65  & 29.90 \\
Entropy \eqref{eq:maxEntropy}  & 32.81  & 41.12  & 68.70  & 3.63  & 46.98  & 47.54  & 27.65  & 22.73  & 49.27 & 2.36  & 32.98  & 33.21 \\
Area \eqref{eq:minSupp} (Exp)  & 33.12  & 22.50  & 73.75  & -0.73  & 42.94  & 43.12  & 26.31  & 19.53  & 33.36  & -0.42  & 26.36  & 26.40 \\
Area \eqref{eq:minSupp} (Gaussian)  & 28.40  & 21.07  & 54.02  & -0.59  & 34.36  & 34.50  & 26.15 & 19.20 & 30.70 & -0.41  & 25.31  & 25.35 \\
Area \eqref{eq:minSupp} (Lorentzian)  & 28.51  & 21.69  & 57.41  & -0.92  & 35.71  & 35.86  & 26.30  & 19.63  & 33.77  & -0.59  & 26.52  & 26.57 \\
Area \eqref{eq:minSupp} (Hyperbolic)  & 27.39  & 20.65  & 50.78  & -0.84  & 32.82  & 32.94  & 26.27  & 19.35  & 32.01  & -0.41  & 25.84  & 25.88 \\
Range \eqref{eq:SupportPDFSimplified} (Exp) & 32.18 & 37.73 & 61.70 & 3.44 & 43.32 & 43.87 & 24.48 & 19.53 & 45.99 & 1.56 & 29.86 & 30.00\\ %
Local Variance~\eqref{eq:local_variance_aggr}  & 26.95  & 20.49  & 28.87  & 1.11  & 25.34  & 25.44  & 26.48  & 20.32  & 25.66  & 0.88  & 24.06  & 24.15 \\
Local Mean Square  & 28.47  & 25.58  & 47.79  & 3.79  & 33.28  & 33.95  & 26.74  & 21.25  & 31.41  & 1.26  & 26.30  & 26.47 \\
Local Mean Absolute Deviation  & 47.61  & 53.19  & 84.85  & 4.55  & 61.67  & 61.89  & 26.42  & 20.03  & 29.41  & 0.95  & 25.22  & 25.29\\
Local Mean Absolute Value  & -  & -  & -  & -  & -  & -  & 27.19  & 22.34  & 41.57  & 2.22  & 30.10  & 30.37 \\
Moran's Index~\eqref{eq:DefMoranIndex}  & 28.32  & 21.57  & 47.31  & 1.39  & 32.32  & 32.40  & 27.70  & 20.99  & 44.18 & 1.32  & 30.90  & 30.96\\
Geary's Contiguity Ratio~\eqref{eq:DefGearyC}  & 27.63  & 20.78  & 31.44  & 0.89  & 26.56  & 26.61  & 26.52  & 20.25  & 28.90 & 1.19  & 25.15  & 25.23 \\
Gradient Magnitude \eqref{eq:maxGradient}  & 26.61  & 20.50  & \textbf{24.69} & 0.61  & \textbf{23.85} & \textbf{23.93}  & 26.34  & 20.01  & 24.40 & \textbf{-0.20} & 23.54 & 23.58\\
Laplacian Magnitude \eqref{eq:maxLaplacian}  & 27.24  & 20.74  & 26.76  & 0.64  & 24.85  & 24.91  & 26.29  & 20.25  & 24.46 & 0.61  & 23.60  & 23.67\\
Hessian Magnitude \eqref{eq:maxHessian}  & 27.20  & 22.40  & 26.81  & 0.95  & 25.39  & 25.47  & 26.36  & 20.24  & 24.61  & 0.80  & 23.66  & 23.74\\
Difference of Gaussians (DoG) & \textbf{24.51} & 18.01 & 30.97 & 1.45 & 24.41 & 24.50 & 24.29 & 17.85 & 24.30 & 0.75 & \textbf{22.10} & \textbf{22.15}\\
Laplacian of the Gaussian (LoG) & 24.66 & \textbf{17.96} & 32.85 & 1.27 & 25.10 & 25.15 & \textbf{24.20} & \textbf{17.78} & 30.04 & 1.30 & 23.94 & 24.01 \\
Variance of Laplacian  & 26.85  & 27.84  & 25.10  & 0.77  & 26.52  & 26.59  & 26.26  & 20.23  & \textbf{24.38}  & 0.53  & 23.56  & 23.62\\
Variance of Gradient  & 26.85  & 27.84  & 25.10  & 0.77  & 26.52  & 26.60  & 26.63  & 20.29  & 25.44 & 0.96  & 24.12  & 24.22\\
Variance of Squared Gradient  & 26.80  & 21.14  & 30.36  & 1.29  & 25.97  & 26.10  & 26.72  & 20.94  & 25.63 & 1.13  & 24.32  & 24.43\\
Mean Timestamp on Pixel~\cite{Mitrokhin18iros}  & 64.94  & 87.19  & 211.4  & 0.94  & 121.1  & 121.2  & -  & -  & - & -  & -  & -\\
\bottomrule
\end{tabular}
\end{adjustbox}
\vspace{-0.5ex}
 \caption{\label{tab:rot:poster:all} \emph{Accuracy Comparison of Focus Loss
Functions} on the \texttt{\small{}{}poster\_rotation} sequence from dataset~\cite{Mueggler17ijrr}. 
Angular velocity errors (in \si{\deg/\second}) of the motion compensation method~\cite{Gallego18cvpr} (with or without polarity) with respect to motion-capture system.
The six columns per case are the errors in each component of the angular velocity and their mean, standard deviation and RMS values.
Processing $\numEvents = \SI{30000}{events}$, warped onto an image of $240\times 180$ pixels (DAVIS camera~\cite{Brandli14ssc}).
On each column, the best value is highlighted in bold.}
\end{table*}

Tables~\ref{tab:rot:boxes:all} and~\ref{tab:rot:poster:all} provide further quantitative evaluations of the considered focus loss functions.
Results correspond to the {\small\texttt{boxes}} and {\small\texttt{poster}} sequences in~\cite{Mueggler17ijrr}, undergoing a rotational motion with velocities close to \SI{1000}{\degree/\second}.
Looking at the RMS errors, these are small compared to the excursion of the signal.
The RMS columns of these tables are summarized in Table~\ref{tab:boxrot:poster:all}.

\section{Additional Plots of Focus Loss Functions in Optical Flow Space}
\global\long\def\oflowsurfwidth{0.175\linewidth}
\begin{figure*}[t!]
	\centering
    {\small
    \setlength{\tabcolsep}{2pt}
	\begin{tabular}{
	>{\centering\arraybackslash}m{\oflowsurfwidth} 
	>{\centering\arraybackslash}m{\oflowsurfwidth}
	>{\centering\arraybackslash}m{\oflowsurfwidth} 
	>{\centering\arraybackslash}m{\oflowsurfwidth}}
		\includegraphics[width=\linewidth]{images/optical_flow/cvpr18/preview.pdf}
		&
		\includegraphics[width=\linewidth]{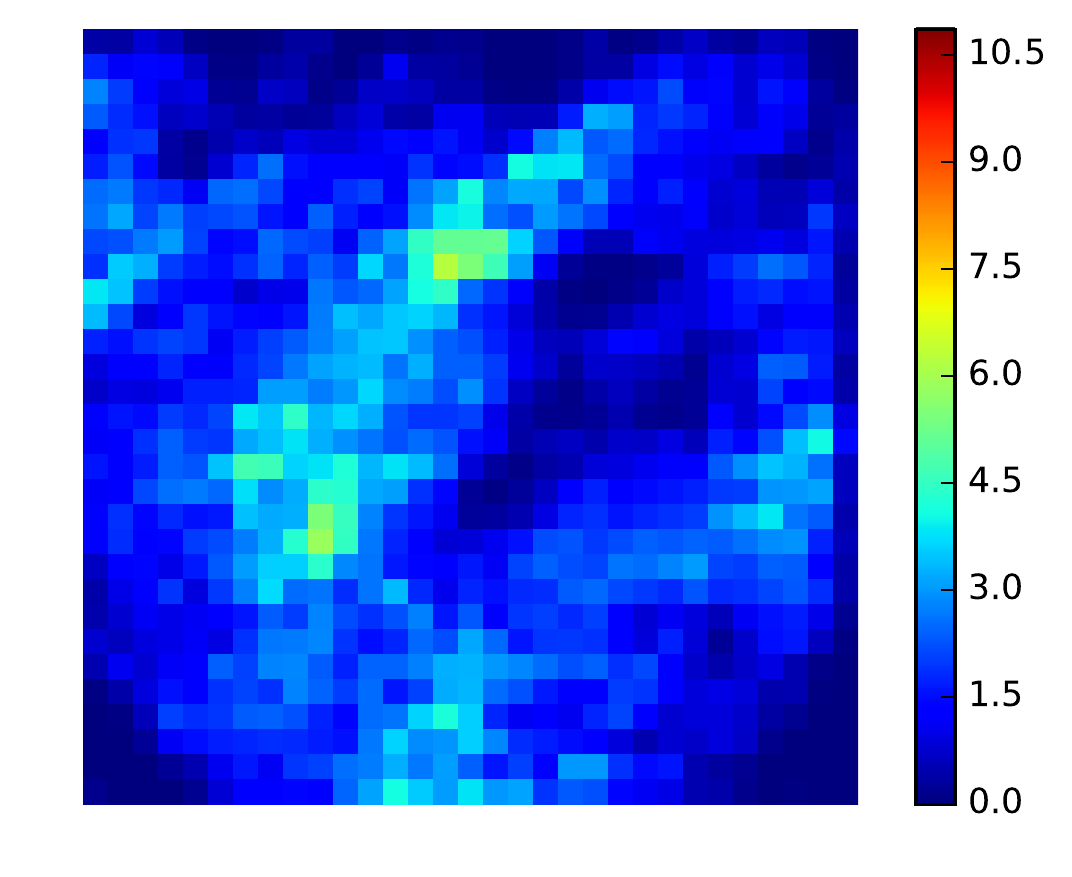}
		&
		\includegraphics[width=\linewidth]{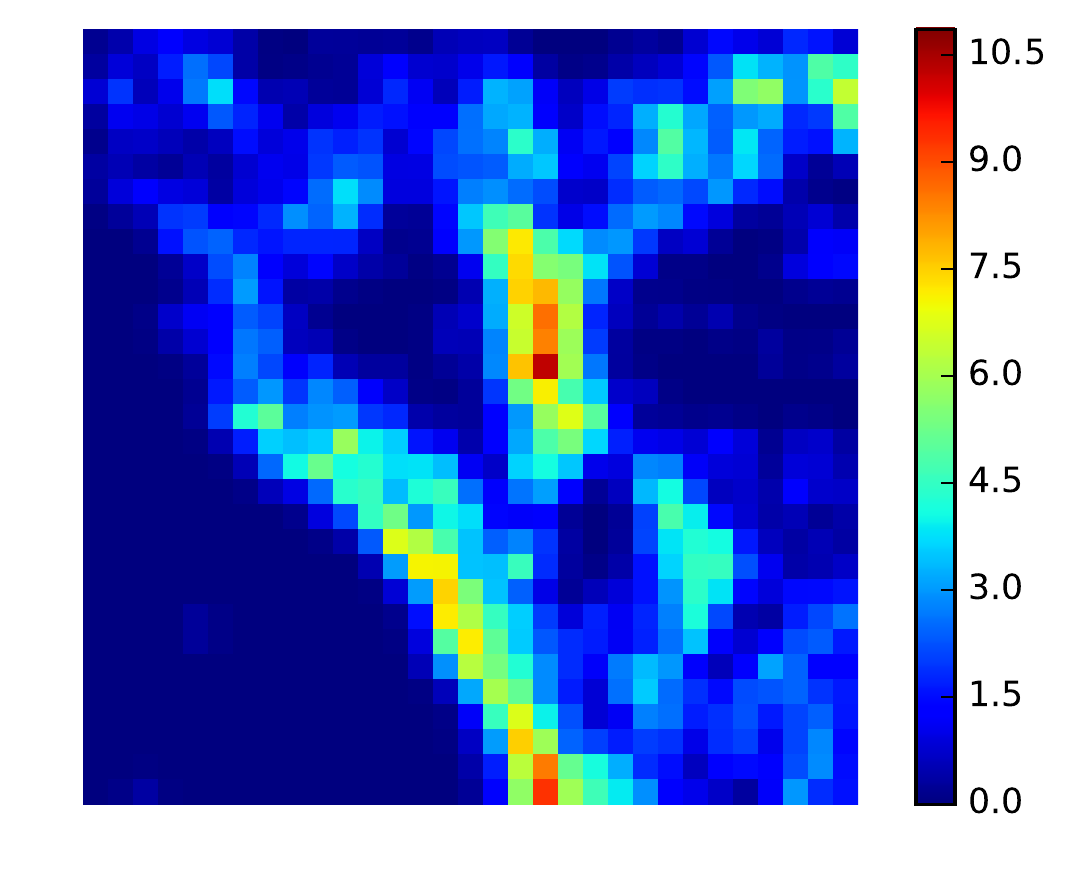}
		&
		\includegraphics[width=\linewidth]{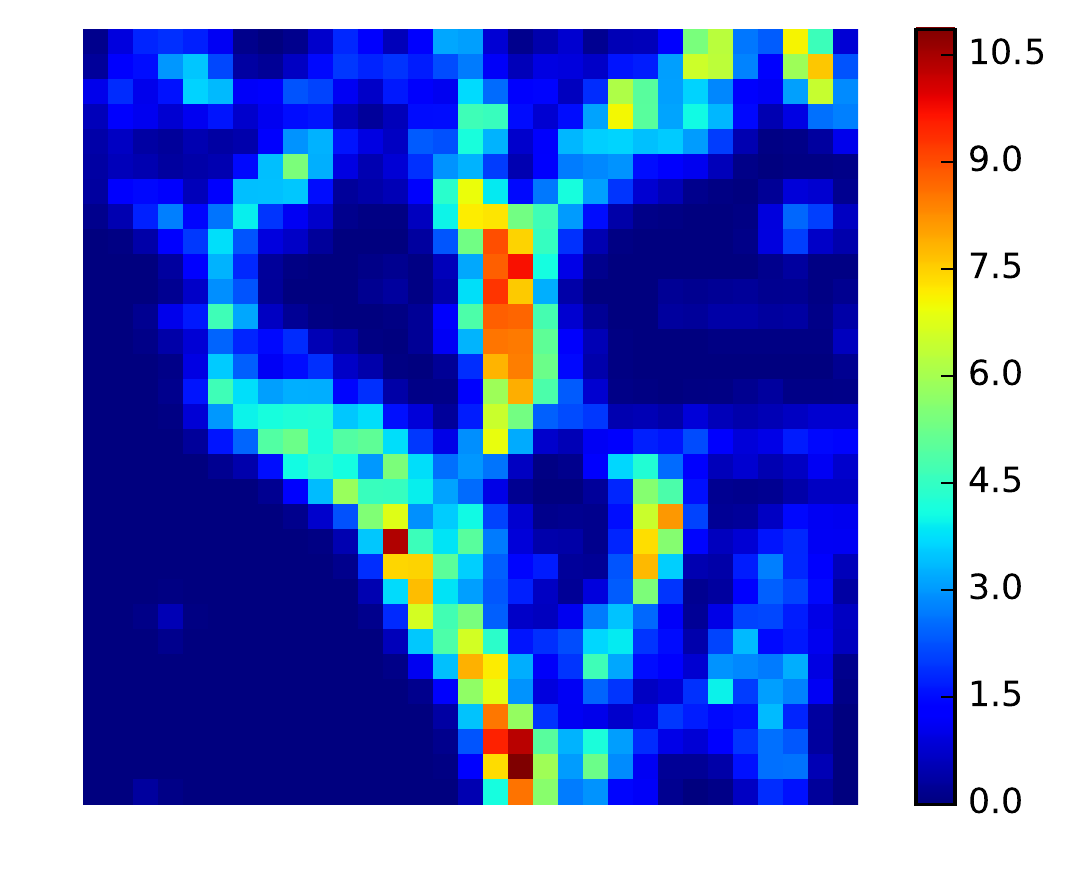}
		\\\addlinespace[-1ex]
		Scene and patch & IWE using flow $\bparams_0$ & IWE using flow $\bparams_1$ & IWE using flow $\bparams_2$
		\\
		\midrule
		\includegraphics[width=\linewidth]{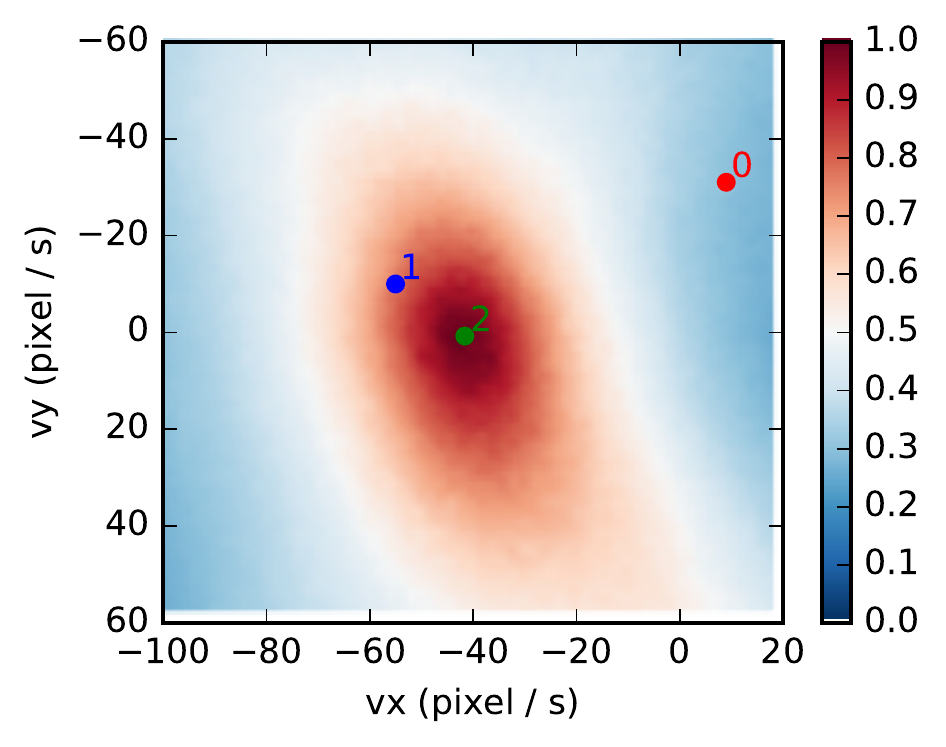}
		&
		\includegraphics[width=\linewidth]{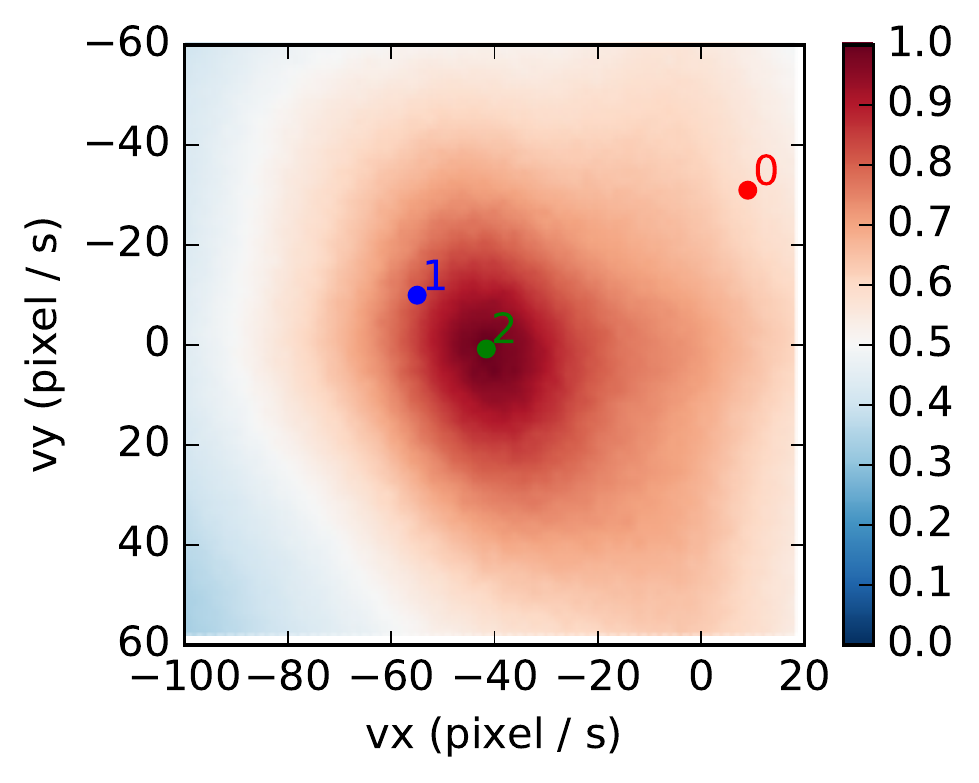}
		&
		\includegraphics[width=\linewidth]{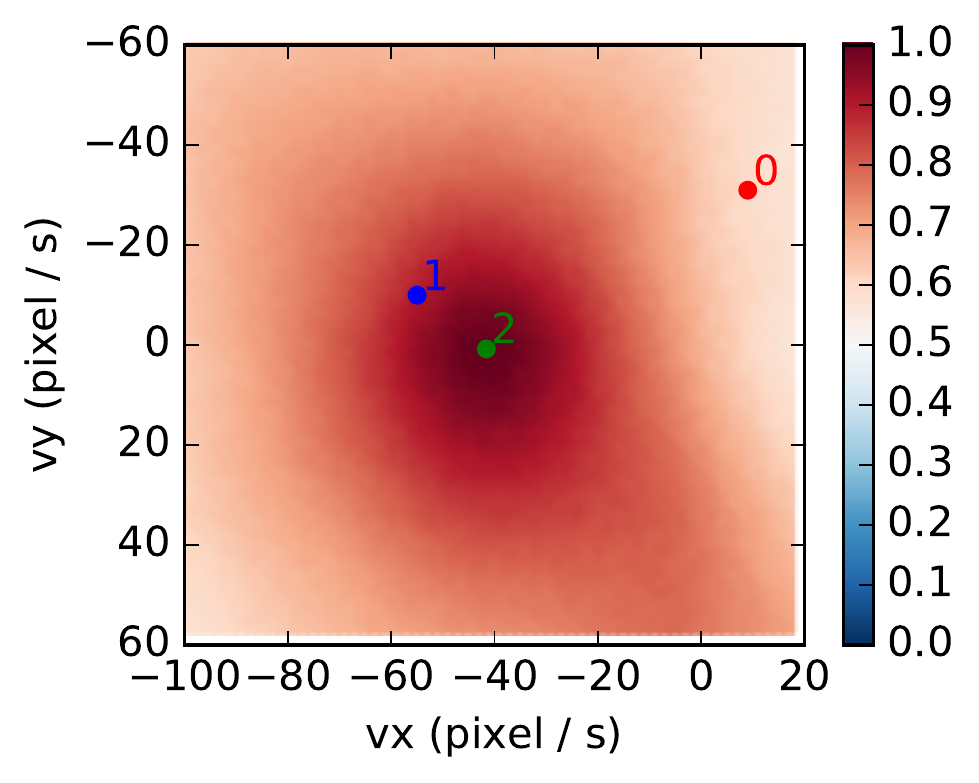}
		&
		\includegraphics[width=\linewidth]{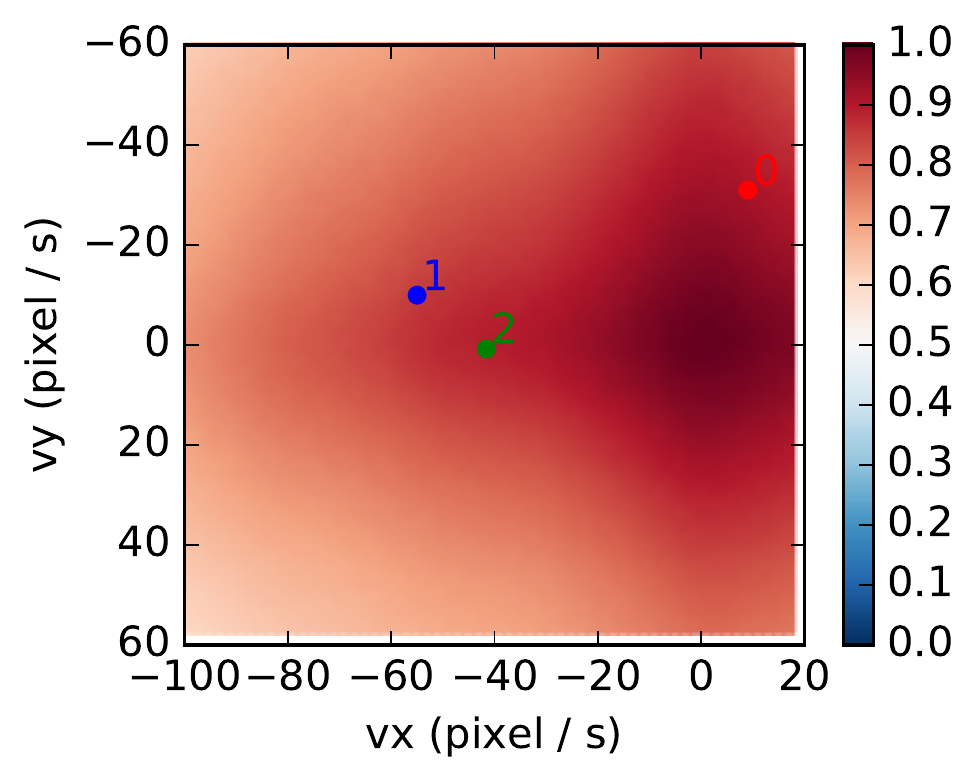}
		\\\addlinespace[-1ex]
		Variance & Mean Square & MAD & MAV
		\\\addlinespace[1ex]

		\includegraphics[width=\linewidth]{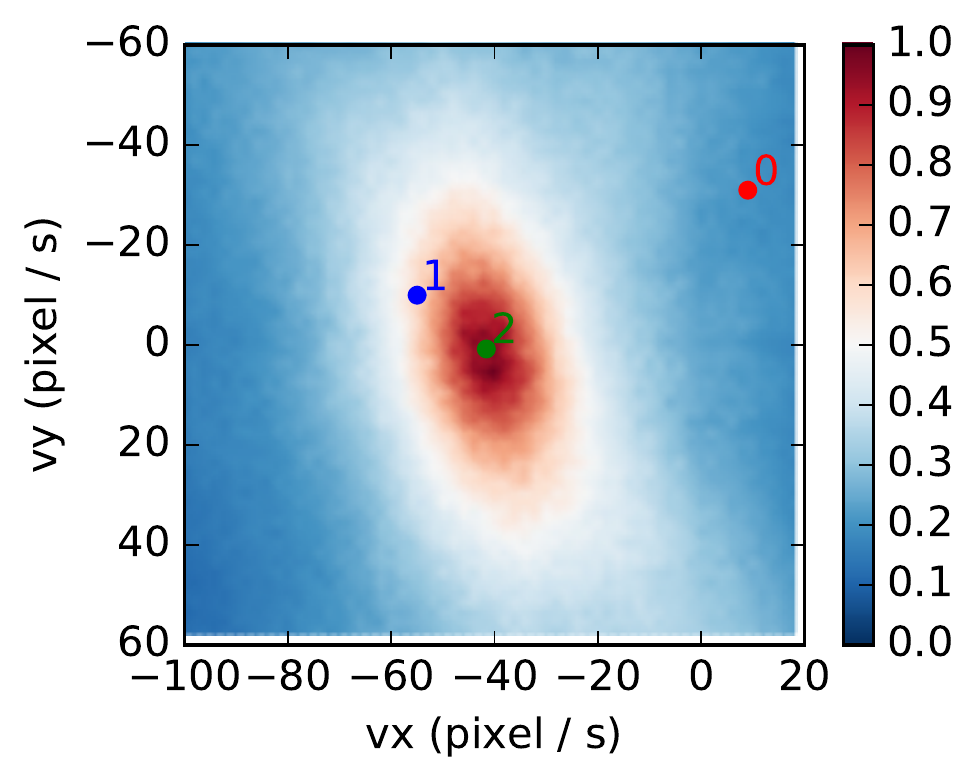}
		&
		\includegraphics[width=\linewidth]{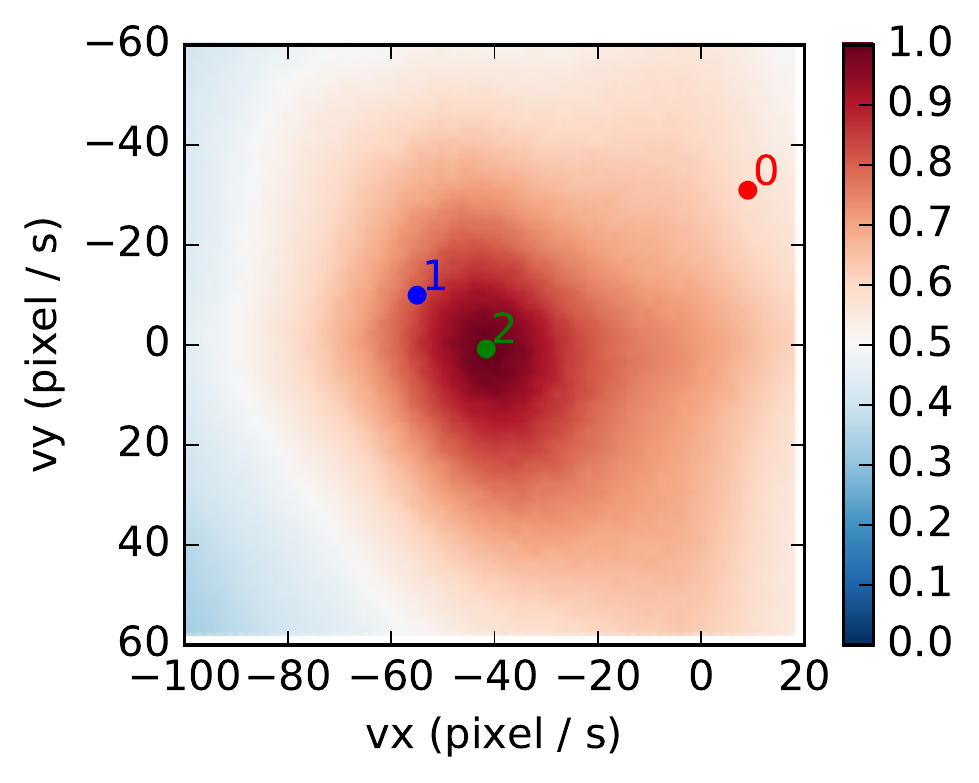}
		&
		\includegraphics[width=\linewidth]{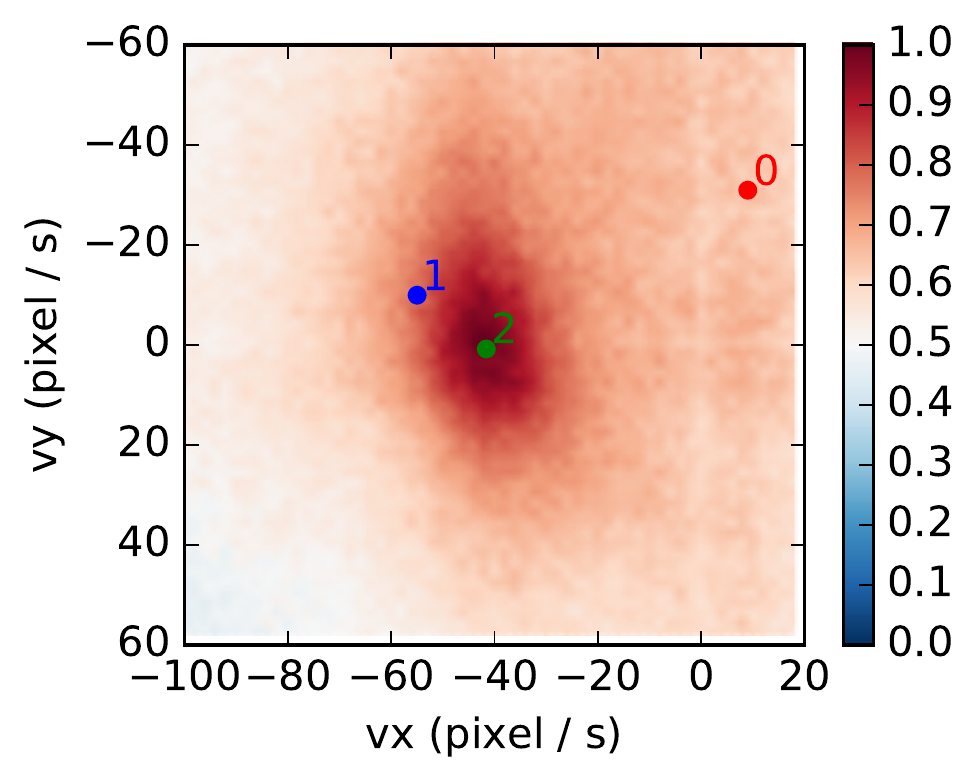}
		&
		\includegraphics[width=\linewidth]{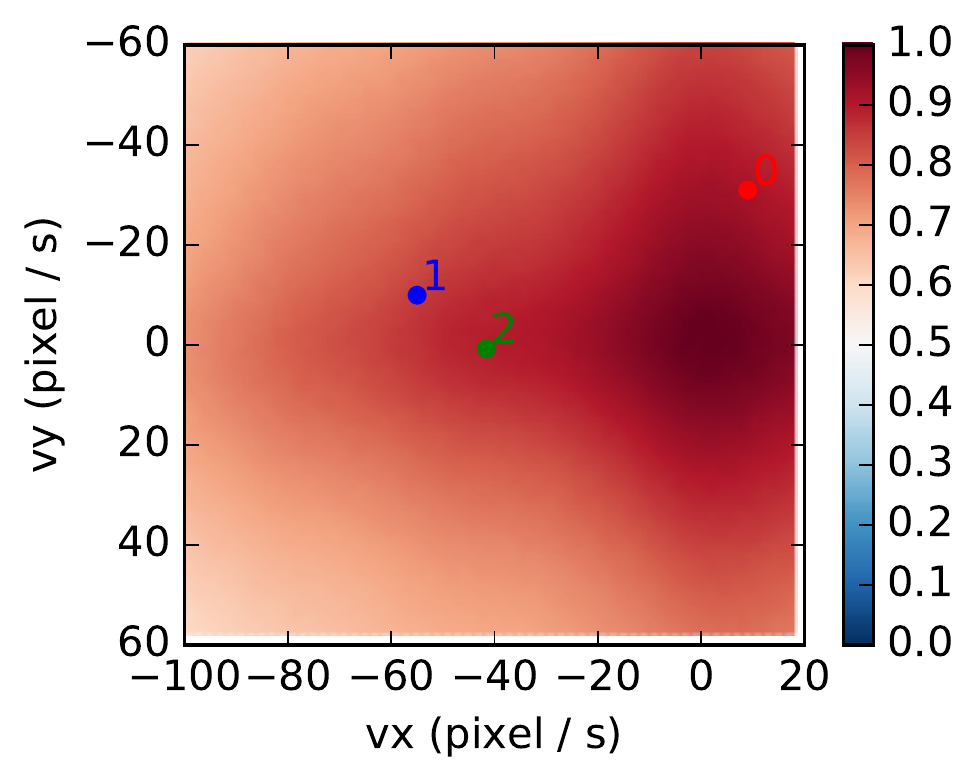}
		\\\addlinespace[-1ex]
		Local Variance & Local MS & Local MAD & Local MAV
		\\\addlinespace[1ex]
		
		\includegraphics[width=\linewidth]{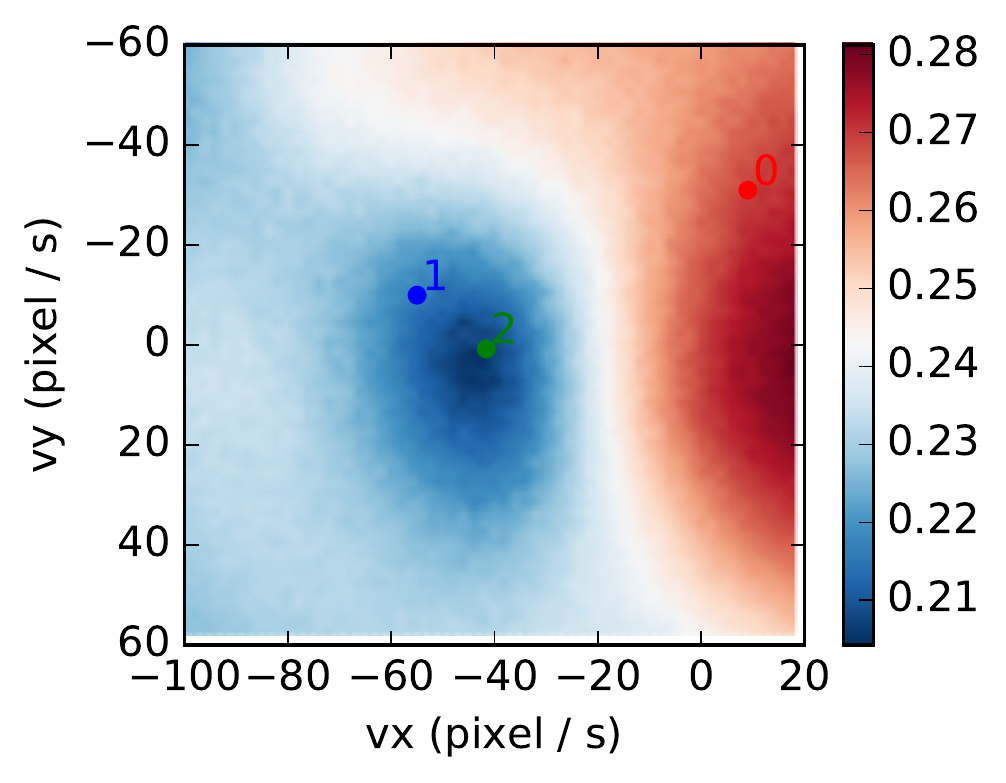}
		&
		\includegraphics[width=\linewidth]{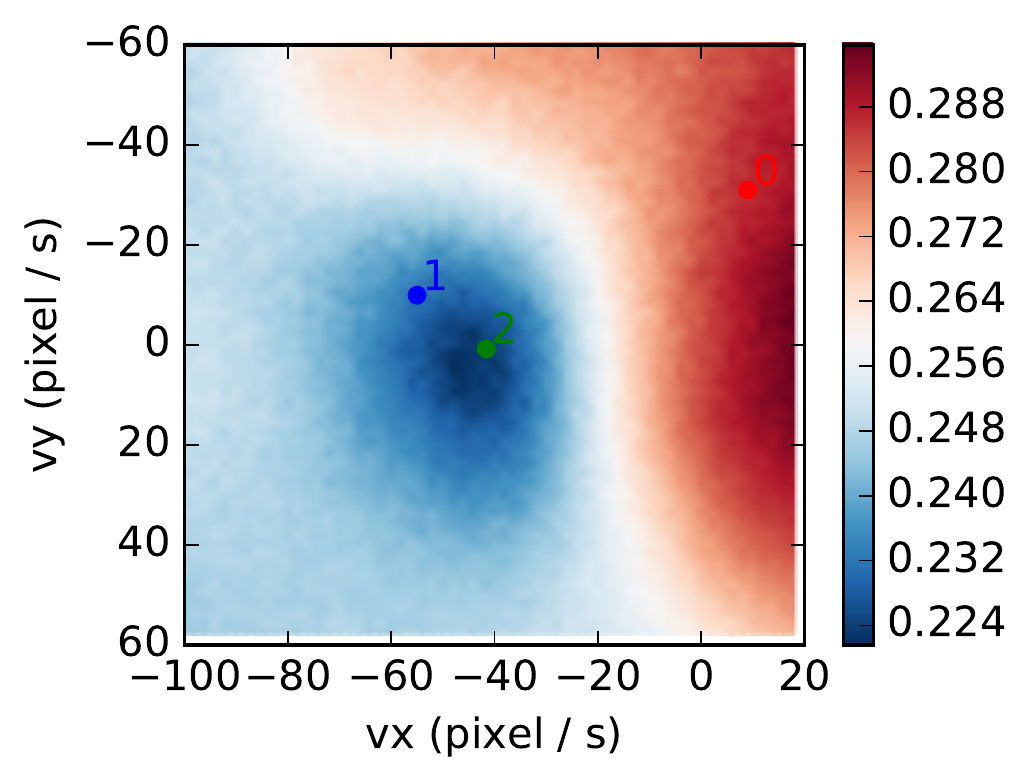}
		&
		\includegraphics[width=\linewidth]{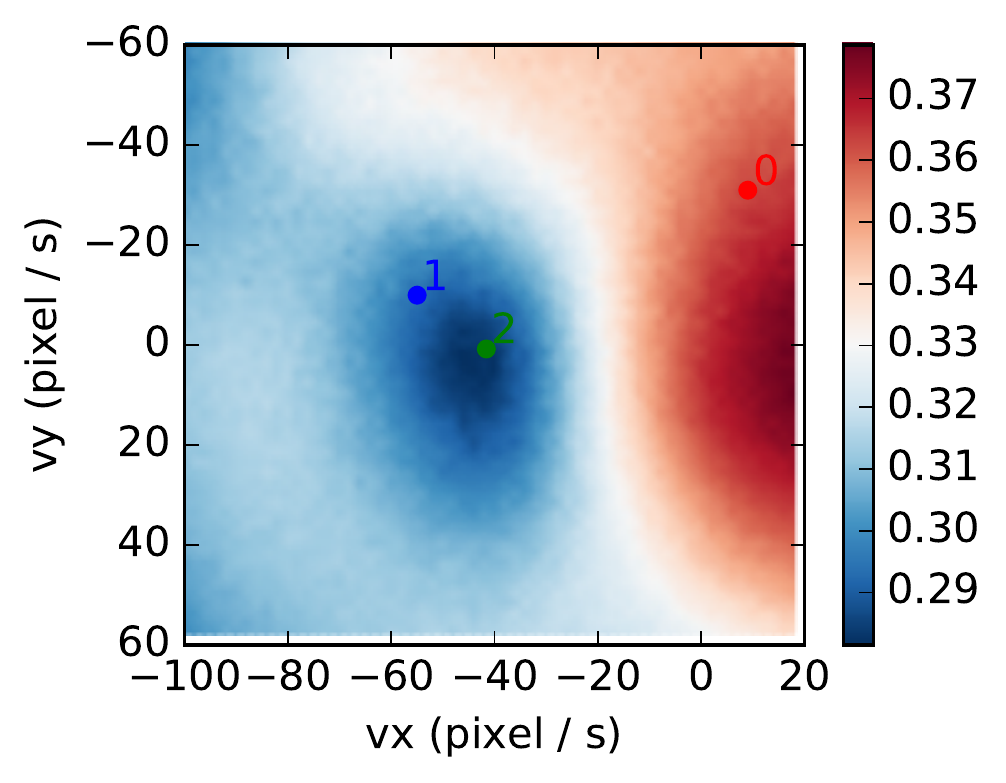}
		&
		\includegraphics[width=\linewidth]{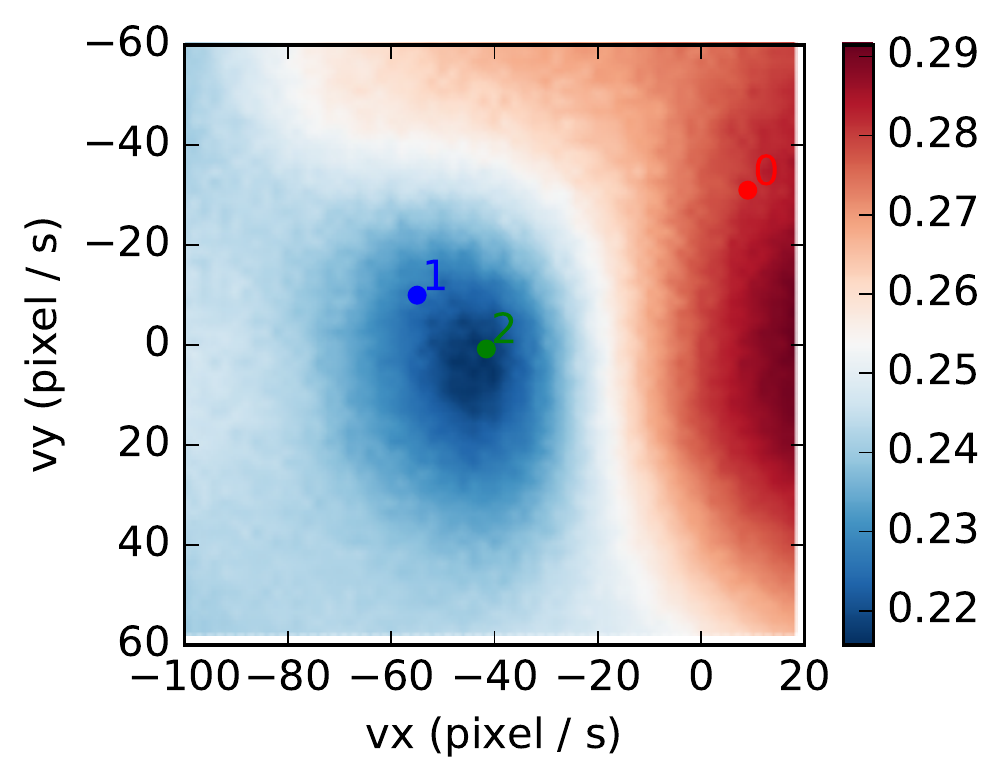}
		\\\addlinespace[-1ex]
		Area (Exp) & Area (Gaussian) & Area (Lorentz) & Area (Hyperbolic)
		\\\addlinespace[1ex]

		\includegraphics[width=\linewidth]{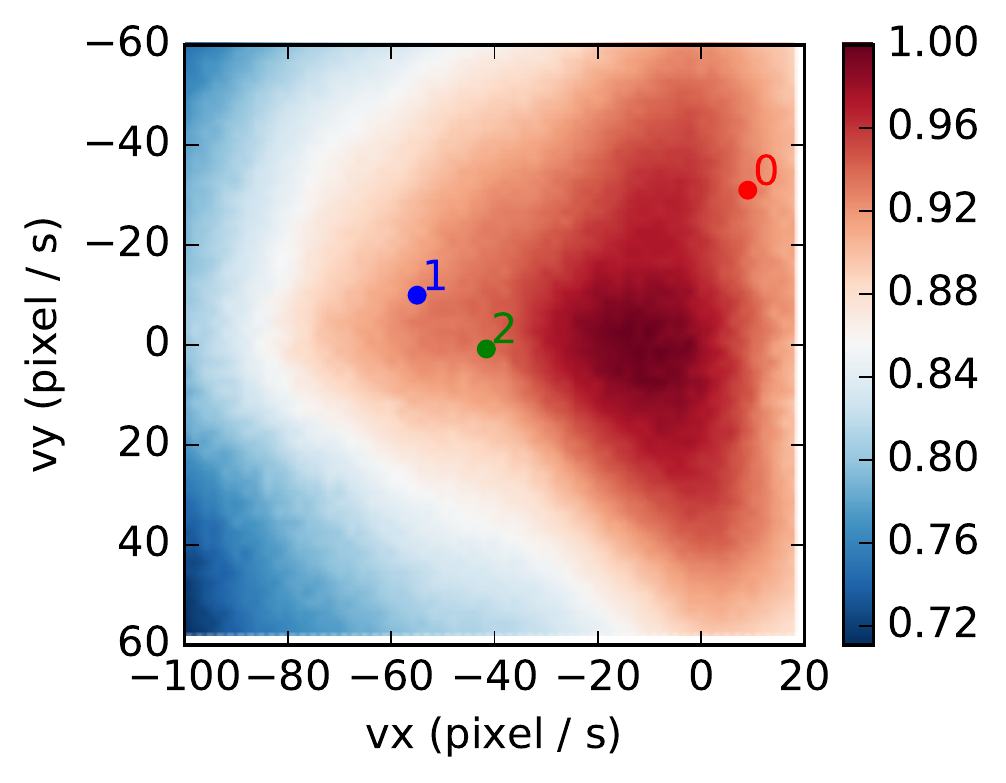}
		&
		\includegraphics[width=\linewidth]{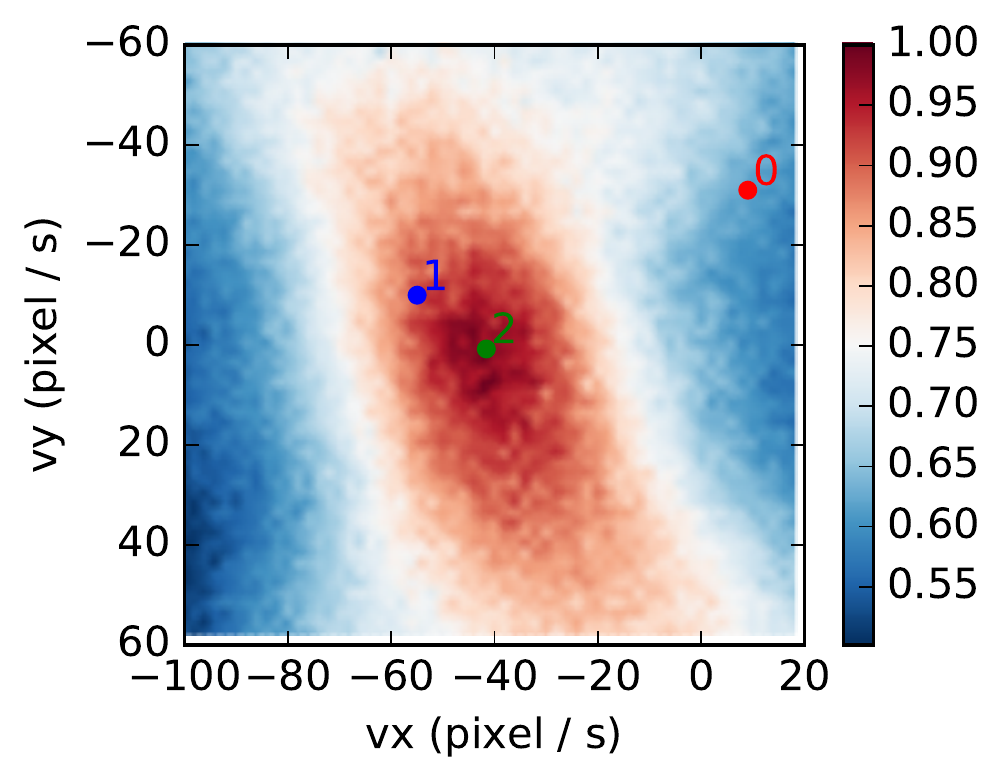}
		&
		\includegraphics[width=\linewidth]{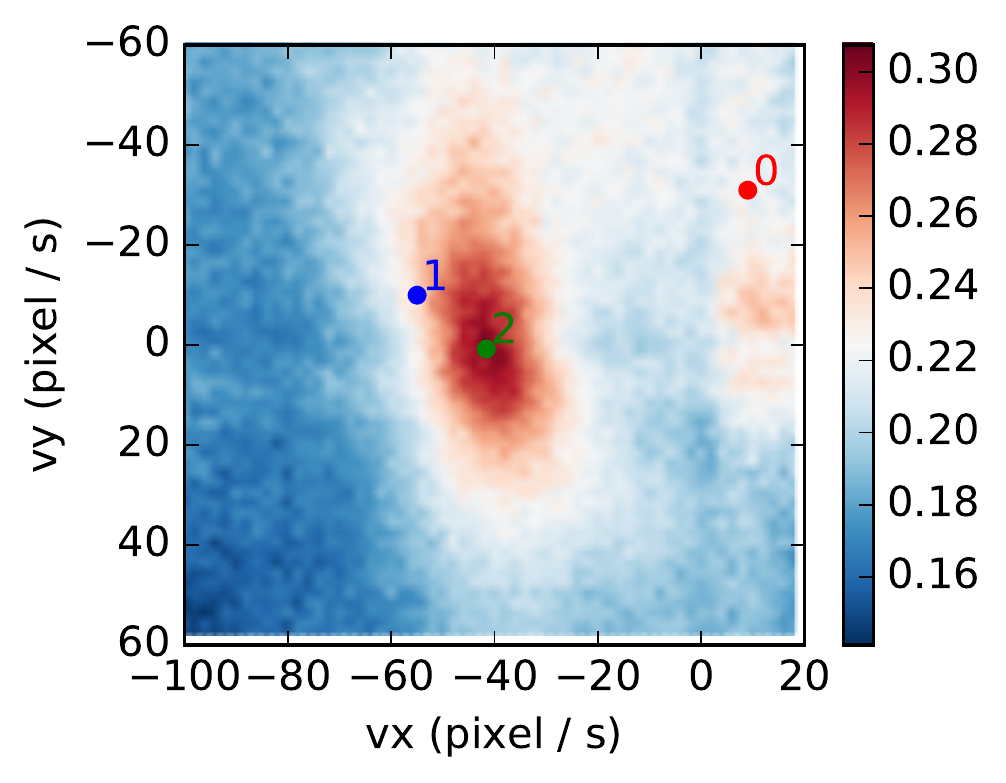}
		&
		\includegraphics[width=\linewidth]{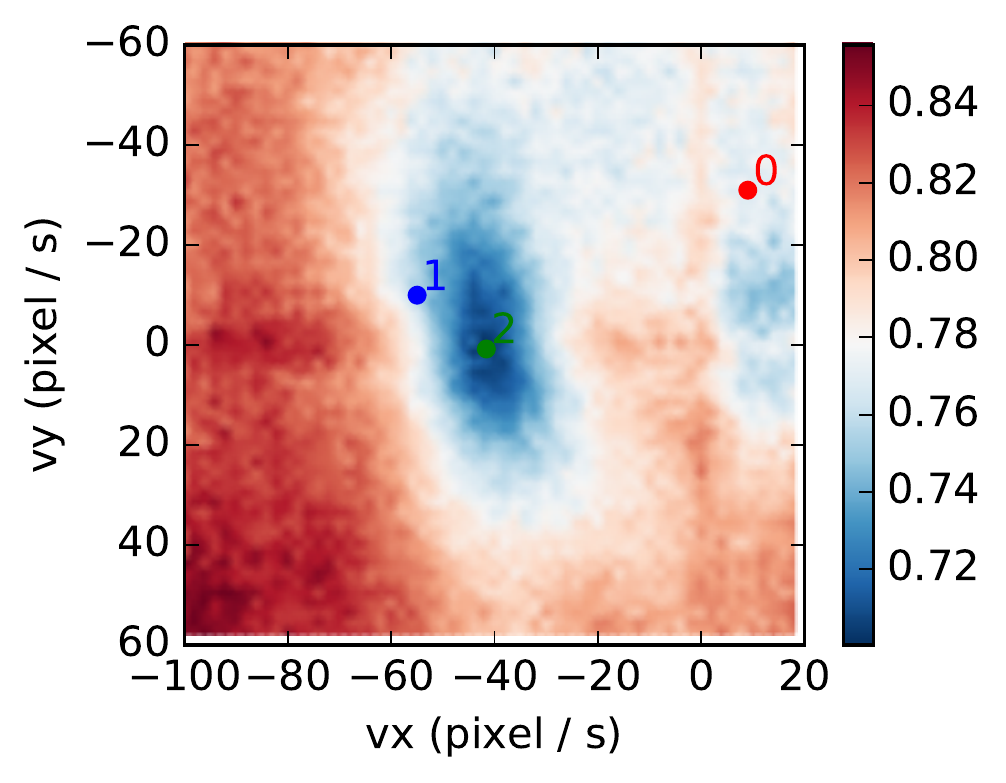}
		\\\addlinespace[-1ex]
		Entropy & Range (Exp) & Geary's $C$ & Moran's Index
		\\\addlinespace[1ex]

		\includegraphics[width=\linewidth]{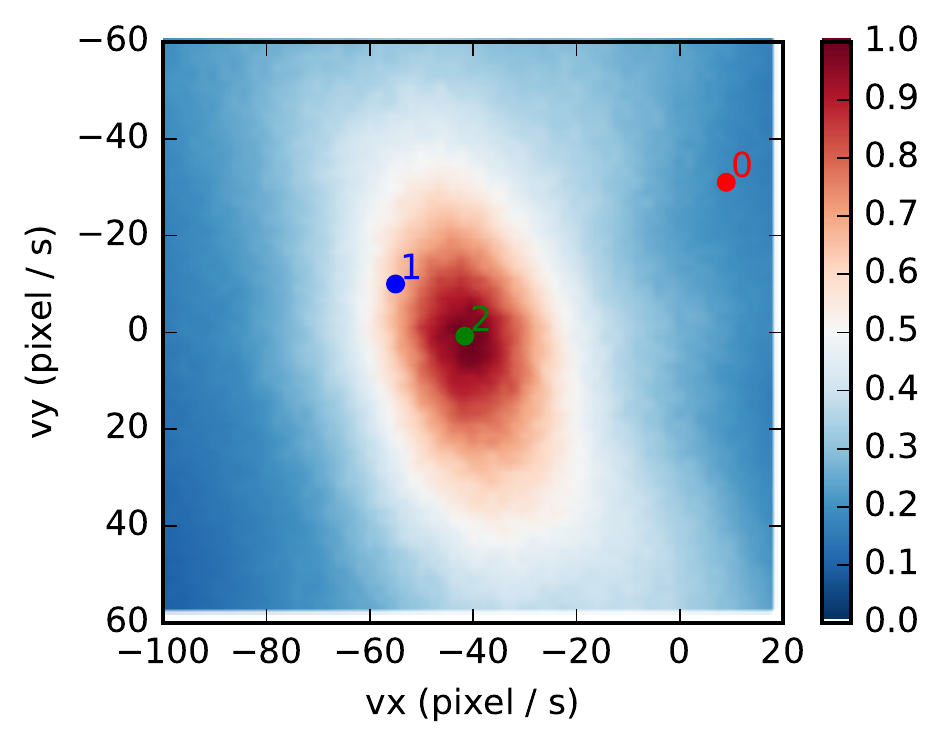}
		&
		\includegraphics[width=\linewidth]{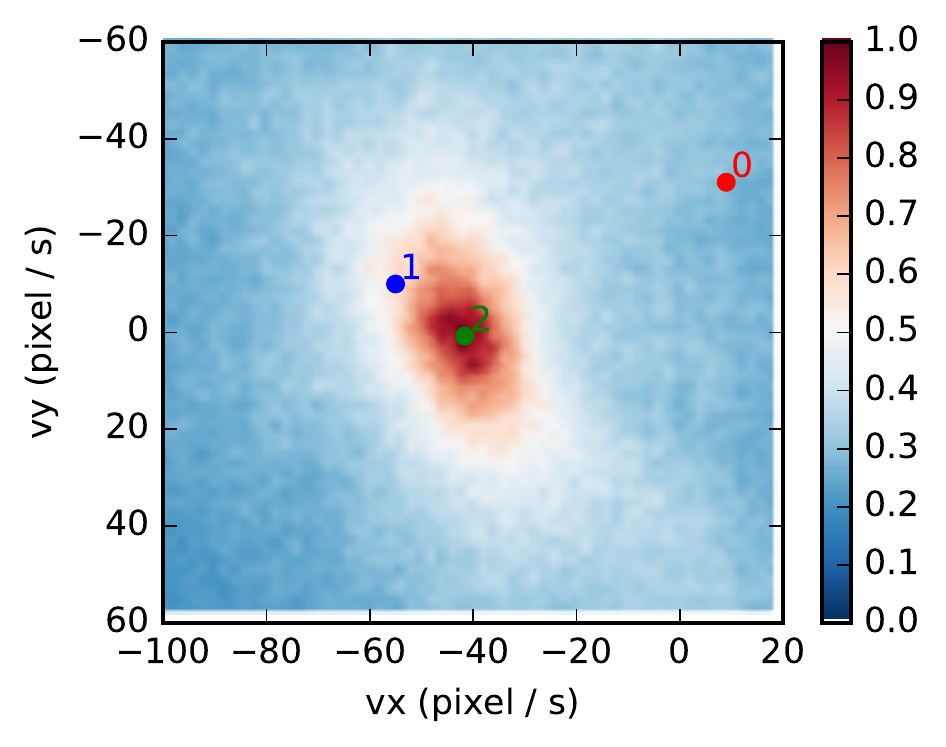}
		&
		\includegraphics[width=\linewidth]{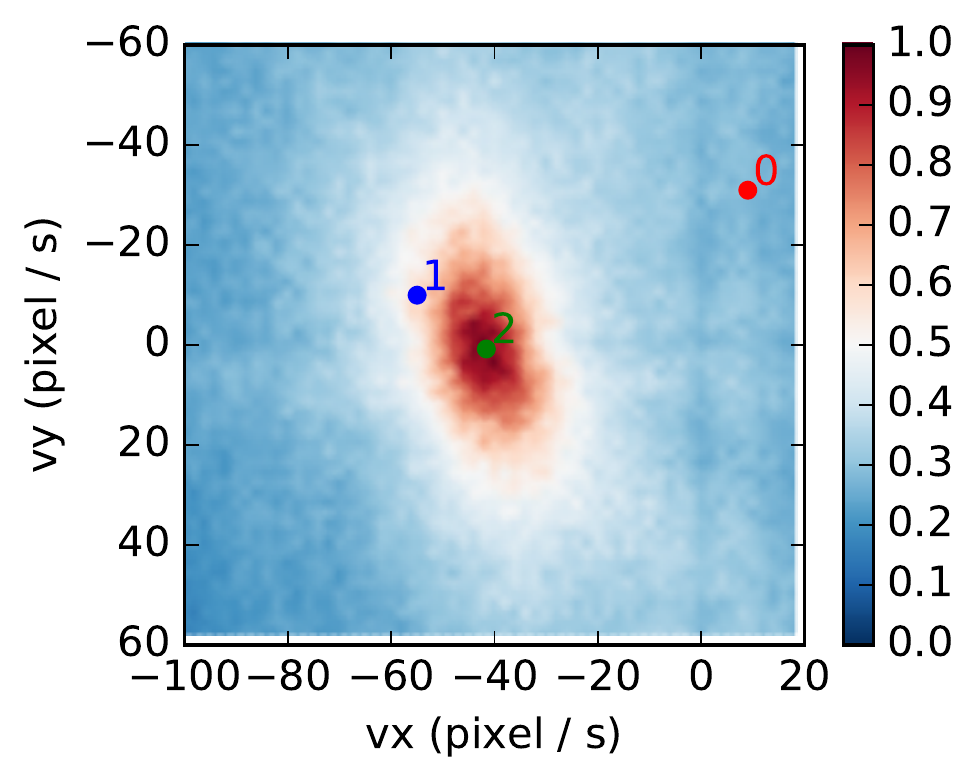}
		&
		\includegraphics[width=\linewidth]{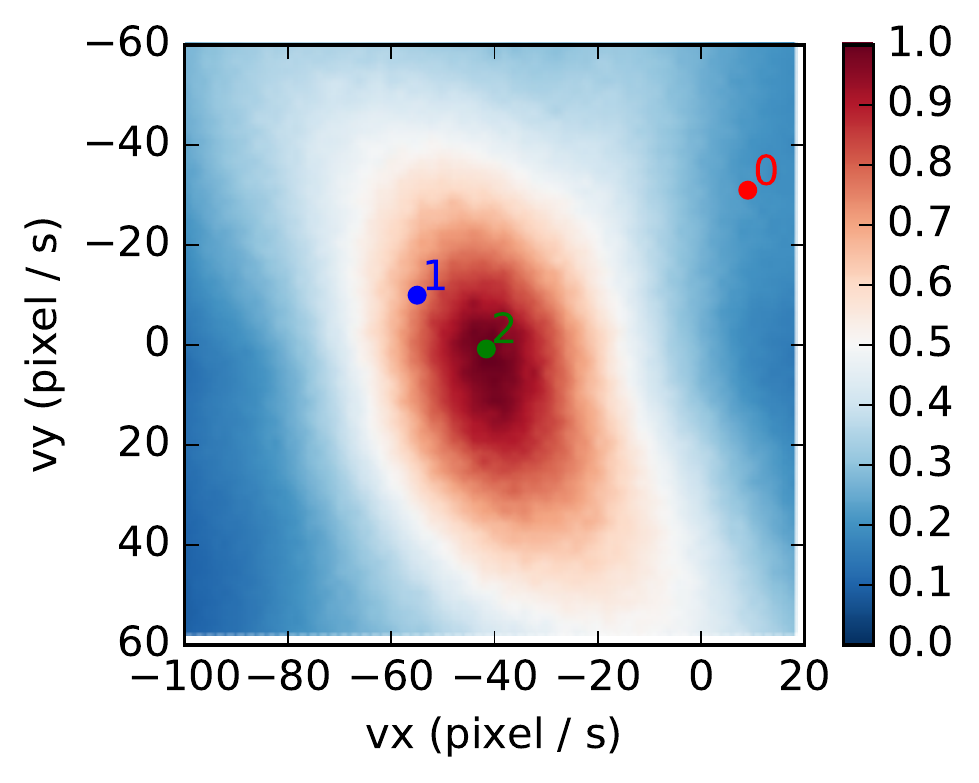}
		\\\addlinespace[-1ex]
		Gradient Magnitude & Laplacian Magnitude & Hessian Magnitude & DoG Magnitude
		\\\addlinespace[1ex]

		\includegraphics[width=\linewidth]{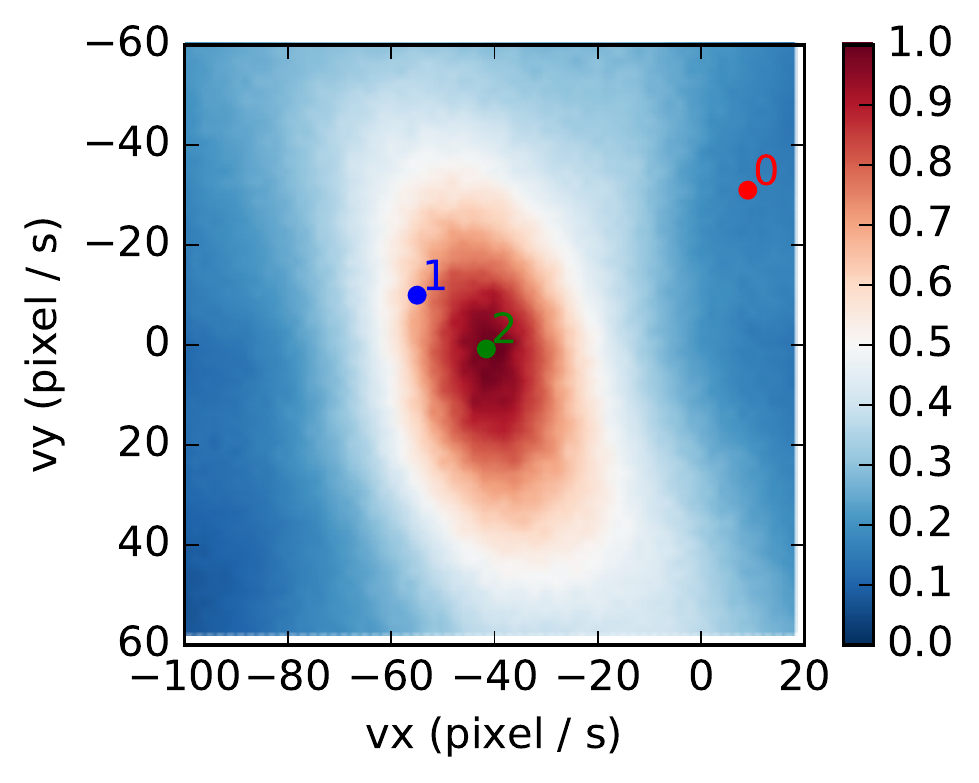}
		&
		\includegraphics[width=\linewidth]{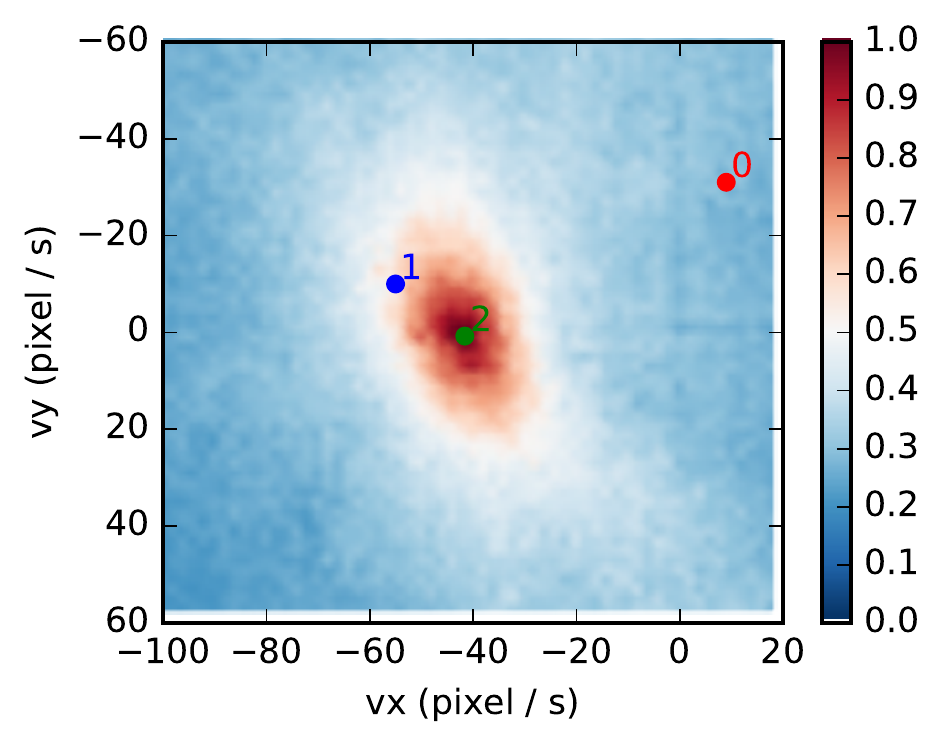}
		&
		\includegraphics[width=\linewidth]{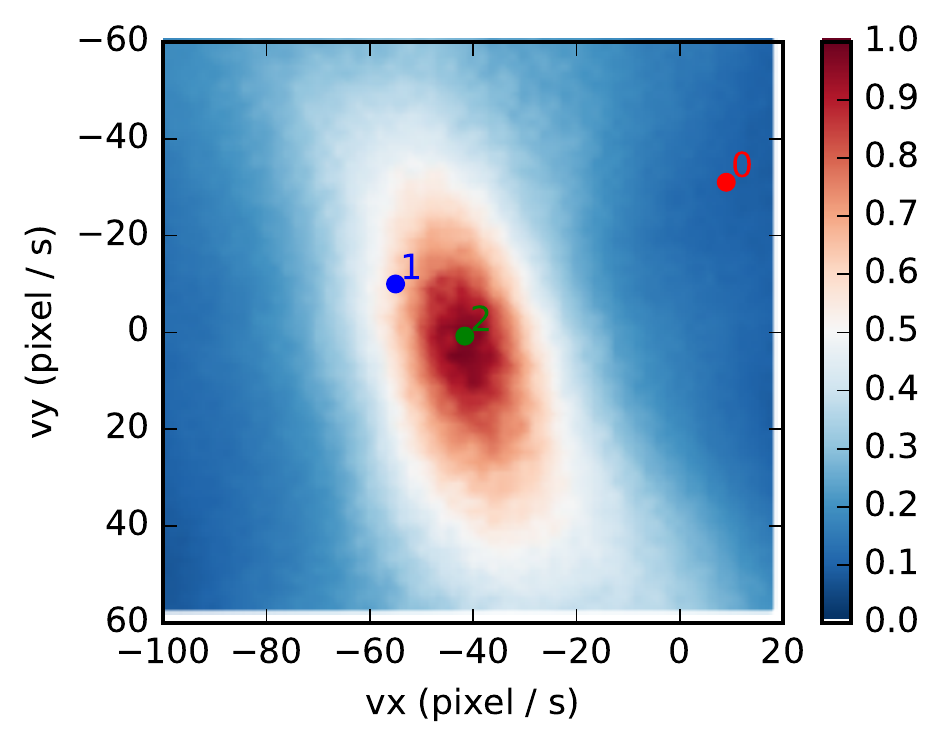}
		&
		\includegraphics[width=\linewidth]{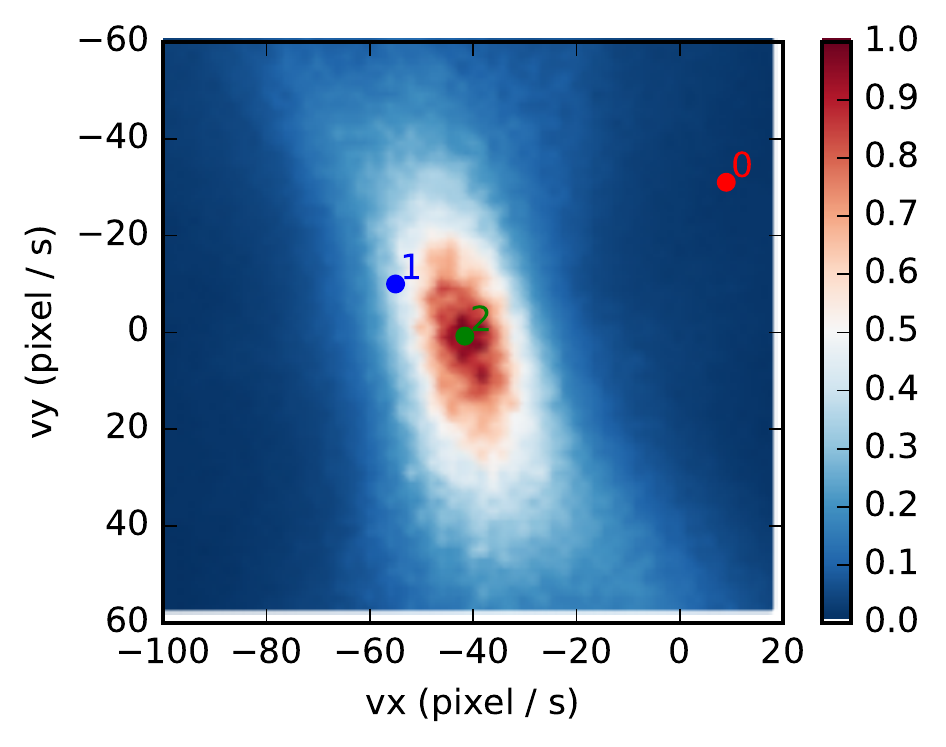}
		\\\addlinespace[-1ex]
		LoG Magnitude & Var. of Laplacian & Var. of Grad. & Var. of Squared Grad.
		\\

	\end{tabular}
	}
	\caption{Visualization of the Focus Loss Functions (as heat maps, pseudo-colored from blue to red).
	The top-left image shows a selected patch (highlighted in yellow) and three candidate flow vectors $\{\bparams_i\}_{i=0}^2$.
	The ground truth flow is close to $\bparams_2 = (-40,0)\si{pixel/\second}$.
	The top row also shows the warped events (IWE patch) using the three flow vectors, without using polarity in the IWE.
	The remaining rows show the focus loss functions in optical flow (i.e., image velocity) space.}
	\label{fig:boxes:oflowpatch:surfaces}
\end{figure*}

\begin{figure*}[t!]
	\centering
    {\small
    \setlength{\tabcolsep}{2pt}
	\begin{tabular}{
	>{\centering\arraybackslash}m{\oflowsurfwidth} 
	>{\centering\arraybackslash}m{\oflowsurfwidth}
	>{\centering\arraybackslash}m{\oflowsurfwidth} 
	>{\centering\arraybackslash}m{\oflowsurfwidth}}
		\includegraphics[width=\linewidth]{images/optical_flow/cvpr18/preview.pdf}
		&
		\includegraphics[width=\linewidth]{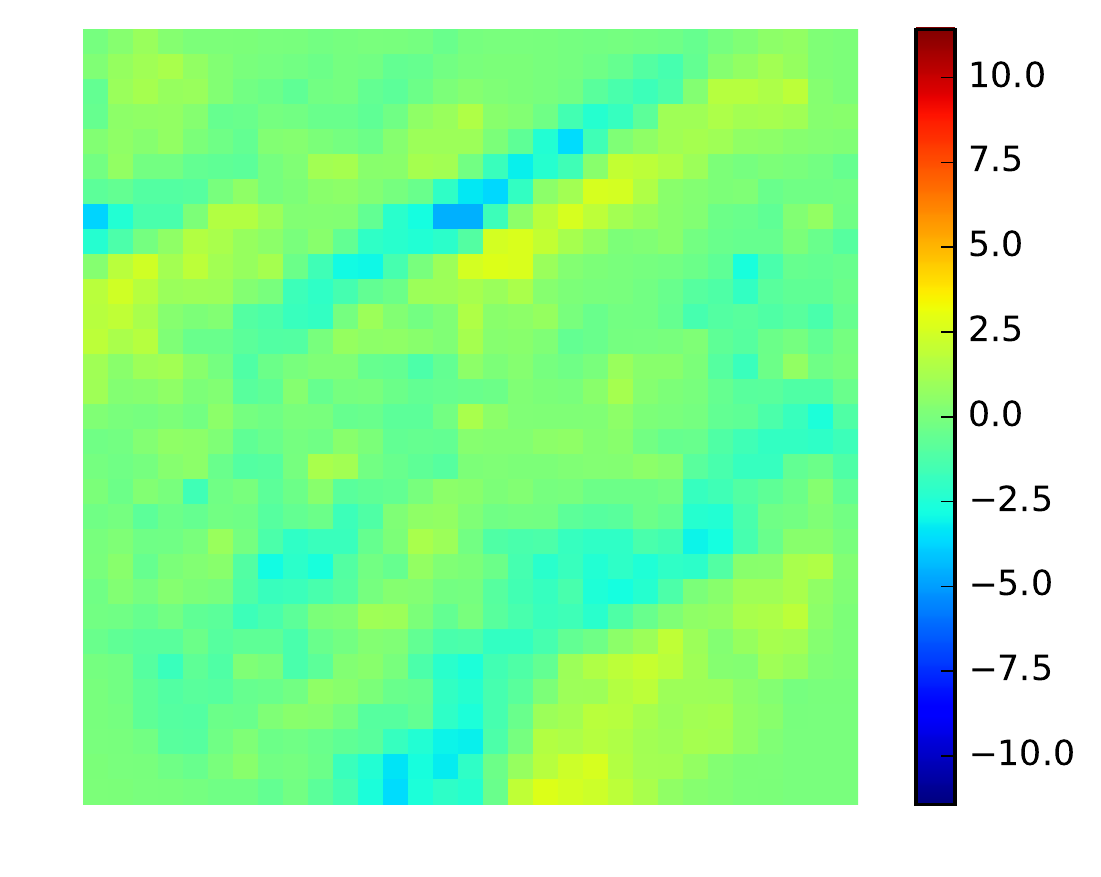}
		&
		\includegraphics[width=\linewidth]{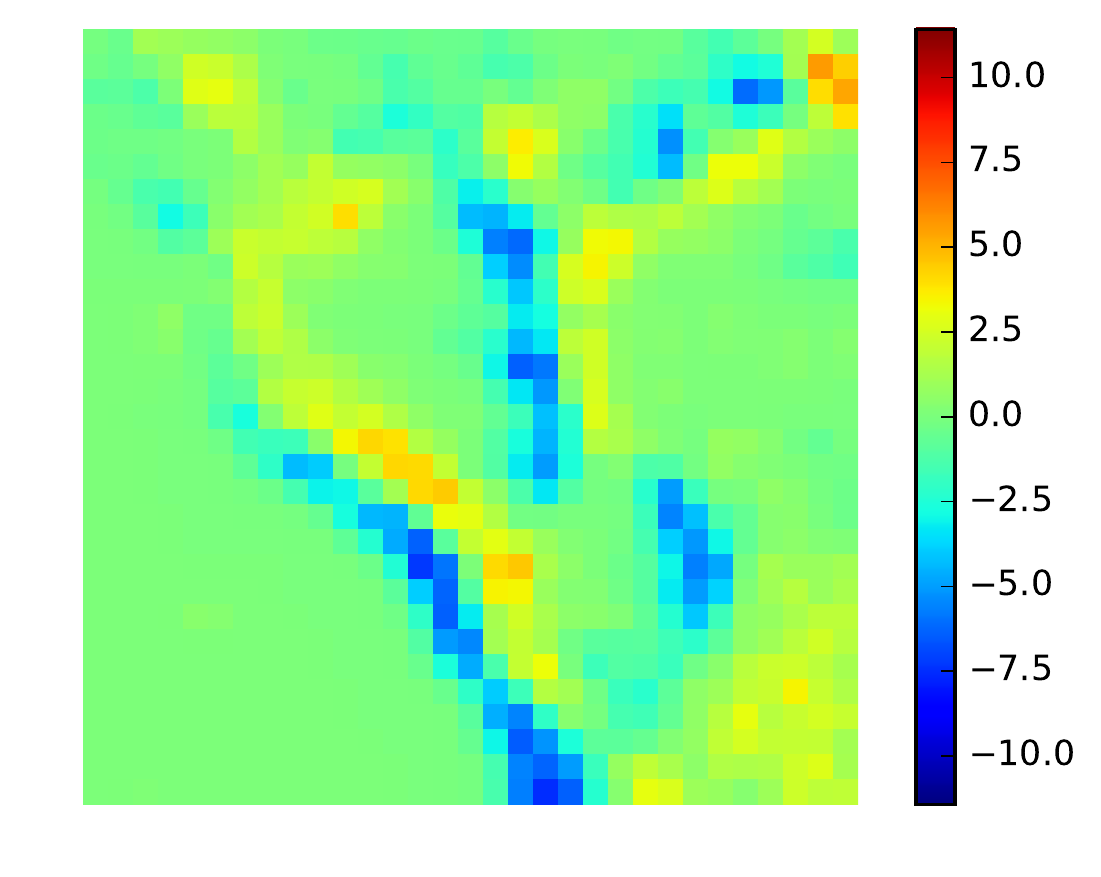}
		&
		\includegraphics[width=\linewidth]{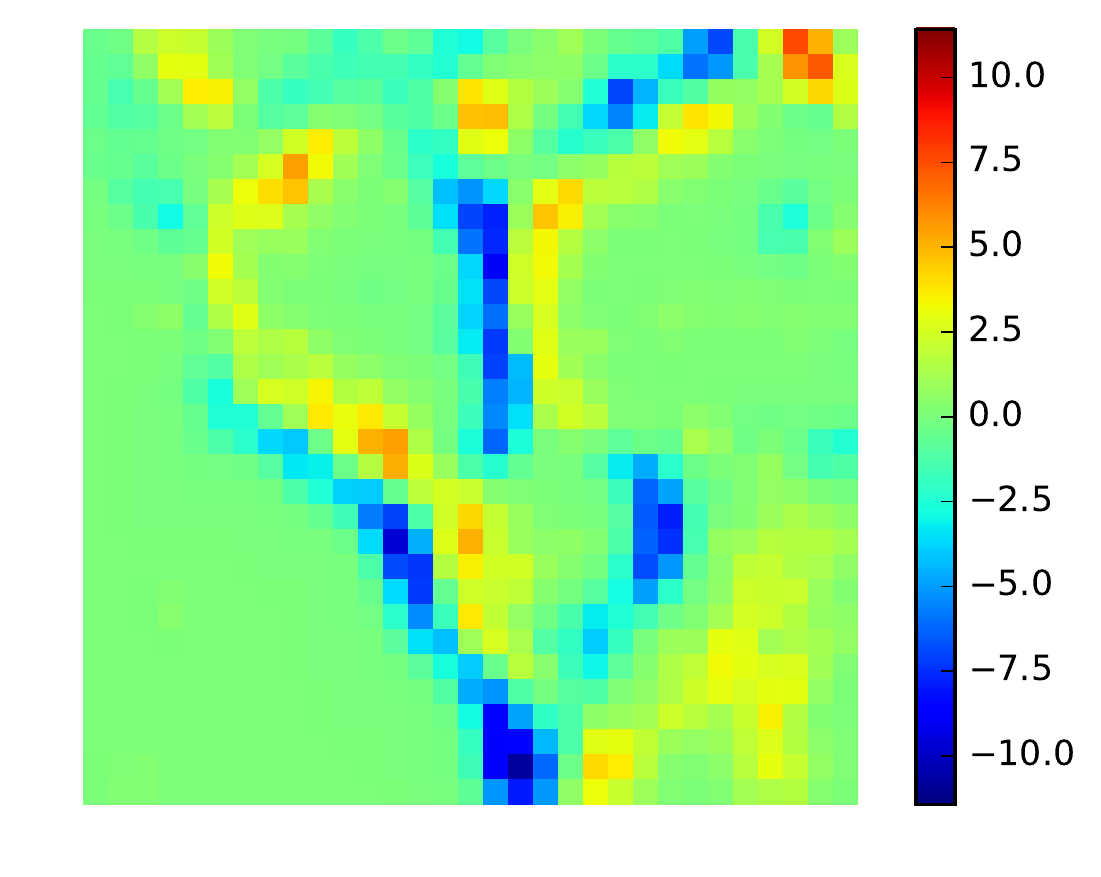}
		\\\addlinespace[-1ex]
		Scene and patch & IWE using flow $\bparams_0$ & IWE using flow $\bparams_1$ & IWE using flow $\bparams_2$
		\\
		\midrule
		\includegraphics[width=\linewidth]{images/optical_flow/cvpr18/with_polarity/cost_function_0_variance.pdf}
		&
		\includegraphics[width=\linewidth]{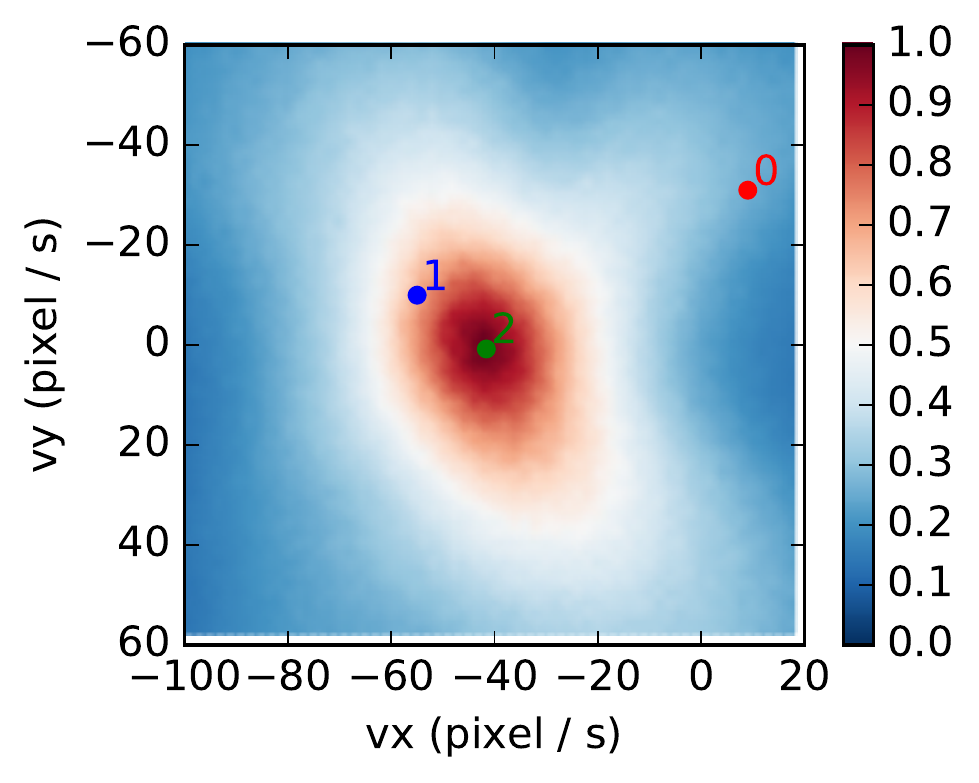}
		&
		\includegraphics[width=\linewidth]{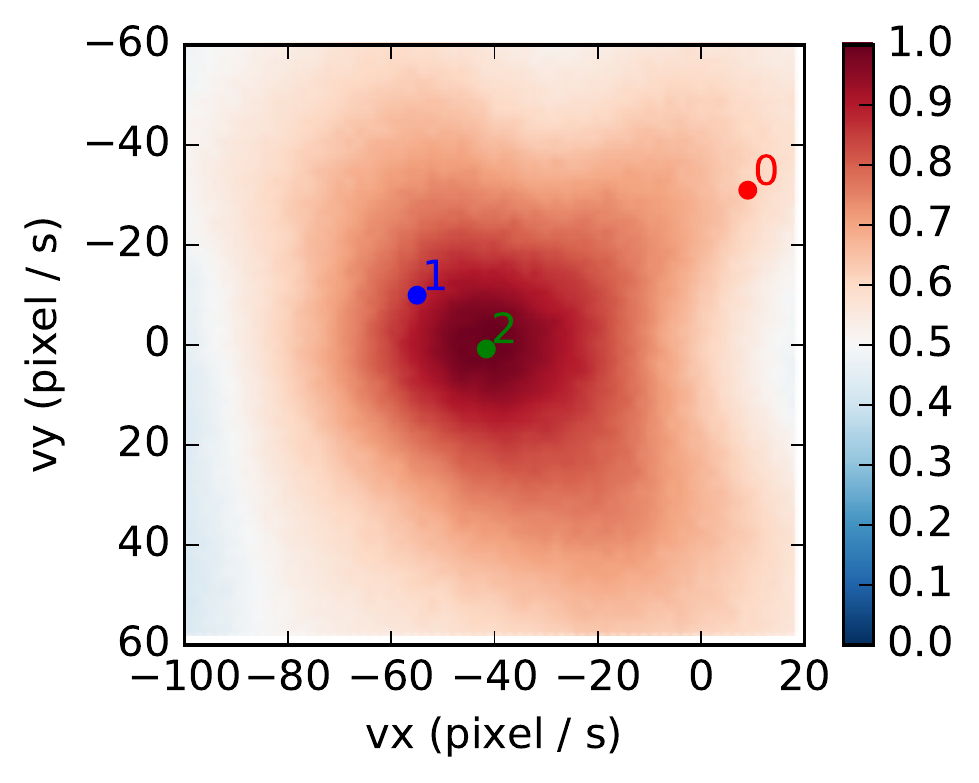}
		&
		\includegraphics[width=\linewidth]{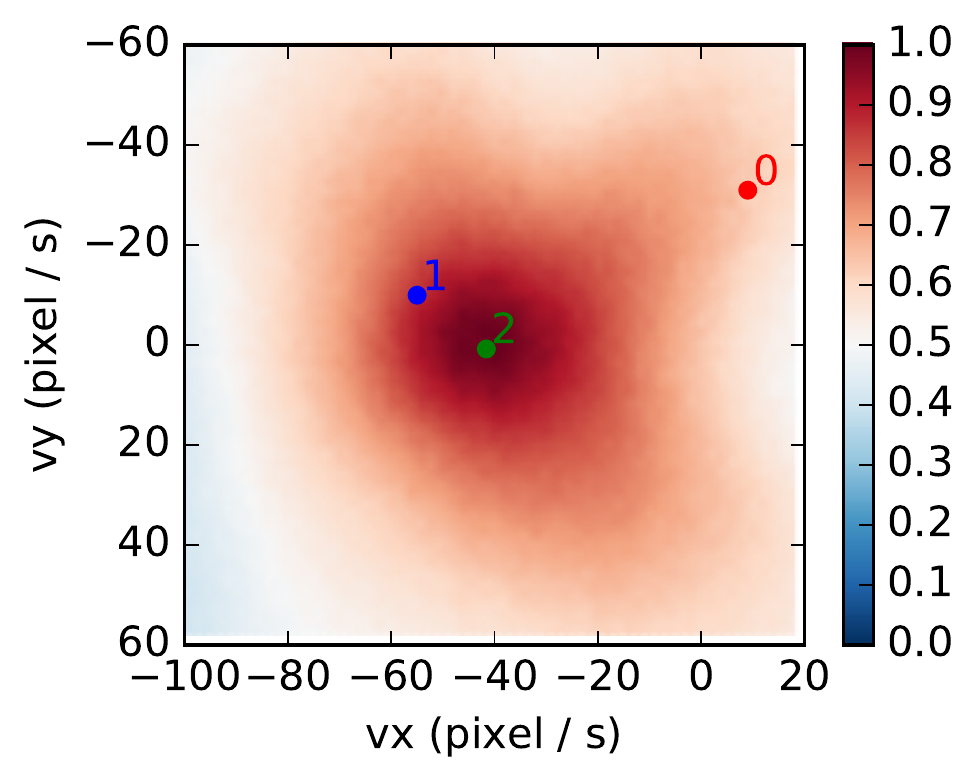}
		\\\addlinespace[-1ex]
		Variance & Mean Square & MAD & MAV
		\\\addlinespace[1ex]

		\includegraphics[width=\linewidth]{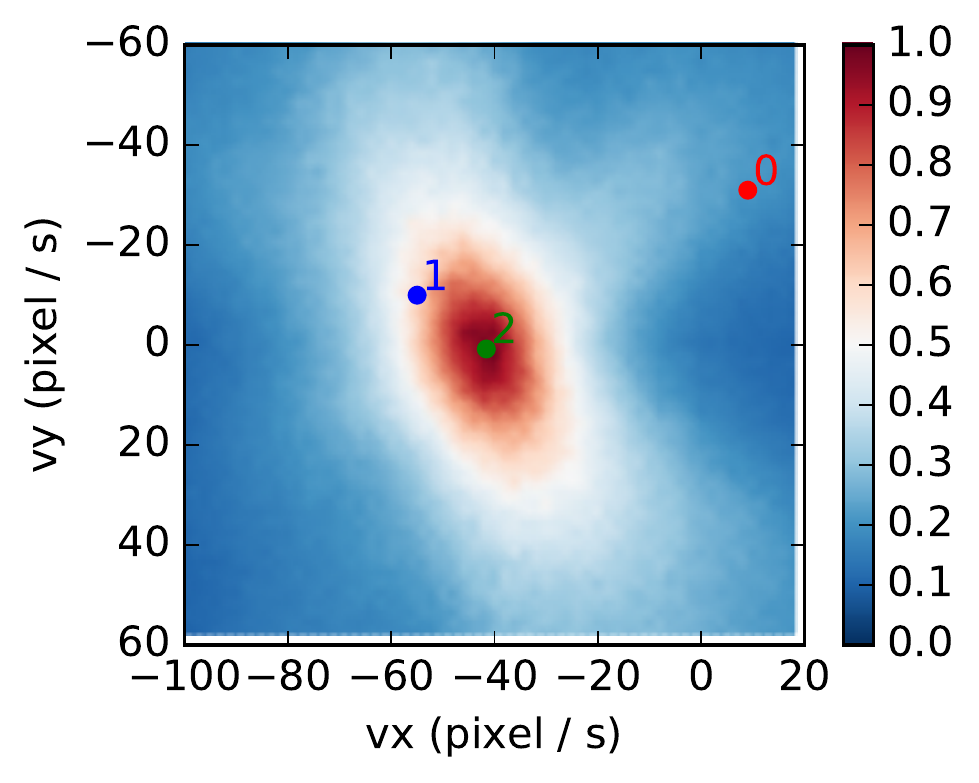}
		&
		\includegraphics[width=\linewidth]{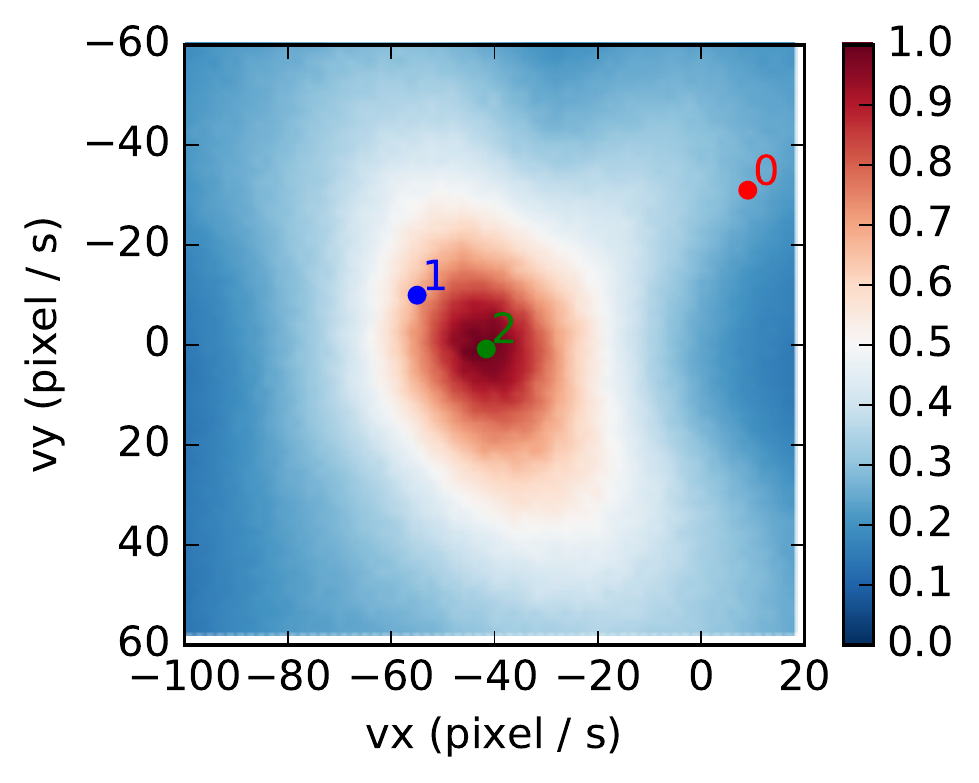}
		&
		\includegraphics[width=\linewidth]{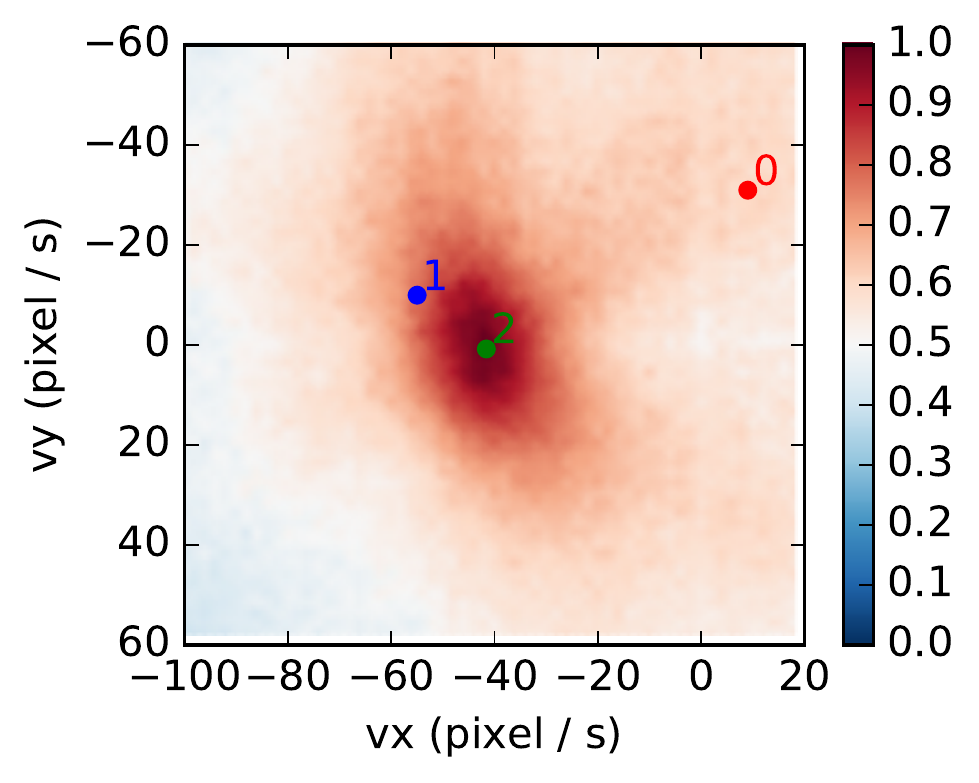}
		&
		\includegraphics[width=\linewidth]{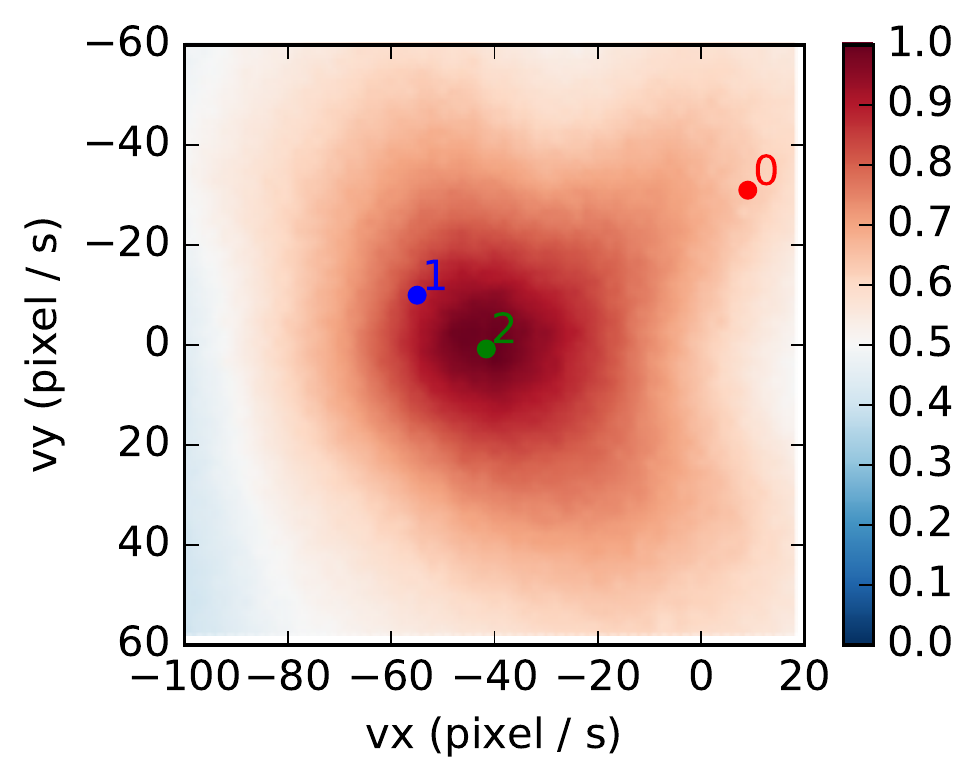}
		\\\addlinespace[-1ex]
		Local Variance & Local MS & Local MAD & Local MAV
		\\\addlinespace[1ex]
		
		\includegraphics[width=\linewidth]{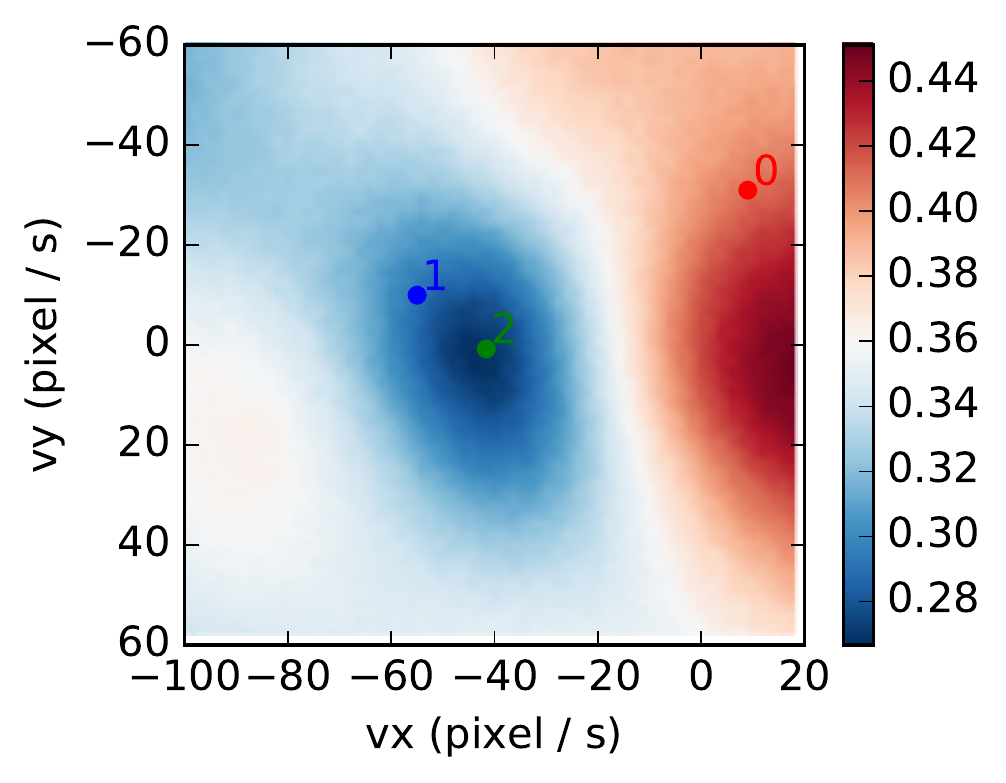}
		&
		\includegraphics[width=\linewidth]{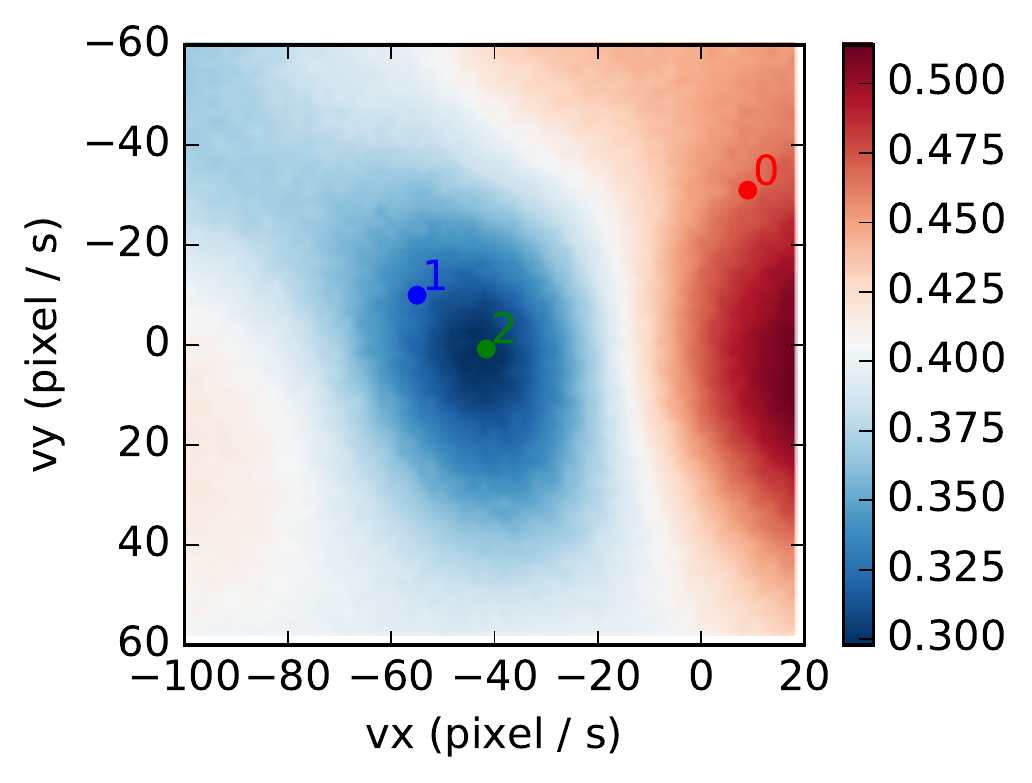}
		&
		\includegraphics[width=\linewidth]{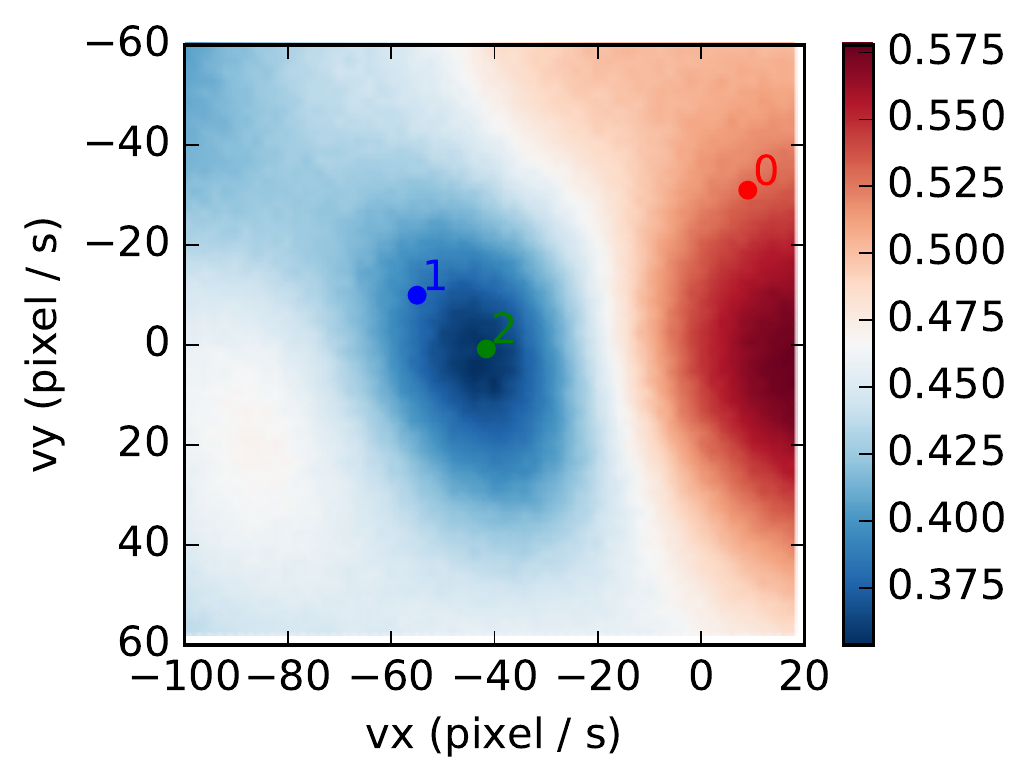}
		&
		\includegraphics[width=\linewidth]{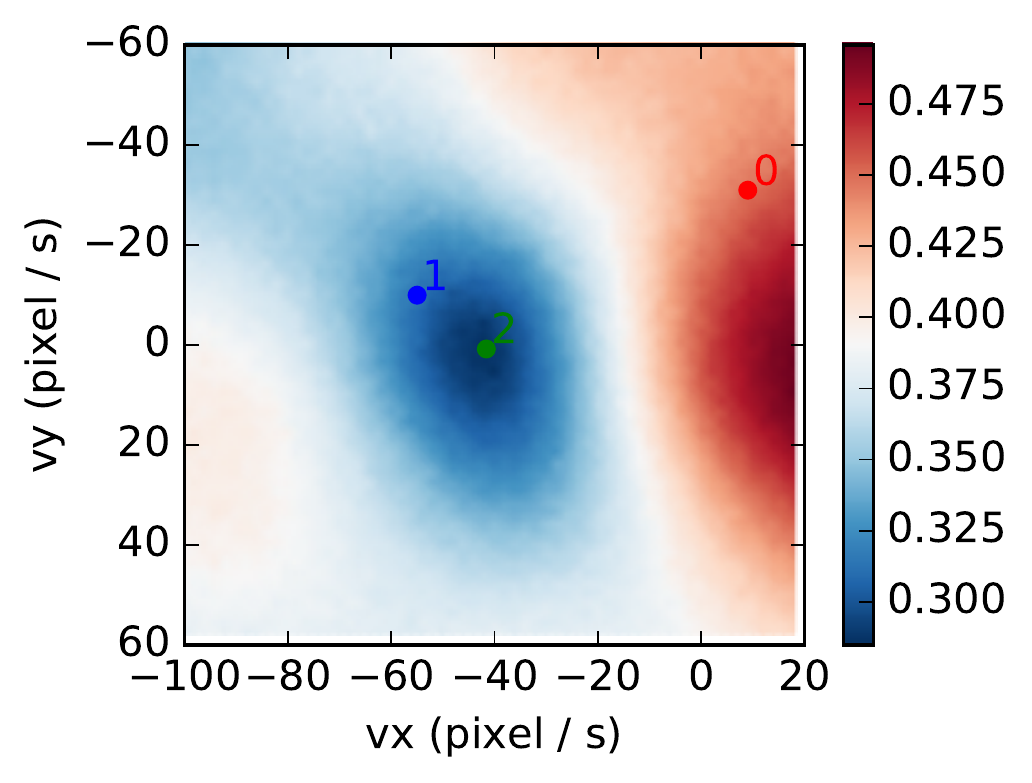}
		\\\addlinespace[-1ex]
		Area (Exp) & Area (Gaussian) & Area (Lorentz) & Area (Hyperbolic)
		\\\addlinespace[1ex]

		\includegraphics[width=\linewidth]{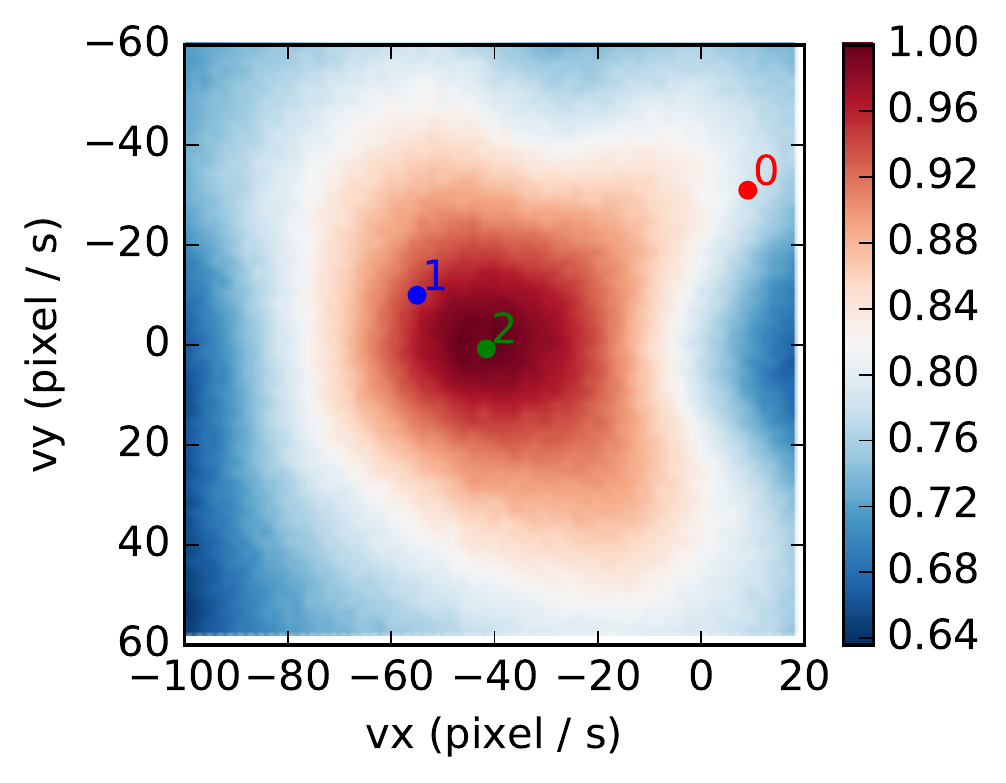}
		&
		\includegraphics[width=\linewidth]{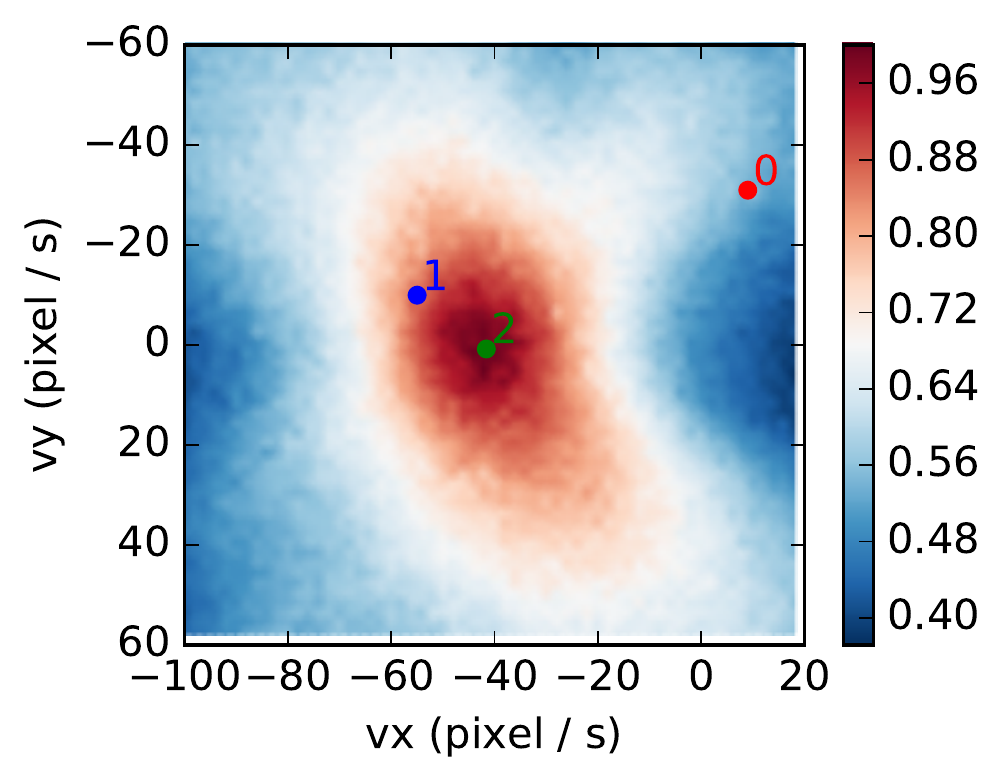}
		&
		\includegraphics[width=\linewidth]{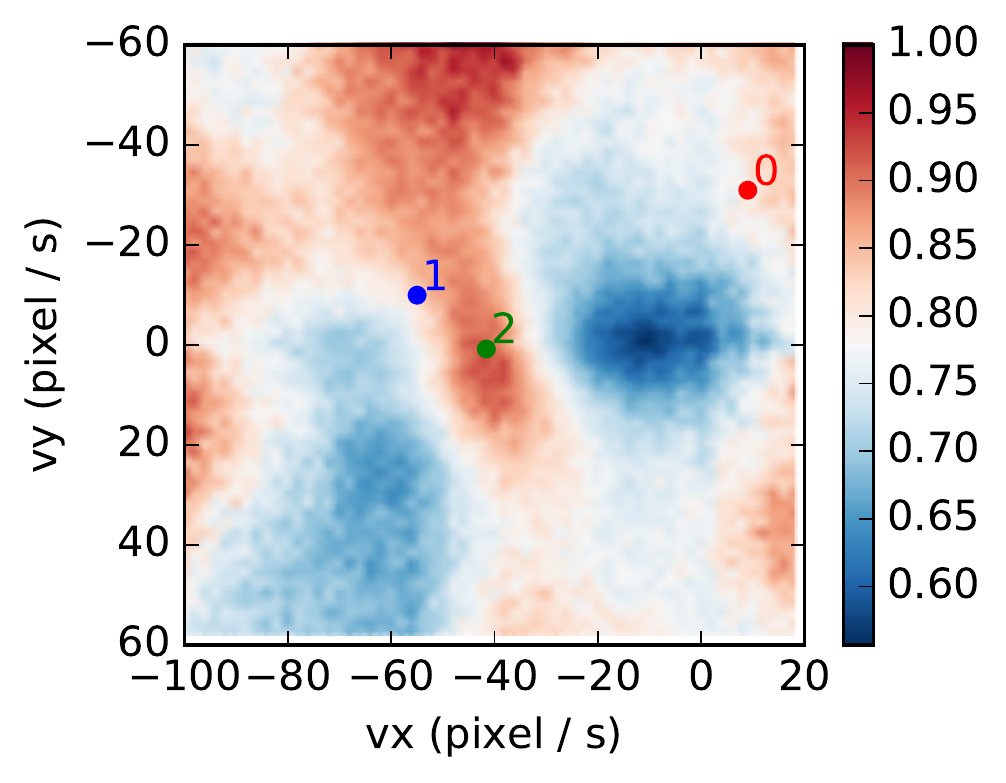}
		&
		\includegraphics[width=\linewidth]{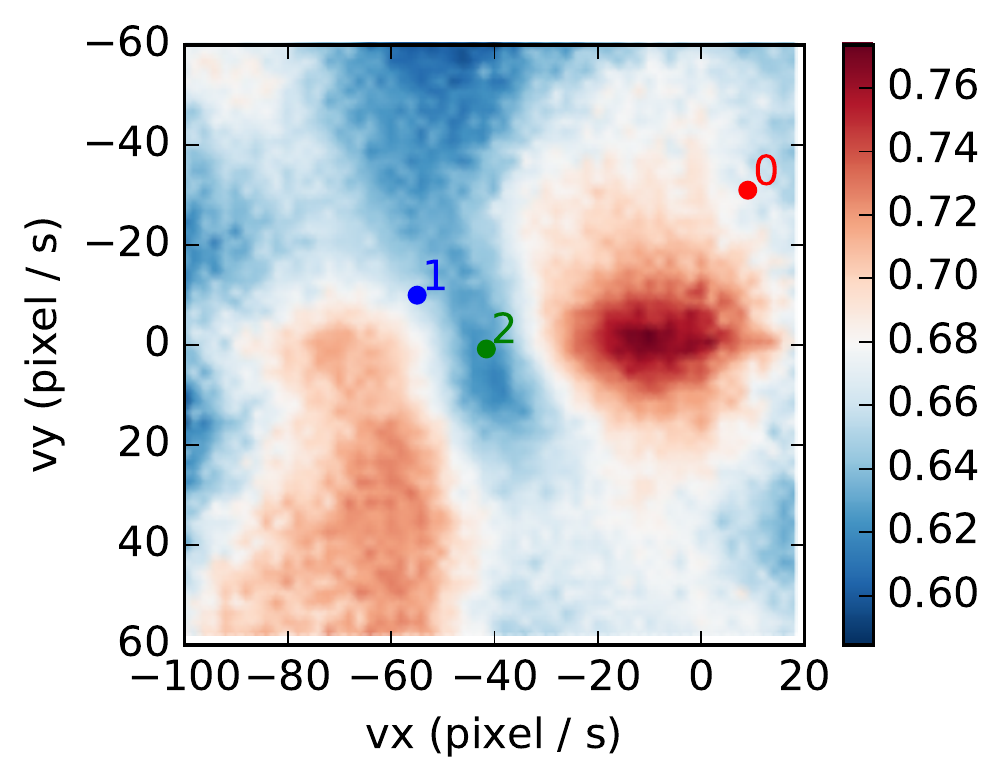}
		\\\addlinespace[-1ex]
		Entropy & Range (Exp) & Geary's $C$ & Moran's Index
		\\\addlinespace[1ex]

		\includegraphics[width=\linewidth]{images/optical_flow/cvpr18/with_polarity/cost_function_0_grad_magnitude_squared.pdf}
		&
		\includegraphics[width=\linewidth]{images/optical_flow/cvpr18/with_polarity/cost_function_0_laplacian_magnitude_squared.pdf}
		&
		\includegraphics[width=\linewidth]{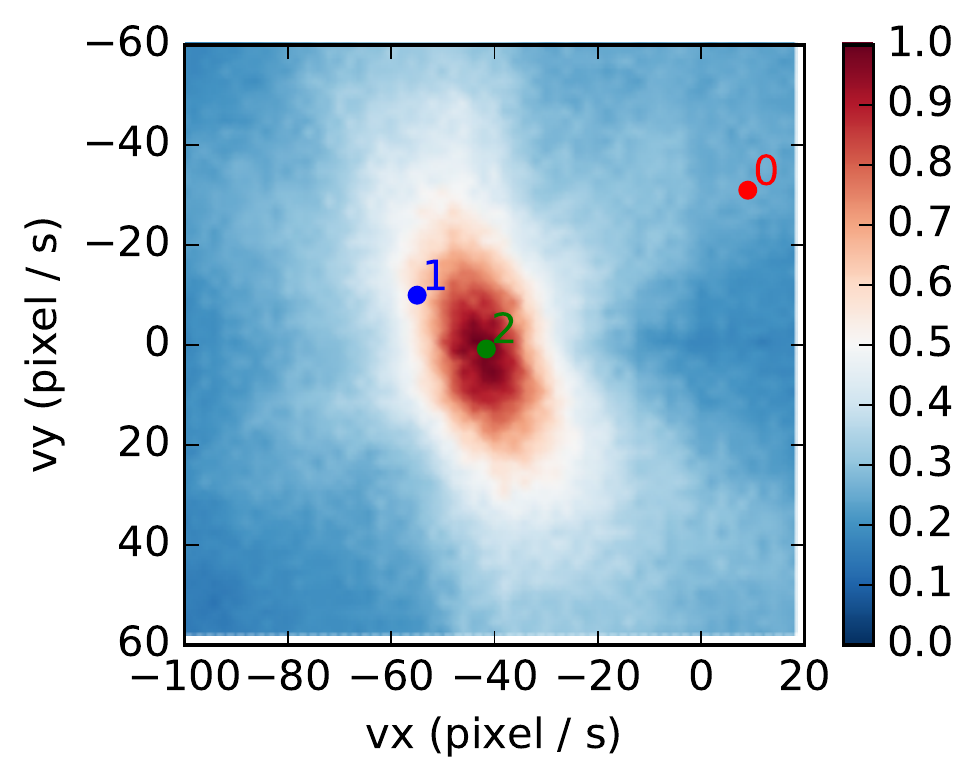}
		&
		\includegraphics[width=\linewidth]{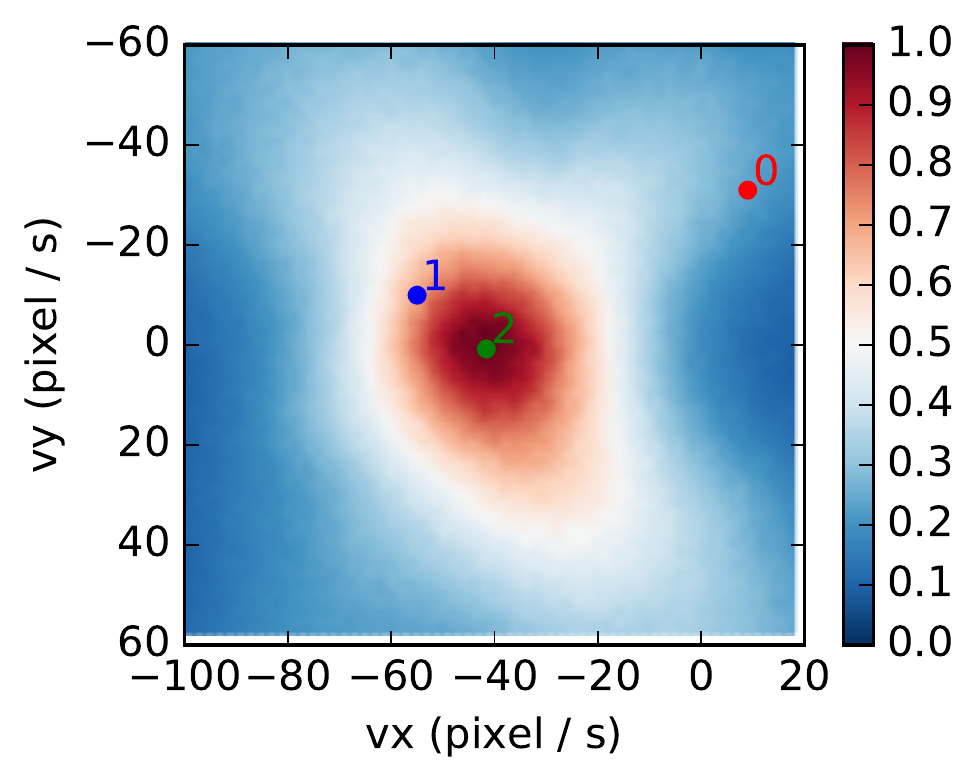}
		\\\addlinespace[-1ex]
		Gradient Magnitude & Laplacian Magnitude & Hessian Magnitude & DoG Magnitude
		\\\addlinespace[1ex]

		\includegraphics[width=\linewidth]{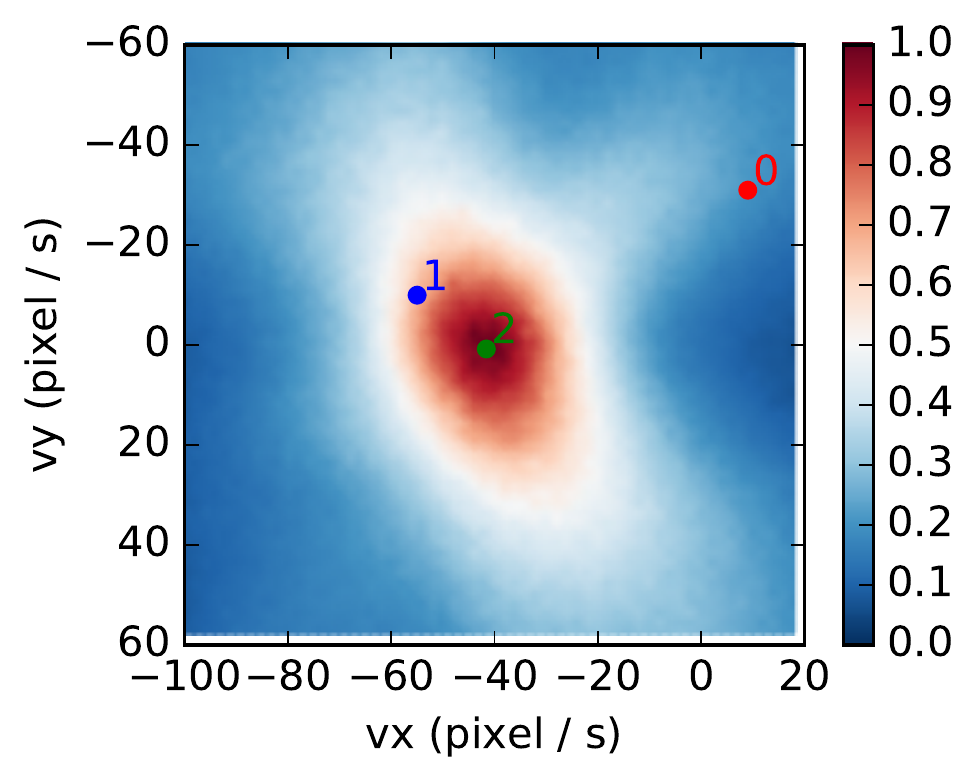}
		&
		\includegraphics[width=\linewidth]{images/optical_flow/cvpr18/with_polarity/cost_function_0_var_of_laplacian.pdf}
		&
		\includegraphics[width=\linewidth]{images/optical_flow/cvpr18/with_polarity/cost_function_0_var_of_grad_magnitude.pdf}
		&
		\includegraphics[width=\linewidth]{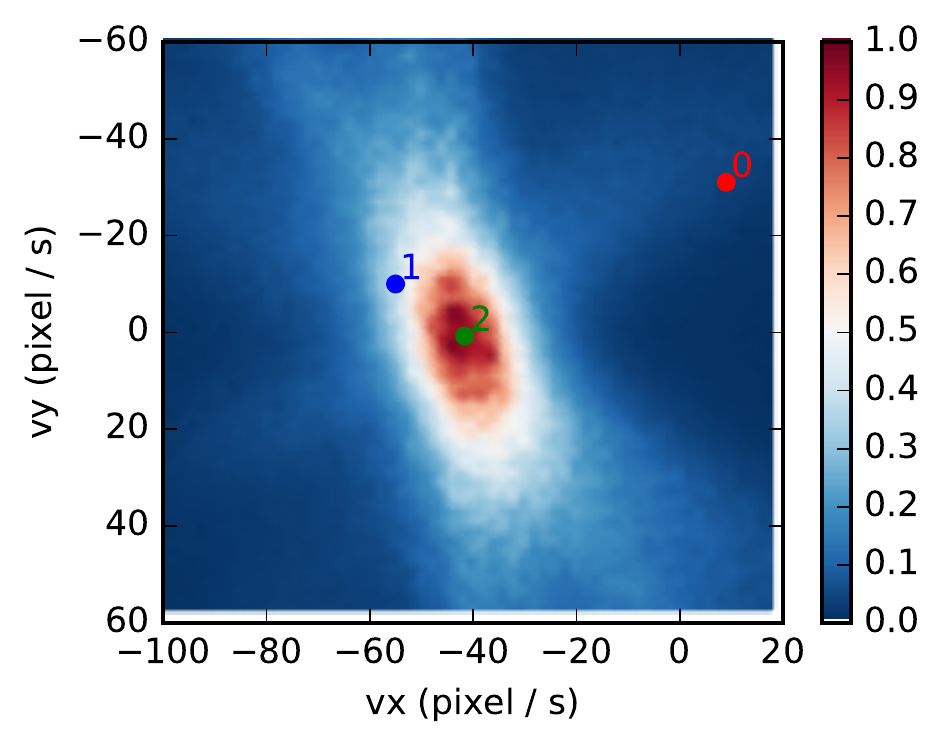}
		\\\addlinespace[-1ex]
		LoG Magnitude & Var. of Laplacian & Var. of Grad. & Var. of Squared Grad.
		\\

	\end{tabular}
	}
	\caption{Visualization of the Focus Loss Functions (as heat maps, pseudo-colored from blue to red).
	Same notation as in Fig.~\ref{fig:boxes:oflowpatch:surfaces}, but using polarity in the IWE.}
	\label{fig:boxes:oflowpatch:surfaces:withpol}
\end{figure*}

Figs.~\ref{fig:boxes:oflowpatch:surfaces} and \ref{fig:boxes:oflowpatch:surfaces:withpol} present experiments with events in a small space-time window ($31 \times 31$ pixels and $\Delta t = \SI{200}{\milli\second}$), yielding approximately $\SI{2000}{events}$, 
from a sequence of the dataset~\cite{Mueggler17ijrr}.
The goal of these figures is to visualize the ``shape'' or ``signature'' of the focus loss functions (as heat maps, pseudo-colored from blue to red).

The top-left image shows the patch on the intensity frame (not used) corresponding to the space-time window of events,
highlighted in yellow, 
and three candidate flow vectors $\{\bparams_i\}_{i=0}^2$ (marked with red, blue and green arrows, respectively).
The ground truth flow is close to $\bparams_2 = (-40,0)\si{pixel/\second}$.
The top row also shows the warped events (IWE patch) using the three flow vectors, without polarity in the IWE (Fig.~\ref{fig:boxes:oflowpatch:surfaces}) or with polarity (Fig.~\ref{fig:boxes:oflowpatch:surfaces:withpol}).
The remaining rows show the focus loss functions in optical flow space, with $\pm \SI{60}{pixel/second}$ around $\bparams_2$.
Some focus functions are designed to be maximized (and therefore should present a local maximum at $\bparams_2$), 
while others are designed to be minimized (and should present a local minimum at $\bparams_2$).

\paragraph*{Without Polarity.}
Rows 2 and 3 of the figures show the variance, MS, MAD, MAV and their aggregation of local versions.
They are all visualized in the range $[0,1]$, by dividing by the maximum value of the focus function.
Without using polarity (Fig.~\ref{fig:boxes:oflowpatch:surfaces}), the variance presents a nice peak at the correct optical flow $\bparams_2$, the peaks of the MS and MAD functions are not as pronounced, and the MAV function does not have the maximum at the ground truth location (we explained that, without polarity, the MAV cannot be used to estimate $\bparams$).
The local versions (third row of Fig.~\ref{fig:boxes:oflowpatch:surfaces}) are slightly narrower than the global versions.
The fourth row presents the four area-based focus losses (Section~\ref{sec:app:MinSupport}), whose goal is to be minimized, and indeed, they present a local minimum at $\bparams_2$.
There are not big differences in these four area-based losses.
The fifth row shows more statistics-based losses.
The range and Geary's C show a local maximum at the correct flow.
Moran's index shows a local minimum, as expected, at the correct flow.
The entropy, without using polarity, does not have a local maximum at the correct flow. 
Instead, using polarity (Fig.~\ref{fig:boxes:oflowpatch:surfaces:withpol}), it does have a local maximum at $\bparams_2$.
The last two rows of Figs.~\ref{fig:boxes:oflowpatch:surfaces} and~\ref{fig:boxes:oflowpatch:surfaces:withpol} show focus loss functions based on the IWE derivatives and their variances (composite losses).
They all present a clear peak at the correct depth (as the case of the variance and local variance); some of them are more narrow than others (all are visualized in the range $[0,1]$, for ease of comparison). 
The gradient magnitude (based on Sobel operator), the DoG magnitude and the LoG magnitude seem to be the smoothest of these two rows.

\paragraph*{With Polarity.}
Fig.~\ref{fig:boxes:oflowpatch:surfaces:withpol} shows the results on the same experiment as Fig.~\ref{fig:boxes:oflowpatch:surfaces}, but using event polarity in the IWE.
In a scene with approximately equal number of dark-to-bright and bright-to-dark transitions, the number of positive and negative events is approximately balanced, and so, the mean of the IWE is approximately zero. 
Thus, in this case, the MS is approximately equal to the variance, and the MAV approximates the MAD. 
This is noticeable in the second row of Fig.~\ref{fig:boxes:oflowpatch:surfaces:withpol}.
A similar trend is observe in the local versions of the four above statistics, albeit the local variance and MAD present narrower peaks than the local MS and MAV, respectively.
The area-based focus functions are computed by splitting the events according to polarity, computing the areas of the two resulting IWEs and adding their area values. 
The corresponding plots, in the fourth row of Fig.~\ref{fig:boxes:oflowpatch:surfaces:withpol} are similar to those without polarity (Fig.~\ref{fig:boxes:oflowpatch:surfaces}, except for the vertical scale).
The entropy and range improve if event polarity is used, basically because the PDF becomes double-sided and it allows us to distinguish positive and negative IWE edges/values.
Moran's index and Geary's $C$ ratio are not good focus losses if polarity is used, since they present brittle local minimum/maximum, respectively.
The derivative-based losses present a clear peak at the correct flow, and slightly more pronounced than their counterparts in Fig.~\ref{fig:boxes:oflowpatch:surfaces}.

\section{Additional Plots on Depth Estimation}

\begin{figure*}
\centering
\subfloat[\label{fig:depth:patch:sliceDSI}Image of Warped Events IWE (i.e., slice of the Disparity Space Image (DSI)~\cite{Rebecq17ijcv}) corresponding to depth $\bparams\equiv Z = \SI{1.11}{\meter}$. 
Color scale represents the event count, from blue (few events) to red (many events).]{\includegraphics[width=0.475\linewidth]{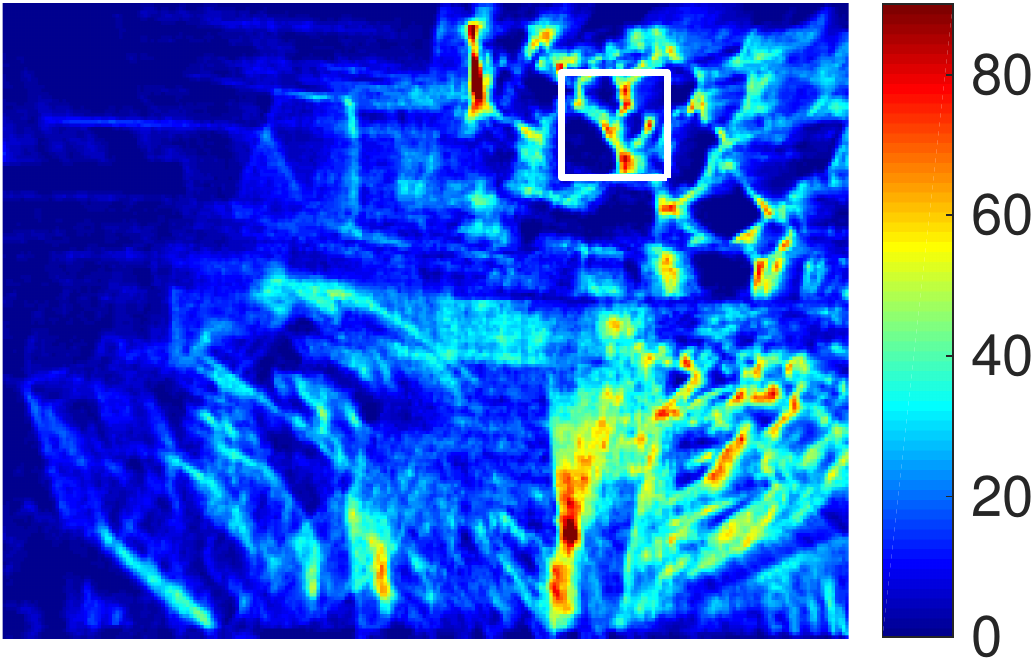}}\\
\subfloat[\label{fig:depth:patch:focus}
Focus functions vs depth for the events corresponding to the patch highlighted in Fig.~\ref{fig:depth:patch:sliceDSI}.]{\includegraphics[width=\linewidth]{images/depth/cvpr19_all_valid_curves-crop.pdf}}
\caption{\label{fig:depth:patch}\emph{Depth Estimation}.
Focus functions in Fig.~\ref{fig:depth:patch:focus} are shown normalized to the range $[0,1]$ for easier visualization.}
\end{figure*}

Fig.~\ref{fig:depth:patch} shows in more detail Fig.~\ref{fig:depthVsDepth}: depth estimation for a patch from a sequence of the dataset~\cite{Mueggler17ijrr}.
The sequence was recorded with a DAVIS camera~\cite{Brandli14ssc}, camera poses were recorded by a motion-capture system, and the camera was calibrated, so the only unknown is the scene structure (i.e., depth). 
Fig.~\ref{fig:depth:patch:focus} shows how the values of the focus functions vary with respect to the depth parameter $\bparams \equiv Z$, for the events corresponding to the patch highlighted in Fig.~\ref{fig:depth:patch:sliceDSI}.
Remarkably, the focus curves have a smooth variation, with a clear extrema around the correct depth value.
Fig.~\ref{fig:depth:patch:sliceDSI} shows the warped events (IWE) for a depth $Z=\SI{1.11}{\meter}$, close to the peaks of the focus curves. 
The IWE is pseudo-colored, from few event count (blue) to large event count (red).
It shows that the events produce a sharp image at the patch location, whereas other parts of the image are ``out of focus'' since events do not align at that depth~\cite{Rebecq17ijcv,Gallego18cvpr}.
Clearly, some focus curves are narrower than others (Fig.~\ref{fig:depth:patch:focus}), showing better properties for determining the optimal depth location.
We observe that the entropy and the range curves have wide peaks, and therefore do not determine depth very precisely.

\global\long\def\sliderdepthwidth{0.195\linewidth}
\begin{figure*}[t!]
	\centering
    {\small
    \setlength{\tabcolsep}{2pt}
	\begin{tabular}{
	>{\centering\arraybackslash}m{0.6cm} 
	>{\centering\arraybackslash}m{\sliderdepthwidth} 
	>{\centering\arraybackslash}m{\sliderdepthwidth}
	>{\centering\arraybackslash}m{\sliderdepthwidth} 
	>{\centering\arraybackslash}m{\sliderdepthwidth}}
		\rotatebox{90}{\makecell{Depth map}}
		&
		\includegraphics[width=\linewidth]{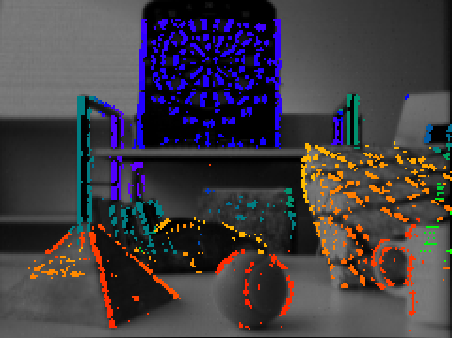}
		&
		\includegraphics[width=\linewidth]{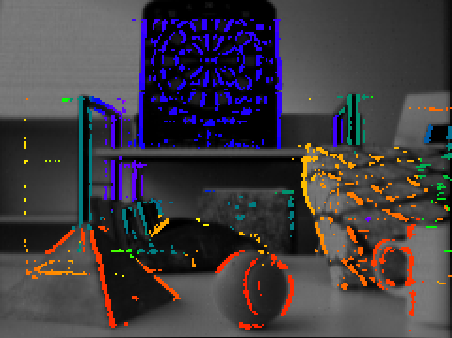}
		&
		\includegraphics[width=\linewidth]{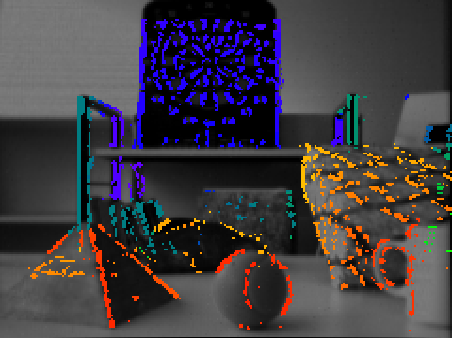}
		&
		\includegraphics[width=\linewidth]{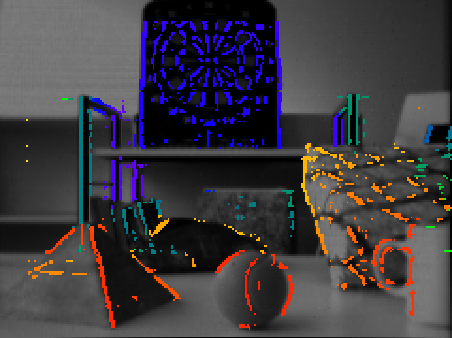}
		\\

		\rotatebox{90}{\makecell{Focus confidence map}}
		&
		\includegraphics[width=\linewidth]{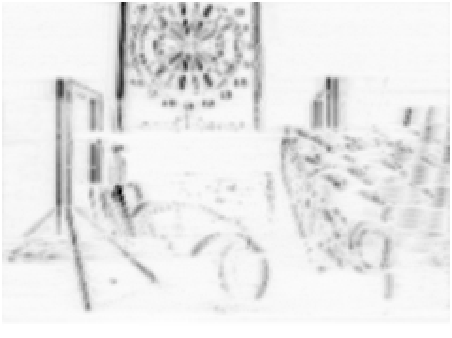}
		&
		\includegraphics[width=\linewidth]{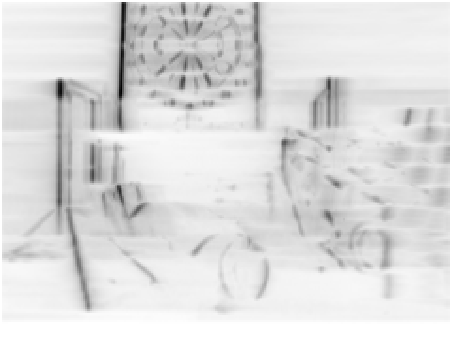}
		&
		\includegraphics[width=\linewidth]{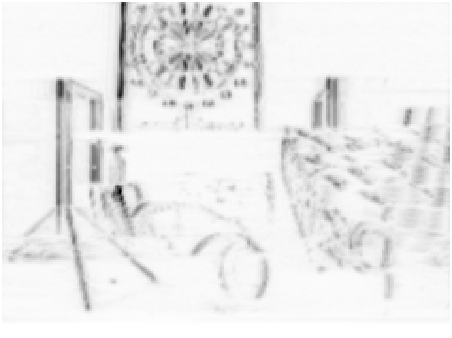}
		&
		\includegraphics[width=\linewidth]{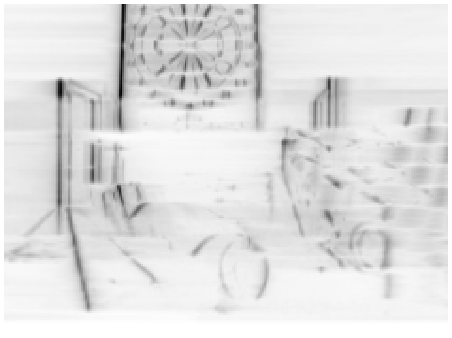}
		\\

		\rotatebox{90}{\makecell{3D point cloud}}
		&
		\includegraphics[width=\linewidth]{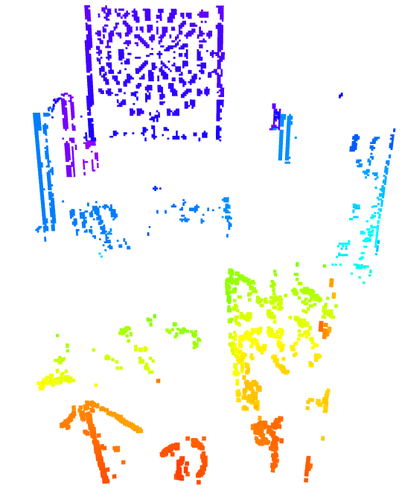}
		&
		\includegraphics[width=\linewidth]{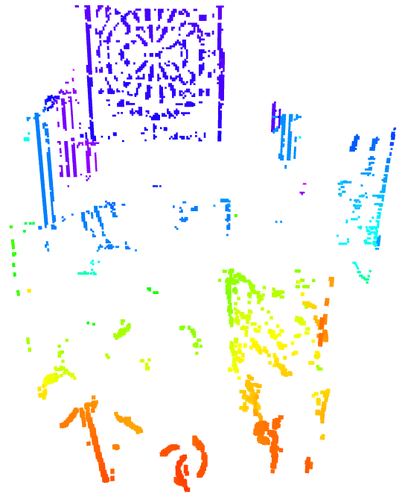}
		&
		\includegraphics[width=\linewidth]{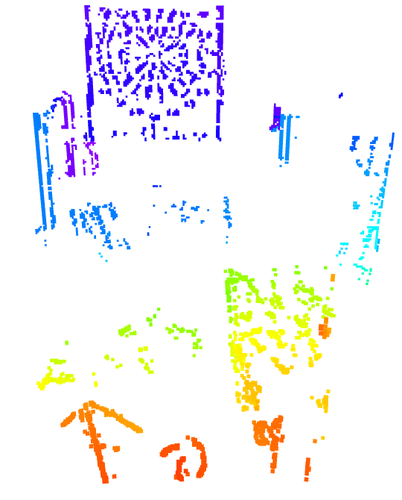}
		&
		\includegraphics[width=\linewidth]{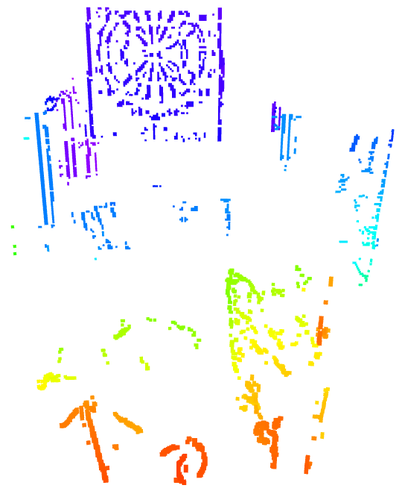}
		\\

		& Local Variance & Local MS & Local MAD & Gradient Magnitude
		\\\addlinespace[3ex]

		\rotatebox{90}{\makecell{Depth map}}
		&
		\includegraphics[width=\linewidth]{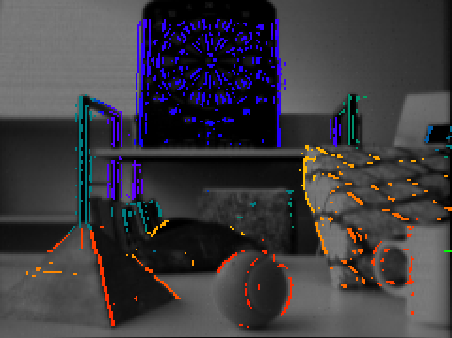}
		&
		\includegraphics[width=\linewidth]{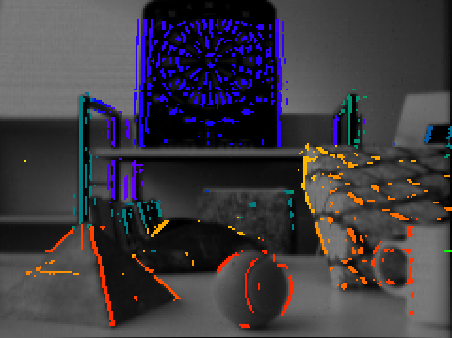}
		&
		\includegraphics[width=\linewidth]{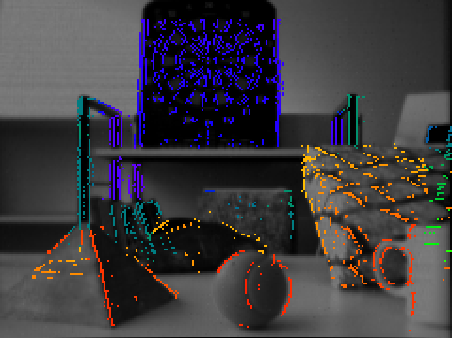}
		&
		\includegraphics[width=\linewidth]{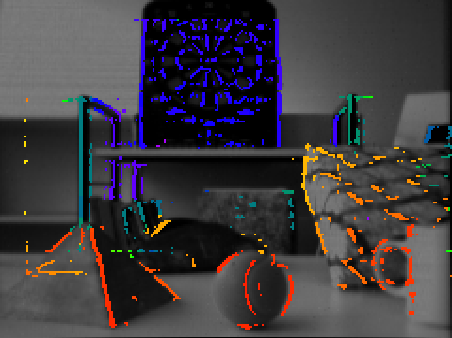}
		\\

		\rotatebox{90}{\makecell{Focus confidence map}}
		&
		\includegraphics[width=\linewidth]{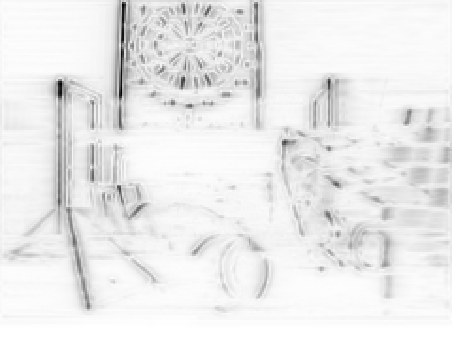}
		&
		\includegraphics[width=\linewidth]{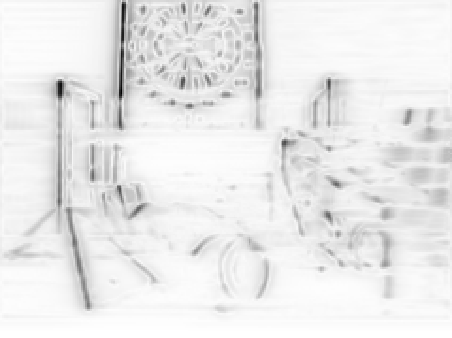}
		&
		\includegraphics[width=\linewidth]{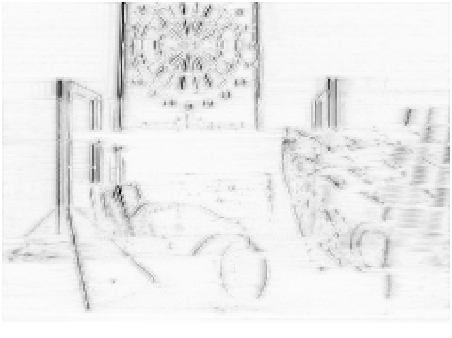}
		&
		\includegraphics[width=\linewidth]{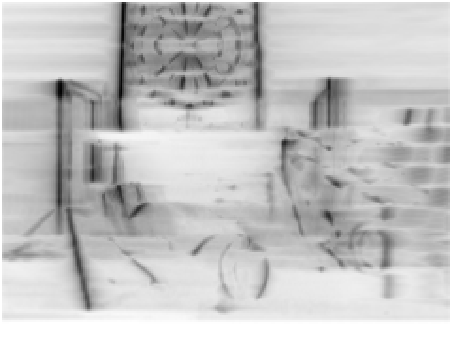}
		\\

		\rotatebox{90}{\makecell{3D point cloud}}
		&
		\includegraphics[width=\linewidth]{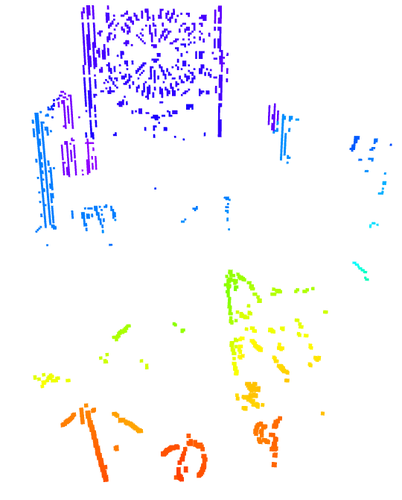}
		&
		\includegraphics[width=\linewidth]{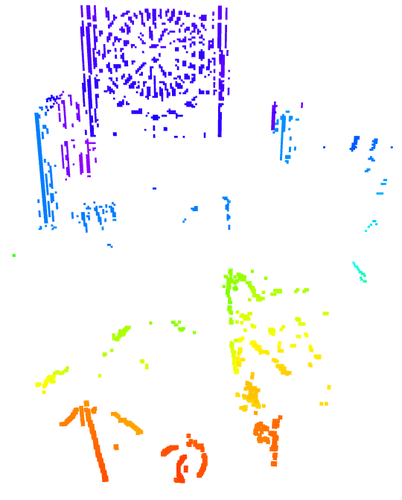}
		&
		\includegraphics[width=\linewidth]{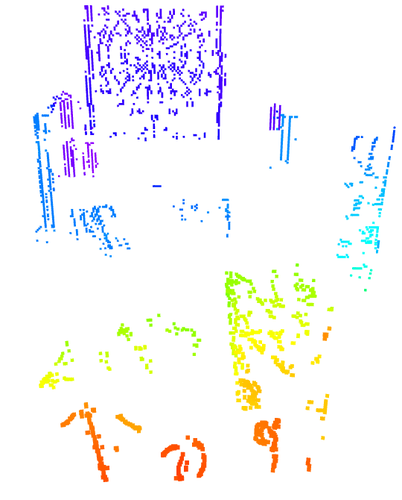}
		&
		\includegraphics[width=\linewidth]{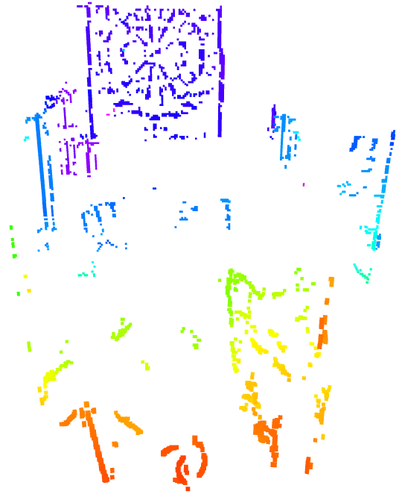}
		\\

		& LoG Magnitude & DoG Magnitude & Laplacian Magnitude & Area (Exp)
		\\
	\end{tabular}
	}
	\caption{\emph{3-D reconstruction} of a scene using several focus loss functions.
	Semi-dense depth maps (overlaid on grayscale frames) and point clouds are pseudo-colored according to depth, from red (close) to blue (far), in the range \SIrange{0.45}{2.4}{\meter}.	
    Sequence {\small\texttt{slider\_depth}} from~\cite{Mueggler17ijrr}, with $\numEvents = \SI{1000000}{events}$ over a time span of \SI{2.93}{\second} and a camera baseline of \SI{85}{\centi\meter}.
	\label{fig:depth:slider}}
\end{figure*}

\paragraph*{Semi-dense 3D Reconstruction.}
Fig.~\ref{fig:depth:slider} shows depth estimation for every pixel of a reference view along the trajectory of the event camera.
For every pixel, we compute focus curves, as those in Fig.~\ref{fig:depth:patch:focus}, and select the depth at the peak.
To capture fine spatial details, the focus functions are computed on patches of $3\times 3$ pixels in the reference view, weighted by a Gaussian kernel to emphasize the contribution of the center pixel.
We also record the value of the focus function at the peak for every pixel of the reference view.
These values are displayed as a ``focus confidence map'' in Fig.~\ref{fig:depth:slider}.
For better visualization, the focus values are represented in negative form, from bright (low focus value) to dark (high focus value).
The confidence map is used to select the pixels in the reference view with largest focus, i.e., the pixels for which depth is most reliably estimated.
The above selection yields a semi-dense depth map, which is displayed color coded, overlaid on the intensity frame from the DAVIS camera~\cite{Brandli14ssc} at the reference view.
We used adaptive thresholding~\cite[p.780]{Gonzalez09book} on the focus confidence map, and a median filter to remove spike noise from the depth map. 
As it is seen, depth is most reliably estimated at strong brightness edges of the scene. 
Finally, the depth map is also visualized as a point cloud, color-coded according to depth (Fig.~\ref{fig:depth:slider}).

The figure compares some representative focus functions.
In general, we obtain good depth 3D reconstructions with the methods tested. 
Some methods produce slightly noisier 3D reconstructions than others, and some recover more edges than others.
This is due to both the shape of the focus confidence maps and the adaptive thresholding parameters.
We observe that focus functions as simple as the local mean square (MS) or the local mean absolute deviation (MAD) produce good results.
These semi-dense 3D reconstruction methods may be used as the mapping module of an event-based visual odometry system, such as~\cite{Rebecq17ral}, to enable camera pose estimation from the 3D reconstructed scene.

\clearpage
\onecolumn
\section{Analytical Derivatives of Focus Loss Functions}
\label{sec:app:Derivs}

In this section, we provide the analytical derivatives of some of the focus loss functions used. 
A significant advantage of the proposed focus loss functions is that they are defined in terms of well-known analytical operations, and so, we can use powerful tools from Calculus to compute and simplify their derivatives.
This becomes useful in the optimization framework of Fig.~\ref{fig:blockDiagramSimple}, both, for speed-up and increased accuracy over numerical derivatives.

Let the vector of parameters defining the space-time warping of events be $\bparams=(\theta_{1},\ldots,\theta_{M})^{\top} \in \R^M$.

\paragraph{Derivative of the IWE.}
The derivative of the IWE~\eqref{eq:IWE} with respect to the warp parameters $\bparams$ is, replacing the Dirac delta with an approximation, such as a Gaussian, $\delta (\bx) \approx \cN(\bx;\bzero,\epsilon^2\mId)$, and using the chain rule,
\begin{equation}
\label{eq:DerivIWE}
\prtl{\IWE}{\bparams} 
= -\sum_{k=1}^{\numEvents} b_k \,\nabla\cN \bigl(\bx - \bx'_k(\bparams);\bzero,\epsilon^2\mId\bigr) \, \prtl{\bx'_k(\bparams)}{\bparams},
\end{equation}
where $\nabla \cN$ is the gradient of the 2D Gaussian PDF,
and the derivative of the warp is a purely geometric term: 
$\prtl{\bx'_k(\bparams)}{\bparams} = \Warp'(\bx'_k,t_k;\bparams).$

\paragraph{Loss Function: Mean Square (MS).}
The derivative of the MS, \eqref{eq:MeanSquareIWE}, is, by the chain rule,
\begin{equation}
\label{eq:DerivGlobalMS}
\prtl{}{\bparams} \text{MS}(\IWE) 
= \frac1{|\Omega|} \int_{\Omega} 2 \IWE(\bx) \prtl{\IWE(\bx)}{\bparams} d \bx.
\end{equation}

\paragraph{Loss Function: Variance.}
The derivative of the variance of the IWE~\eqref{eq:IWEVariance} is given in~\cite{Gallego17ral}.
Letting 
\begin{equation}
\IWE^c(\bx) \doteq \IWE(\bx) - \mu(\IWE)
\end{equation}
be the centered IWE, 
the derivative of the variance is, by the chain rule,
\begin{equation}
\prtl{}{\bparams}\variance(\IWE) 
= \frac{1}{|\Omega|} \int_{\Omega} 2\IWE^c(\bx) \prtl{\IWE^c(\bx)}{\bparams} d\bx,
\end{equation}
(formally, the same formula as~\eqref{eq:DerivGlobalMS}, but with the centered IWE playing the role of the IWE in~\eqref{eq:DerivGlobalMS}) with 
\begin{equation}
\label{eq:DerivCenteredIWE}
\prtl{\IWE^c}{\bparams} 
= \prtl{\IWE}{\bparams}  - \mu\left(\prtl{\IWE}{\bparams} \right)
\end{equation}
since the mean $\mu(\cdot)$ and the derivative are linear operators, and therefore, commute.
The previous result~\eqref{eq:DerivIWE} may be substituted in~\eqref{eq:DerivCenteredIWE}.

\paragraph{Loss Function: Mean Absolute Value (MAV).}
Derivative of the MAV is 
\begin{equation}
\prtl{}{\bparams} \text{MAV}(\IWE) 
= \frac1{|\Omega|} \int_{\Omega} \text{sign}(\IWE(\bx)) \prtl{\IWE(\bx)}{\bparams} d\bx.
\end{equation}
From a numerical point of view, it is sensible to replace $\text{sign}(x)$ by a smooth approximation, e.g., 
$\text{sign}(x) \approx \tanh(kx)$, with parameter $k\gg 1$ controlling the width of the transition around $x=0$ (see~\cite{Gallego14tcsi}).

\paragraph{Loss Function: Mean Absolute Deviation (MAD).}
The derivative of the MAD, \eqref{eq:MeanAbsoluteDevIWE}, is, by the chain rule and using~\eqref{eq:DerivCenteredIWE},
\begin{equation}
\prtl{}{\bparams} \text{MAD}(\IWE)
= \frac1{|\Omega|} \int_{\Omega} \text{sign}(\IWE^c(\bx)) \prtl{\IWE^c(\bx)}{\bparams} d \bx.
\end{equation}

\paragraph{Loss Function: Entropy.}
The derivative of entropy~\eqref{eq:maxEntropy} is, stemming from~\eqref{eq:EntropyLogPDFImageStatement},
\begin{equation}
\label{eq:DerivEntropyOfImageWarpedEv}
\prtl{}{\bparams}H\left(p_{I}(z)\right)
=\int_{\Omega} -\frac{1}{|\Omega|}\frac{p'_{I}(I(\bx))}{p_{I}(I(\bx))}\,
\prtl{I(\bx)}{\bparams}\,d\bx,
\end{equation}
where $p_{I}'(z) = dp_{I}/dz$.
This formula can be obtained by differentiating~\eqref{eq:EntropyLogPDFImageStatement} 
and applying the chain rule,
\begin{equation}
\prtl{}{\bparams}\log p_{I}\left(I(\bx)\right)
=\frac{p'_{I}\left(I(\bx)\right)}{p_{I}\left(I(\bx)\right)}\,\prtl{}{\bparams}I(\bx).
\end{equation}

\emph{Implementation Details:} 
We approximate the density function $p_{I}(z)$ by a smooth histogram of the IWE~\eqref{eq:IWE}.
First, a high-resolution histogram (e.g., 200 bins) is computed and normalized to unit area (like a PDF), and then it is smoothed by a Gaussian filter, e.g., of standard deviation $\sigma=5$ bins.
The smoothed PDF is the convolution
\begin{equation}
\label{eq:IWEPDFsmooth}
p_{I}^{\sigma}(z)\doteq p_{I}(z)\ast\cN(z;0,\sigma^{2}),
\end{equation}
and it is straightforward to show, using the convolution properties and interchanging the order of integration, that it is equivalent to the function obtained by replacing the Dirac delta in~\eqref{eq:PDFImage} with a smooth approximation (such as the Gaussian: $\delta\leftarrow\cN$):
\begin{equation}
p_{I}^{\sigma}(z)=\frac{1}{|\Omega|}\int_{\Omega}\cN(z-I(\bx);0,\sigma^{2})d\bx.
\end{equation}
Smoothing (i.e., filtering) mitigates the noise due to bin discretization and improves robustness (size of the basin of attraction) of the entropy-based focus function~\eqref{eq:maxEntropy}. 
Thus, the term in the integrand of~\eqref{eq:DerivEntropyOfImageWarpedEv} is approximated by
\begin{equation}
\frac{p'_{I}(z)}{p_{I}(z)}\approx\frac{(p_{I}^{\sigma})'(z)}{p_{I}^{\sigma}(z)},
\end{equation}
where the derivative $(p_{I}^{\sigma})'$ is computed using central, finite-differences on the samples of $p_{I}^{\sigma}(z)$. 
Linear interpolation is used to interpolate the samples of $p_{I}^{\sigma}$ and $(p_{I}^{\sigma})'$.

\paragraph{Loss Function: Image Area.}
The derivative of the area~\eqref{eq:minSupp} is
\begin{equation}
\prtl{}{\bparams} \supp(\IWE) =
\int_{\Omega} \rho(\IWE(\bx)) \prtl{\IWE(\bx)}{\bparams} d\bx,
\end{equation}
with weighting functions $\rho(\lambda)$.
See Section~\ref{sec:Focus:MinSupport} for different choices (exponential, Gaussian, Lorentzian and Hyperbolic).

\paragraph{Loss Function: Image Range.}
The derivative of the support of the PDF of the IWE~\eqref{eq:SupportPDFSimplified} is
\begin{equation}
\prtl{}{\bparams} \supp(p_{\IWE})
=\frac{1}{|\Omega|}\int_{\Omega}\rho'\left(p_{I}(I(\bx))\right)\prtl{I(\bx)}{\bparams}d\bx,
\end{equation}
where $\rho'(\lambda) = d \rho / d\lambda$ is the derivative of the weighting function.
This can be shown using the chain rule and the derivative of the PDF of the IWE~\eqref{eq:PDFImage}, expressed in terms of the derivative of the Dirac delta.

\paragraph{Result: Derivative of a Convolution.} 
The derivative of a convolution of the IWE with a kernel $K(\bx)$ is computed component-wise:
\begin{equation}
\label{eq:DerivConvIWE}
\underbrace{\prtl{}{\bparams} \underbrace{(\IWE * K)(\bx)}_{1\times 1}}_{1\times M}
=\left(\ldots, \prtl{}{\theta_i} (\IWE * K)(\bx), \ldots \right)
=\left(\ldots, \prtl{\IWE(\bx)}{\theta_i} * K(\bx), \ldots \right)
\doteq \underbrace{\prtl{\IWE(\bx)}{\bparams}}_{1\times M} * \underbrace{K(\bx)}_{1\times 1}.
\end{equation}

\paragraph{Loss Function: Local MS.}
The derivative of the aggregated local MS of the IWE is, by the chain rule and~\eqref{eq:DerivConvIWE},
\begin{equation}
\prtl{}{\bparams}\int_{\Omega} \IWE^{2}(\bx) * G_{\sigma}(\bx) \,d\bx
=\int_{\Omega} \left( 2 \IWE(\bx) \prtl{\IWE(\bx)}{\bparams}\right) * G_{\sigma}(\bx) \,d\bx.
\end{equation}

\paragraph{Loss Function: Local Variance.}

The derivative of the aggregated local variance of the IWE is, by the chain rule on~\eqref{eq:local_variance_aggr},
\begin{equation}
\prtl{}{\bparams}\int_{\Omega} \variance(\bx; \IWE) \,d\bx
= \int_{\Omega} \prtl{}{\bparams} \variance(\bx; \IWE) \,d\bx,\end{equation}
where, using~\eqref{eq:local_variance_conv} and result~\eqref{eq:DerivConvIWE}, the integrand becomes
\begin{equation}
\prtl{}{\bparams} \variance(\bx; \IWE) 
\stackrel{\eqref{eq:local_variance_conv}}{\approx} \left(2 \IWE(\bx) \prtl{\IWE(\bx)}{\bparams}\right) * G_{\sigma}(\bx) 
-2 (\IWE(\bx)*G_{\sigma}(\bx)) \left( \prtl{\IWE(\bx)}{\bparams} * G_{\sigma}(\bx)\right).
\end{equation}

\paragraph{Loss Function: Local MAV.}
The derivative of the aggregated local MAV of the IWE is, using the chain rule and the compact notation~\eqref{eq:DerivConvIWE},
\begin{equation}
\prtl{}{\bparams}\int_{\Omega} |\IWE(\bx)|*G_{\sigma}(\bx) \,d\bx
=\int_{\Omega} \left( \text{sign}(\IWE(\bx)) \prtl{\IWE(\bx)}{\bparams}\right) * G_{\sigma}(\bx) \,d\bx.
\end{equation}

\paragraph{Loss Function: Local MAD.}
Letting $\IWE^c(\bx) \doteq \IWE(\bx) - \mu(\IWE)(\bx)$ be the locally-centered IWE, with local mean $\mu(\IWE)(\bx)\doteq \IWE(\bx) * G_{\sigma}(\bx)$,
the derivative of the aggregated local MAD of the IWE is, 
using the chain rule and the compact notation~\eqref{eq:DerivConvIWE},
\begin{equation}
\prtl{}{\bparams}\int_{\Omega} |\IWE^c(\bx)|*G_{\sigma}(\bx) \,d\bx
=\int_{\Omega} \left( \text{sign}(\IWE^c(\bx)) \prtl{\IWE^c(\bx)}{\bparams}\right) * G_{\sigma}(\bx) \,d\bx,
\end{equation}
with derivative of the locally-centered IWE
\begin{equation}
\label{eq:DerivCenteredIWELocal}
\prtl{\IWE^c(\bx)}{\bparams} 
= \prtl{\IWE(\bx)}{\bparams} - \mu\left(\prtl{\IWE}{\bparams} \right)(\bx)
\end{equation}
since the local mean and the derivative are linear operators, and hence, commute.

\paragraph{Loss Function: Gradient Magnitude.}
The derivative of the squared magnitude of the gradient of the IWE~\eqref{eq:maxGradient} is
\begin{equation}
\prtl{}{\bparams}\| \nabla I \|_{L^{2}(\Omega)}^{2} 
=\int_{\Omega} 2 (\nabla I(\bx))^\top \prtl{\nabla I(\bx)}{\bparams} d\bx,
\label{eq:DerivGradMagwrtParams}
\end{equation}
with $2\times M$ matrix (assuming equality of mixed derivatives by Schwarz's theorem),
\begin{equation}
\label{eq:DerivGradAtPointwrtParams}
\prtl{\nabla I(\bx)}{\bparams}
=\left(\begin{array}{c}
\prtl{}{\bparams}I_{x}\\[0.5ex]
\prtl{}{\bparams}I_{y}
\end{array}\right)
=\left(\begin{array}{c}
\prtl{}x\prtl I{\bparams}\\[0.5ex]
\prtl{}y\prtl I{\bparams}
\end{array}\right).
\end{equation}
Hence, \eqref{eq:DerivGradMagwrtParams} becomes
\begin{equation}
\eqref{eq:DerivGradMagwrtParams} 
=\int_{\Omega}2\left(I_{x}(\bx)\prtl{}x\left(\prtl{I(\bx)}{\bparams}\right) + I_{x}(\bx)\prtl{}y \left(\prtl{I(\bx)}{\bparams}\right)\right)d\bx.
\end{equation}

\paragraph{Loss Function: Laplacian Magnitude.}
Following similar steps as for the gradient magnitude, the derivative of the magnitude of the Laplacian of the IWE~\eqref{eq:maxLaplacian} is 
\begin{equation}
\prtl{}{\bparams}\|\Delta I\|_{L^{2}(\Omega)}^{2}
= \int_{\Omega}2\Delta I(\bx)\,\Delta \left(\prtl{I(\bx)}{\bparams}\right)d\bx.
\end{equation}
where we defined
\begin{equation}
\label{eq:LaplacianPrtlIWE}
\Delta \left(\prtl{I(\bx)}{\bparams}\right)
\doteq \prtl{^{2}}{x^{2}}\left(\prtl{I(\bx)}{\bparams}\right) + \prtl{^{2}}{y^{2}}\left(\prtl{I(\bx)}{\bparams}\right).
\end{equation}

\paragraph{Loss Function: Hessian Magnitude.}
The derivative of Hessian magnitude~\eqref{eq:maxHessian} is
\begin{equation}
\label{eq:DerivHessian}
\prtl{}{\bparams} \|\hessian(I)\|_{L^{2}(\Omega)}^{2} = 
\sum_{i,j} \prtl{}{\bparams} \|I_{x_i x_j}\|_{L^{2}(\Omega)}^{2},
\end{equation}
where $x_i,x_j$ are variables $x,y$ of the image plane.
Using once more Schwarz's theorem to swap the differentiation order, 
each of the four terms in the sum~\eqref{eq:DerivHessian} is
\begin{equation}
\prtl{}{\bparams} \|I_{x_i x_j}\|_{L^{2}(\Omega)}^{2} 
= \int_{\Omega} 2 I_{x_i x_j}(\bx) \left(\prtl{I(\bx)}{\bparams}\right)_{x_i x_j} d\bx,
\end{equation}
where subscripts $x_i,x_j$ indicate differentiation with respect to those variables.

\paragraph{Loss Function: Difference of Gaussians (DoG).}
The derivative of the squared difference of Gaussians applied to the IWE can be written in a compact way, using~\eqref{eq:DerivConvIWE} on the $\text{DoG}(\bx)\doteq (G_{\sigma_1}-G_{\sigma_2})(\bx)$ filter, with $\sigma_1 > \sigma_2$, as
\begin{equation}
\prtl{}{\bparams} \int_{\Omega} \left(\IWE(\bx) * \text{DoG}(\bx)\right)^{2} \,d\bx
= \int_{\Omega} 2 \left(\IWE(\bx) * \text{DoG}(\bx)\right)
 \left(\prtl{\IWE(\bx)}{\bparams} * \text{DoG}(\bx)\right)
\,d\bx,
\end{equation}
In expanded form,
\begin{equation}
\prtl{}{\bparams} \int_{\Omega} \left(\IWE(\bx) * \text{DoG}(\bx)\right)^{2} \,d\bx
= \int_{\Omega} 2 \left((\IWE*G_{\sigma_1})(\bx) -(\IWE*G_{\sigma_2})(\bx)\right) 
\left(\prtl{\IWE(\bx)}{\bparams}*G_{\sigma_1}(\bx) -\prtl{\IWE(\bx)}{\bparams}*G_{\sigma_2}(\bx)\right)
\,d\bx
\end{equation}
In the experiments, we used $\sigma_1 = \SI{1}{pixel}$ and $\sigma_2 = 3\sigma_1$.

\paragraph{Loss Function: Laplacian of the Gaussian (LoG).}
The derivative of the squared difference of the Laplacian of the Gaussian of the IWE can be computed from the formula for the DoG, 
using the fact that the DoG approximates the LoG if $\sigma_2 \approx 1.6 \sigma_1$.

\paragraph{Loss Function: Variance of the Laplacian.}
Derivative of the variance of the Laplacian:
\begin{equation}
\label{eq:VarLaplacian}
\variance(\Delta\IWE) \doteq \frac1{|\Omega|}\int_{\Omega} (\Delta\IWE(\bx) - \mu_{\Delta\IWE} )^2 d\bx, 
\end{equation}
where the mean is $\mu_{\Delta\IWE}\doteq \frac1{|\Omega|}\int_{\Omega} \Delta\IWE(\bx)d\bx$.
The derivative of~\eqref{eq:VarLaplacian} with respect to the parameters $\bparams$ is, using~\eqref{eq:LaplacianPrtlIWE}, 
\begin{equation}
\label{eq:VarLaplacianDeriv}
\prtl{}{\bparams} \variance(\Delta\IWE) 
= \frac1{|\Omega|}\int_{\Omega} 2(\Delta\IWE - \mu_{\Delta\IWE} )
\left(\Delta\left(\prtl{\IWE}{\bparams}\right) - \mu\left(\Delta\left(\prtl{\IWE}{\bparams}\right)\right) \right)
d\bx.
\end{equation}

\paragraph{Loss Function: Variance of the Squared Gradient Magnitude.}
The derivative of the variance of the squared gradient magnitude
\begin{equation}
\label{eq:VarSquaredGradientMag}
\variance(\|\nabla\IWE\|^2) \doteq \frac1{|\Omega|} 
\int_{\Omega} (\|\nabla\IWE(\bx)\|^2 - \mu_{\|\nabla\IWE\|^2})^2 d\bx,
\end{equation}
with mean $\mu_{\|\nabla\IWE\|^2}\doteq \frac1{|\Omega|}\int_{\Omega} \|\nabla\IWE(\bx)\|^2 d\bx$,
is, using~\eqref{eq:DerivGradAtPointwrtParams},
\begin{equation}
\prtl{}{\bparams}\variance(\|\nabla\IWE\|^2) 
= \frac1{|\Omega|} \int_{\Omega} 2(\|\nabla\IWE\|^2 - \mu_{\|\nabla\IWE\|^2}) \left(2(\nabla \IWE)^\top \prtl{\nabla \IWE}{\bparams} - \mu\left(2(\nabla I)^\top \prtl{\nabla I}{\bparams}\right) \right)
d\bx
\end{equation}

\paragraph{Loss Function: Variance of the Gradient Magnitude.}
Derivative of the variance of the gradient magnitude
\begin{equation}
\label{eq:VarGradientMag}
\variance(\|\nabla\IWE\|) \doteq \frac1{|\Omega|} 
\int_{\Omega} (\|\nabla\IWE(\bx)\| - \mu_{\|\nabla\IWE\|})^2 d\bx,
\end{equation}
with mean $\mu_{\|\nabla\IWE\|}\doteq \frac1{|\Omega|}\int_{\Omega} \|\nabla\IWE(\bx)\| d\bx$,
is, also using~\eqref{eq:DerivGradAtPointwrtParams},
\begin{equation}
\prtl{}{\bparams}\variance(\|\nabla\IWE\|) 
= \frac1{|\Omega|} \int_{\Omega} 2(\|\nabla\IWE\| - \mu_{\|\nabla\IWE\|}) \left(\frac{(\nabla \IWE)^\top}{\|\nabla \IWE\|} \prtl{\nabla \IWE}{\bparams} - \mu\left(\frac{(\nabla \IWE)^\top}{\|\nabla \IWE\|} \prtl{\nabla I}{\bparams}\right) \right)
d\bx.
\end{equation}

\cleardoublepage
\twocolumn
{\small
\bibliographystyle{ieeetr_fullname} %

}

\end{document}